\documentclass[a4paper,fleqn]{cas-dc}

\usepackage[numbers]{natbib}
\usepackage{diagbox}
\usepackage{tikz}
\usepackage[absolute,overlay]{textpos}
\usepackage{blindtext}
\usepackage{relsize}
\usepackage{hyperref}

\tikzset{fontscale/.style={font=\relsize{#1}}}

\begin{document}
\let\WriteBookmarks\relax
\def\floatpagepagefraction{1}
\def\textpagefraction{.001}
\shorttitle{}
\shortauthors{Jon Gutiérrez-Zaballa et~al.}

\begin{textblock*}{21cm}(1.25cm, 0.55cm)
    \begin{tikzpicture}
      \draw (0,0) rectangle (17.25,1); 
      \end{tikzpicture}
  \end{textblock*} 
  
\begin{textblock*}{21cm}(-0.5cm, 0.55cm)
\begin{tikzpicture}
    \node (center) {};
    \path (center)+(10.5,4-0.25) node [fontscale=-1] (name) {Final published version of the manuscript can be found at \href{https://doi.org/10.1016/j.sysarc.2024.103242}{10.1016/j.sysarc.2024.103242}.};
    \path (center)+(10.5,4+0.25) node [fontscale=-1] (name) {\copyright 2024. This manuscript version is made available under the CC-BY-NC-ND 4.0 license.};
    \end{tikzpicture}
\end{textblock*} 

\title [mode = title]{Evaluating Single Event Upsets in Deep Neural Networks for Semantic Segmentation: an embedded system perspective}                      
\tnotemark[1]
\tnotetext[1]{This work was partially supported by the Basque Government under grants PRE\_2023\_2\_0148 and KK-2023/00090, and by the Spanish Ministry of Science and Innovation under grant PID2020-115375RB-I00.}

\author[1]{Jon Gutiérrez-Zaballa}[orcid=0000-0002-6633-4148]
\cormark[1]
\ead{j.gutierrez@ehu.eus}

\affiliation[1]{organization={Dept. of Electronics Technology, University of the Basque Country},
                city={Bilbao},
                postcode={48013}, 
                country={Spain}}

\author[1]{Koldo Basterretxea}[orcid=0000-0002-5934-4735]

\author[2]{Javier Echanobe}[orcid=0000-0002-1064-2555]

\affiliation[2]{organization={Dept. of Electricity and Electronics, University of the Basque Country},
                city={Leioa},
                postcode={48940}, 
                country={Spain}}

\cortext[cor1]{Corresponding author}

\begin{abstract}
As the deployment of artifical intelligence (AI) algorithms at edge devices becomes increasingly prevalent, enhancing the robustness and reliability of autonomous AI-based perception and decision systems is becoming as relevant as precision and performance, especially in applications areas considered safety-critical such as autonomous driving and aerospace.
This paper delves into the robustness assessment in embedded Deep Neural Networks (DNNs), particularly focusing on the impact of parameter perturbations produced by single event upsets (SEUs) on convolutional neural networks (CNN) for image semantic segmentation.
By scrutinizing the layer-by-layer and bit-by-bit sensitivity of various encoder-decoder models to soft errors, this study thoroughly investigates the vulnerability of segmentation DNNs to SEUs and evaluates the consequences of techniques like model pruning and parameter quantization on the robustness of compressed models aimed at embedded implementations.
The findings offer valuable insights into the mechanisms underlying SEU-induced failures that allow for evaluating the robustness of DNNs once trained in advance.
Moreover, based on the collected data, we propose a set of practical lightweight error mitigation techniques with no memory or computational cost suitable for resource-constrained deployments.
The code used to perform the fault injection (FI) campaign is available at \url{https://github.com/jonGuti13/TensorFI2}, while the code to implement proposed techniques is available at \url{https://github.com/jonGuti13/parameterProtection}.
\end{abstract}

\begin{keywords}
Single Bit Upsets \sep Robustness evaluation \sep Model compression \sep Embedded Artificial Intelligence \sep Semantic segmentation
\end{keywords}

\maketitle

\section{Introduction}
There is an increasing interest in developing domain-specific processors for the deployment of artificial intelligence (AI) algorithms at the edge and at the endpoints.
This trend is driven by the necessity of freeing endpoint AI-based systems from having to send acquired data to external, more powerful computing machines to be processed and wait for a response before any action can be executed.
The goal is to make embedded AI systems completely autonomous, avoid security and reliability issues associated with data transmission, and reduce response latency.
Achieving embedded AI processing autonomy makes it possible to extend the applicability of complex AI algorithms, such as the increasingly widespread deep neural networks (DNN), to domains in which there may be severe communication bandwidth constraints, hard reliability and security specifications and/or real-time response requirements.
Examples of such application areas are intelligent vision, autonomous navigation, advanced driving assistance systems (ADAS), autonomous driving systems (ADS), and aerospace applications such as remote sensing and airborne flight control \cite{le2023improving}.

If AI has to be pervasively integrated in systems that autonomously operate in real-world environments, AI processors must meet not only demanding performance specifications, but also strict safety and reliability standards.
This is nowadays a mayor concern in the development of AI-based systems that must be integrated in safety-critical applications such as in the aerospace \cite{ARP4754} and automotive \cite{ISO26262} domains.
For instance, the main agent that jeopardizes the performance in the aerospace applications is the exposure of aircraft and spacecraft electronics to radiation \cite{martinella2021impact}.
When a radiation particle interacts with electronic components, the logical value stored in a cell may be altered, resulting in a single event upset (SEU) or, more specifically, a single bit upset (SBU) when it only affects a single bit or a multiple bit upset (MBU) if multiple bits are affected.
Field programmable gate arrays (FPGA) are particularly sensitive to SBUs, which can affect both the sequential elements of the implemented circuit (flip-flops or Block RAMs) and the configuration memory (LUTs).
In the simplest case, a SBU will cause a bit of one of the implemented artificial neural network (ANN) model parameters to flip.
Depending on the specific position of the SBU, the network output may change and differ from the expected one (critical errors), putting the safety of the system at risk.
In the context of ADS, soft errors at terrestrial altitudes occur due to the following factors: the interaction of high-energy cosmic neutrons with silicon, the interaction of low-energy cosmic neutrons with high concentrations of 10B in the device and the emission of alpha particles from trace radioactive impurities in the device materials \cite{baumann2005soft}.
According to \cite{hirokawa2016multiple}, when exposed to terrestrial neutrons, a bit in SRAM has a probability of $1.33 * 10^{-24}$ to flip in $1ns$.
As explained in \cite{yan2020single}, for a typical neural network of more than 10 million parameters, the probability of at least one bit-flip occurring in a month is 10\%.
This problem is aggravated as a consequence of increasing miniaturization and the use of lower voltage levels in modern integrated circuits.

In addition to background radiation, embedded devices are also vulnerable to 
other factors such as disturbance errors in storage devices (SRAM and DRAM) and also to deliberate malicious bit-flip attacks (BFA).
Moreover, alternative technologies to traditional meta-oxide-semiconductor devices 
for the more efficient in-memory computation, e.g. memristive devices, are particularly prone to bit instabilities \cite{wang2022advances, zhang2022wesco}.
Disturbance errors are a general class of reliability problem that 
affects memory and storage technologies such as SRAM, 
DRAM, flash and hard-disk \cite{kim2014flipping}.
Write failure and read 
disturb are two major causes of SRAM technology failures tightly related to miniaturisation and low-voltage operation \cite{kim2011variation}.
In the case of DRAM, the so-called RowHammer vulnerability is a well-known 
issue that produces data corruption and the 
appearance of multiple bit-flips \cite{kim2014flipping}.
This has attracted the attention 
of attackers, who can maliciously flip memory bits in DRAM 
without the need of any data access privileges \cite{razavi2016flip, yao2020deephammer}.
If extended to safety-
critical domains such as ADS and aerospace, 
the consequences of data-oriented attacks can be catastrophic, especially when the BFAs 
are targeted \cite{wang2023aegis}.

The outstanding performance of state-of-the-art DNNs in many applications is very often at the expense of increasing their size and complexity dramatically \cite{wang2023internimage}.
Although it strongly depends on the design of the digital circuit that implements their functionalities (the "application layer", \cite{1369494}), large models are generally more robust due to their overparameterized nature \cite{ruospo2022selective, taheri2024exploration}.
However, deploying complex AI models into resource-constrained embedded processing systems usually implies, firstly, applying one or various compression techniques to transform large models into the so-called lightweight deep learning models.
These procedures aim to reduce the computational complexity and the memory footprint of original DNNs maintaining comparable performance.
The most used approaches include network pruning, parameter quantization, knowledge distillation, and architecture search \cite{gholami2022survey}.
Secondly, it is usually needed to design custom processing units that help accelerate algorithm execution to meet speed and power consumption specifications.
This is achieved by tailoring specific pipelines for data parallelism and applying arithmetic optimization techniques to make the most of both available memory and computational resources.
Depending on the selected target technology (the "physical layer", \cite{1369494}), e.g., embedded GPUs, FPGAs, or application-specific integrated circuits (ASIC), there exist different design constraints to consider and specific optimization techniques to apply.
In any case, our concern is focused on the effects of SBUs on the parameterization of the DNNs, i.e., how the values of trainable parameters are altered and the potential consequences of these changes on the performance of the DNN models.

In this paper, we analyse the reliability of encoder-decoder DNN models designed for image segmentation tasks against SBU disturbances.
This analysis has been performed with two main objectives in mind.
Firstly, the focus is on the consequences that compression techniques such as model pruning and parameter quantization may have on the robustness of these models.
Secondly, a more comprehensive study has been conducted to precisely determine how such disturbances modify the model performance in the inference process.
The aim was to use this knowledge to propose design techniques that help improving model robustness, and thus system reliability, with no cost in terms of computational complexity and memory footprint.
This work was originally motivated by the necessity to improve the reliability of some image segmentation DNN models designed to be applied in ADAS/ADS when implemented on the target processing devices, both embedded GPUs and FPGAs.
To date, there are few published studies that analyse in detail and in a statistically significant way the impact of SBUs on the robustness of image segmentation DNN models, particularly when subjected to compression for deployments at the edge.
This paper aims to fill this gap by performing a detailed layer-by-layer and bit-level analysis based on SBU emulation to better understand the mechanisms that produce failures in the behaviour of the models.
The analysis has been performed both for 32-bit floating-point and 8-bit quantized representations to cover final implementations on different devices.
Although the study has been necessarily carried out for a specific model architecture and a particular dataset, the purpose of the work is to provide guidelines and analysis tools applicable to other models, particularly those of the encoder-decoder type.
Moreover, it also has served to program a set of memory-, computation- and training-free procedures for the protection, to a certain extent, of the performance of such models against SBUs.
The code has been made public at \cite{myRepo}.

The main contributions of this paper are:
\begin{itemize}
  \item We analyse in a statistically significant way how SBUs alter the performance of encoder-decoder DNN models in image segmentation tasks, determining the sensitivity of the output to single bit-flips in the model parameters according to the layer depth, the parameter type, and their binary representation.
  \item We analyse how and why model pruning affects the robustness of the models against SBU perturbations.
  \item We study the consequences of applying parameter quantization in terms of model robustness.
  \item We provide in \cite{myTensorFI2fork} a modification of the original TensorFI2 \cite{tensorfi2} code to allow applying fault injection (FI) campaigns on quantized TensorFlow Lite models and on big unquantized TensorFlow2 models too.
  \item We describe some simple rules to improve model robustness by protecting model parameters against the consequences of SBUs with no cost on computational complexity neither on the memory footprint.
  \item We provide in \cite{myRepo} the code to perform fault mitigation based on the above mentioned technique.
\end{itemize}

The paper is organised as follows: Section \ref{sec:relatedWork} includes the related work done in the field of DNN robustness evaluation and hardening techniques against SBUs.
Then, in Section \ref{sec:datasetmodel}, we describe the architecture of the encoder-decoder model as well as the compression techniques whose effect on model robustness will be studied.
All the experimental results in relation with the fault injection campaign are collected in Section \ref{sec:fiCampaign}.
Section \ref{sec:protectionMethod} describes the proposed fault mitigation technique.
Finally, we provide the conclusions in Section \ref{sec:conclusiones}.

\section{Related work}\label{sec:relatedWork}
\subsection{Robustness of ANNs against bit-flips}\label{sec:robustness}
Most published papers on ANN robustness against bit-flips focus on classification tasks, while very few analyse semantic segmentation models \cite{systematicReview}.
The majority of the authors have carried out experimental fault injection campaigns combined with statistical analysis, while only a few have developed a theoretical analysis based on the development of a vulnerability model.

Among the experimental works, papers such as \cite{arechiga2018robustness} and \cite{malekzadeh2021impact} test the robustness of several image classification convolutional neural network (CNN) architectures against single bit-flips in their weights, but with relatively superficial analyses.
In \cite{arechiga2018effect}, the authors compare the robustness of multilayer perceptrons and CNNs of different sizes, concluding that larger networks with more layers are more robust than smaller, shallower networks.
Some authors, such as \cite{neggaz2019cnns}, investigate the impact of individual bit-flips while comparing floating-point and fixed-point representations, concluding that the latter are more robust.

Works like \cite{sabbagh2019evaluating} and \cite{goldstein2020reliability} evaluate the effects of both quantization and pruning on the robustness of models.
The former shows that compressed models are more fault-resilient compared to uncompressed models in terms of bit error rate zero accuracy degradation, but statistical significance due to a short number of experiments leads to a high variance in the prediction results.
The latter finds that integer-only quantization acts as a fault mitigation technique by reducing the overall range of the data.
It also concludes that pruning enhances the resilience of deep models as a consequence of the reduction in the occupied area and execution times.

Some works focus on methods to ensure the statistical significance of fault injection campaigns.
In \cite{bosio2019reliability}, the authors present a methodology to evaluate the impact of permanent faults affecting CNNs in automotive applications.
Similarly, \cite{ruospo2023assessing} describes how to correctly specify statistical fault injections and proposes a data analysis on the parameters of a CNN for image classification tasks to reduce the number of fault injections required to achieve statistically significant results.
\cite{hong2019terminal} presents one of the most exhaustive and complete analysis of the vulnerability of 32-bit floating-point CNNs for image classification tasks.
The author considers various factors such as bit position, bit-flip direction, parameter sign, layer width, activation function, normalization, and model architecture.
The key findings are: the vulnerability is caused by drastic spikes in a parameter value, the spikes in positive parameters are more threatening, an activation function that allows negative outputs renders the negative parameters vulnerable as well, and the dropout and batch normalization (BN) layers are ineffective in preventing the massive spikes that bit-flips cause.
In \cite{narayanan2021fault}, the author presents two exhaustive tools for fault injection in models created with both TensorFlow1 and TensorFlow2 \cite{tensorfi2}, and performs a thorough analysis of the consequences of fault injections in different classification models.
Additionally, the article also explores techniques to identify the source of the error by the analysis of changes in the model's predictions.
In \cite{ruospo2020evaluating, ruospo2021investigating}, floating-point and fixed-point data type model implementations are analysed showing that fixed-point data provide the best trade-off between memory footprint reduction and CNN resilience.
Similarly, \cite{syed2021fault} performs a comprehensive layer-wise fault analysis of homogeneously and heterogeneously quantized DNNs, suggesting that quantizing the DNN model heterogeneously to fewer bits helps increase the model's resiliency.

As a consequence of the growing concern about the threat of deliberate BFAs, some authors are studying the effects of simultaneous bit disturbances across 
multiple model parameters.
In the extensive work presented in \cite{wang2023aegis}, 
BFAs are analysed according to their untargeted or targeted nature, and an effective mitigation methodology against targeted BFAs is proposed.
In \cite{he2020defending}, the 
authors apply a progressive bit search algorithm to investigate the effects of bit-flip-based weight attacks and obtain some relevant observations regarding quantized DNN sensitivity: the most sensitive parameters are the ones close to zero (large parameter shift), the weights in the front-end layers are the most sensitive, and BFAs force almost all inputs to be classified into one particular output class.
Similarly, in \cite{li2020defending}, the authors evaluate the accuracy 
degradation of an 8-bit integer quantized DNN as a consequence of untargeted random BFAs, one-shot BFAs and 
progressive BFAs for image classification tasks.
The authors show that with the most exhaustive BFA, i.e. progressive BFA, the accuracy drops to 1\% after just 5 iterations.
This result aligns with the one described in \cite{rakin2019bit}, where the author of the progressive BFA algorithm shows that an 8-bit integer 
quantized ResNet-18 can malfunction after just flipping 13 weight bits out of 93 million.

One of the first works that attempted to formalize a theoretical method 
to evaluate the robustness of DNN models was presented in \cite{bach2015pixel}.
It describes a layer-wise relevance propagation model based on the analysis of the contribution of individual neurons to the final loss.
With a similar approach, gradient-based methods such as that explained in \cite{choi2019sensitivity}, 
analyse the sensitivity of each network layer according to the 
importance (relevance) of the weights during inference.
In \cite{yan2020single}, the authors present a 
theoretical analysis of error propagation on some commonly 
used processing layers in image classification models when the 
sign bit is flipped.
In \cite{zhang2022estimating}, a vulnerability model based on some key features such as gradient and absolute value of the parameters is constructed to reduce the necessary amount of fault injections to perform robustness analysis.
\cite{chen2019binfi} proposes BinFI, a fault injector for finding safety-critical bits in machine learning applications that significantly outperforms random fault injection methods in terms of computational costs. 
Finally, in \cite{zhan2021improving}, the authors 
formulate a bit-flip-based weight fault propagation model 
for 32-bit floating-point CNNs to analyse the robustness of ReLU-based models and propose a hardening method based on function upper bounding.
Nevertheless, the applicability of these methods to segmentation networks is not straightforward because they are primarily designed for classification networks, where the layers under study are typically simple convolutions and fully-connected layers.
Additionally, gradient-based analyses can be inaccurate due to noisy gradients and challenges such as vanishing or dying gradients associated with common activation functions like Sigmoid and ReLU.

All the above-mentioned works are focused on detection and classification DNN models.
Regarding the few papers that deal with segmentation networks, \cite{burelt2022improving} claims to have performed the first fault injection study of DNNs performing semantic segmentation, proposing a critical/tolerable fault categorisation for a 32-bit floating-point DeeplabV3+ network.
\cite{esposito2023reliability} evaluates the reliability of neural networks for various tasks, including semantic segmentation, implemented in 32-bit floating-point representation on a GPU trained with Supervised Compression for Split computing.
In \cite{govarini2023fast}, the author stresses the importance of statistical significance in the analyses, and proposes a fast reliability methodology exploiting statistical fault injections in a U-Net model for image segmentation.
The author performs a comparison of the results obtained with this method to those obtained by random FI and improperly-defined statistical FI campaigns and shows the inability of the latter campaigns to reveal sensitive parameters of U-Net.

\subsection{Hardening and protection of neural networks against fault occurrence}\label{sec:hardeningTechniques}
Most of the papers that propose methods to protect ANNs against faults are also concerned with image classification models.
According to the proposed protection technique, these can be grouped as those that use redundancy, modifications of the activation functions, modifications of the parameters, and modifications of the training/inference process.

One of the most basic methods for error detection and protection is the use of a checksum together with a replication of the model, which, while highly effective, is also prohibitively costly in terms of memory consumption and computing overhead.
\cite{sabena2013evaluation} and \cite{oliveira2014modern} are two examples where full duplication has been shown to be effective.
As proposed in \cite{weigel2017kernel, libano2018selective, bolchini2022selective}, using a proper model sensitivity analysis makes it possible to optimize redundancy to protect only the most critical layers.
In the same direction, \cite{ruospo2022selective} presents a software methodology based on a triple modular redundancy technique to selectively protect a reduced set of critical neurons, under a single fault assumption, by a majority voter correction technique.
In \cite{dos2018analyzing}, the authors characterize fault propagation not only by exposing the FPGA/GPU to neutron beams but also by performing a thorough fault injection campaign.
Based on the observations, the authors propose a strategy to improve system reliability by adapting algorithm-based fault-tolerant solutions to CNNs, i.e., adding invariants to the code for quick error detection or correction.
In \cite{fernandez2022methodology}, the authors propose protecting the matrix multiplication operation of the CNNs in GPUs based on a three-stage methodology to selectively protect CNN layers to achieve the required diagnostic coverage and performance trade-off: sensitivity analysis to misclassification per CNN layers using a statistical fault injection campaign, layer-by-layer performance impact and diagnostic coverage analysis, and selective layer protection.
Finally, \cite{dos2021reduced} proposes a reduced-precision duplication with comparison technique to improve the reliability of computing devices to reduce overhead.
It is suitable for mixed-precision architectures, such as NVIDIA GPUs.

The bounding of activation functions is an alternative technique to redundancy to reduce implementation overhead.
In \cite{hoang2020ft}, the authors perform a comprehensive error resilience analysis of DNNs for image classification tasks subjected to hardware faults in the weight memory.
Then, ClipAct is applied, an error mitigation technique based on squashing the high-intensity faulty activation values to alleviate the impact of faulty weights on predictions.
In \cite{chen2021low}, Ranger is proposed, a low-cost fault corrector which selectively restricts the ranges of values in specific DNN layers to dampen the large deviations typically caused by transient faults leading to silent data corruptions.
In \cite{ghavami2022fitact}, FitAct is proposed, a low-cost approach to enhance the error resilience of DNNs for image classification tasks by deploying fine-grained post-trainable activation functions.
The main idea is to accurately bound the activation value of each individual neuron via neuron-wise bounded activation functions to prevent fault propagation in the network.
In \cite{taheri2024exploration}, the author presents a comprehensive methodology for exploring and enabling a holistic assessment of the trilateral impact of quantization on model accuracy, activation fault reliability, and hardware efficiency.
The framework allows for the application of different quantization-aware techniques, fault injection, and hardware implementation and directly measure the hardware parameters.
A novel lightweight protection technique integrated within the framework that ensures the dependable deployment, evaluating the maximum values of the layers’ activations and replacing the out-ranged values with either lower or upper-bound to avoid fault propagation, is also proposed.
Finally, \cite{zhan2021improving} presents a boundary-aware ReLU to improve the reliability of DNNs by determining an upper bound of the activation function which is theoretically calculated so that the deviation between the boundary and the original output cannot affect the final result.

Some other works explore methods to enhance network robustness by directly modifying the network parameters.
Based on the findings of \cite{bach2015pixel} about the connection between neuron resilience and its contribution to the final prediction score, the authors of \cite{schorn2019efficient} propose a methodology based on architectural and feature optimization to avoid critical bottlenecks and balance the feature criticality inside each layer.
In \cite{jang2021mate}, the authors propose MATE, which is a low-cost CNN weight error correction technique based on the observation that, as all mantissa bits of the weights are not closely related to accuracy, some of them can be replaced with error correction codes.
Therefore, MATE can provide high data protection with no memory overhead.
In \cite{burel2021zero}, the authors perform a comparison of the robustness among 32-bit floating-point, 16-bit floating-point, and 8-bit integer formats for image classification tasks and propose an opportunistic parity method to detect and mask errors with zero storage overhead.

There are also proposals to harden network performance by the modification of the training or inference process.
The authors of \cite{lee2022bipolar} propose a bipolar vector classifier which can be easily integrated with any CNN structure for image classification tasks that end in a fully connected layer.
The underlying idea is that as the weights of the classifier are binarized to $\pm$ 1, the resulting final feature vector will only contain positive/negative values that will also be binarized to $\pm$ 1, and thus, a pattern will be created.
Each class has a specific reference pattern so, to assign the final feature vector to a certain class, it will be compared with all the reference patterns, and the winning class will be the one with the smallest Hamming distance.
In \cite{gambardella2022accelerated}, the authors compare the accuracy of quantized DNNs (QNNs) for image classification tasks during accelerated radiation testing when trained with different methodologies and implemented with a dataflow architecture in an FPGA.
The authors find that QNNs trained with fault-aware training, a kind of data augmentation methodology to allow the network to also experience errors during training, make QNNs more resilient to SEUs in FPGAs.
In \cite{draghetti2019detecting}, an efficient error detection solution for object detection-oriented CNNs is proposed based on the observation that, in the absence of errors, the differences between the input frames and the inference provided by the CNN should be strictly correlated.

Finally, regarding the specific works that focus on the hardening of DNNs against BFAs, \cite{he2020defending} proposes binarization-aware training and piecewise clustering as methods to enhance the resistance of quantized DNNs.
The authors conclude that applying binary quantization, increasing network capacity, and using dropout or Batch-Norm regularization are effective techniques to build resistance, while applying adversarial weight training or pruning is shown to be ineffective.
In \cite{li2020defending}, a three-step algorithm based on mean calculation, quantization and clipping is proposed to reconstruct the perturbed quantized weight matrix to tolerate the faults caused by BFAs.
The overhead introduced by the process is small and the protected DNN can better cope with progressive BFA than non-protect DNN (accuracy of 60\% and 1\% respectively after 5 iterations).
A completely different approach is presented in \cite{wang2023aegis}, where the authors propose a dynamic multi-exit architecture that trains extra internal classifiers for hidden layers that can tolerate the existing attacks which flip bits in one specific layer. 
The experiments are conducted using well-known DNN structures and image classification datasets.
Apart from the above-described methods, which aim to mitigate the effect of such attacks, there are also some other methods that focus on verifying the integrity of the models.
In \cite{9643556}, the authors propose to extract a unique signature from the original DNN prior to deployment and then verify the inference output on-the-fly while trying to add the minimum performance and resource overhead.
Indeed, due to the strict temporal and memory footprint constraints that edge devices must adhere to, the protection systems of interest to us are those that introduce minimal memory/computation overhead.

\section{Model development and optimization}\label{sec:datasetmodel}
\subsection{Architectural design and training}\label{sec:architecture}
The segmentation model used as reference for this study is a U-Net, an encoder-decoder fully convolutional network (FCN) for image segmentation, but adapted to use hyperspectral images (HSI) as inputs.

\begin{figure}[h!]
\centering
\includegraphics[width=6cm]{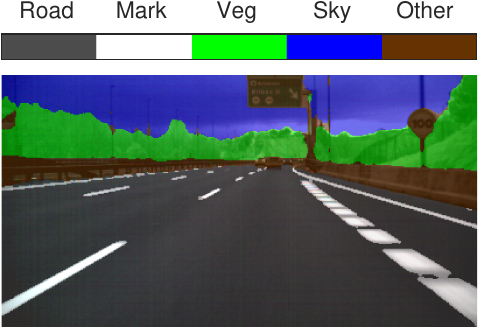}
\caption{A sample segmented image of HSI-Drive v2.0.}
\label{fig:sampleImage}
\end{figure}

The most recent version of the model \cite{gutierrez2023hsi} was trained using the HSI-Drive v2.0 dataset \cite{gutierrez2023hsi}, intended for developing ADAS/ADS systems using HSI (Figure \ref{fig:sampleImage}).
The results on the test set can be found in Table \ref{tab:metricsUnquantizedUnprunedPrunedModel}.

\begin{table}[h!]
\centering
\caption{Segmentation metrics on the test set using the 32-bit floating-point unpruned and pruned models.}
\label{tab:metricsUnquantizedUnprunedPrunedModel}
\resizebox{8cm}{!}{%
\begin{tabular}{c|c|c|c|c|c|c|}
\cline{2-7}
& \multicolumn{3}{c|}{\textbf{Unpruned}} & \multicolumn{3}{c|}{\textbf{Pruned}} \\ \hline
\multicolumn{1}{|c|}{\diagbox[]{\textbf{Class}}{\textbf{Metric}}} & \textbf{Rec.} & \textbf{Prec.} & \textbf{IoU} & \textbf{Rec.} & \textbf{Prec.} & \textbf{IoU} \\ \hline
\multicolumn{1}{|c|}{\textbf{Road}}       & 99.69 & 98.14 & 97.84 & 99.57 & 98.16 & 97.75 \\ \hline
\multicolumn{1}{|c|}{\textbf{Marks}}      & 91.95 & 95.70 & 88.30 & 92.74 & 95.12 & 88.53 \\ \hline
\multicolumn{1}{|c|}{\textbf{Vegetation}} & 97.66 & 96.32 & 94.15 & 97.30 & 96.11 & 93.61 \\ \hline
\multicolumn{1}{|c|}{\textbf{Sky}}        & 96.45 & 96.71 & 93.38 & 96.04 & 97.52 & 93.75 \\ \hline
\multicolumn{1}{|c|}{\textbf{Others}}     & 81.84 & 93.42 & 77.38 & 82.26 & 92.14 & 76.86 \\ \hline
\multicolumn{1}{|c|}{\textbf{Global}}     & 97.23 & 97.17 & 94.67 & 97.12 & 97.06 & 94.48 \\ \hline
\multicolumn{1}{|c|}{\textbf{Weighted}}   & 91.98 & 95.68 & 88.40 & 92.33 & 95.36 & 88.48 \\ \hline
\end{tabular}}
\end{table}

\begin{figure*}[t]
\centering
\includegraphics[width=16.5cm]{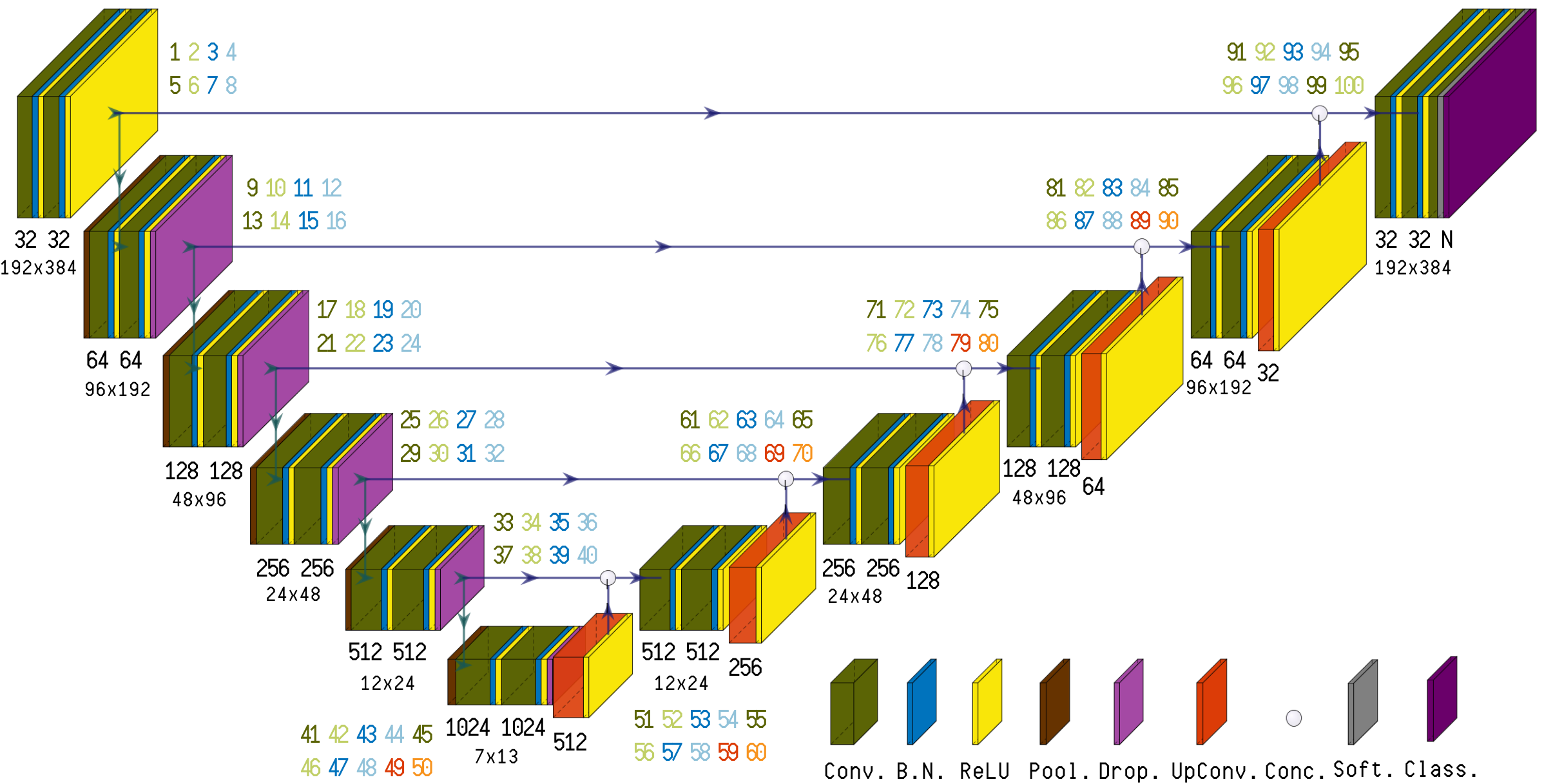}
\caption{Architecture of the unpruned unfolded unquantized U-Net model which has been used as reference.}
\label{fig:unetModelNotPruned}
\end{figure*}

The model, depicted in Figure \ref{fig:unetModelNotPruned}, features a 5-level encoder-decoder architecture comprising two sequences of 3x3 2D convolutional layers (initially with 32 filters) followed by batch normalization and ReLU activation at each level.
Additionally, it includes one 2x2 2D max-pooling layer per encoder level and one 2x2 transposed 2D Convolutional layer per decoder level.
The resulting model comprises 31.14 million parameters and requires 34.60 GFLOPS per inference to execute (Table \ref{tab:modelSize}).
Detailed information regarding the training and testing procedures can be found in \cite{gutierrez2023hsi}.

\begin{table}[h!]
\centering
\caption{Complexity of the different models under study after applying pruning/quantization (status).}
\label{tab:modelSize}
\resizebox{6.5cm}{!}{%
\begin{tabular}{|c|c|c|c|c|}
\hline
\multicolumn{1}{|c|}{\diagbox[]{\textbf{Metric}}{\textbf{Status}}} & \textbf{00} & \textbf{10} & \textbf{01} & \textbf{11} \\ \hline
\textbf{Params. (M)}        & 31.14       & 0.31        & 31.14       & 0.31        \\ \hline
\textbf{Operat. (OPS) }     & 34.60*      & 8.49*       & 34.60       & 8.49       \\ \hline
\textbf{Size (MB)}          & 118.77      & 1.24        & 29.70       & 0.31        \\ \hline
\multicolumn{5}{l}{$*$Before applying quantization, operations are FLOPS.} \\
\end{tabular}}
\end{table}

\subsection{Model Compression}\label{sec:compressionTechniques}
The most prevalent model compression techniques among deep learning developers for later implementation on edge devices involve applying pruning and/or quantization after training the 32-bit floating-point model.
To evaluate how these compression techniques affect the model's robustness against SBUs, several factors must be considered.
First, the size of the model and the architecture designed for its implementation as a digital processor, which directly influences the ratio of unused to used resources and the probability for a SBU to happen.
However, device occupation is not the only parameter influencing the Device Vulnerability Factor (DVF), representing the probability of a configuration bit being critical for the design \cite{deviceReliabilityReport}.
Overparameterized models can absorb more SBUs without producing critical errors since there are more irrelevant weights and biases whose perturbations do not alter the output. 
Thus, it is one of the objectives of this paper to analyse whether large reductions in DNN model sizes can be achieved using these compression techniques without degrading model accuracy or robustness to SBUs.
For this, we are assuming streaming-like architectures with independent resources allocated for all layers so the impact of a SEUs is isolated \cite{TIAN2024115392}.  

\subsection{Pruning}
Pruning facilitates the reduction of both the number of parameters to be stored and the number of computations to be performed.
The concept involves eliminating the least significant parameters of the model.
Depending on which parameters and how they are pruned, pruning can be categorized as fine-grained/sparse/unstructured, where the least important weights are rounded to 0, or coarse-grained/dense/structured, where entire filters are removed from the computational graph.
While the former method generally allows for pruning more weights without harming performance, it is only justified if the processing system is able to optimize multiply and accumulate (MAC) operations with sparse matrices.

In this work, we have applied a conventional structured pruning approach that has been customized to perform model optimization in an iterative manner.
The algorithm basically analyses the computational complexity of each layer while evaluating the impact of the pruning process on the model's accuracy to guarantee negligible impact on overall performance.
As shown in Table \ref{tab:modelSize}, applying this method a 99\% reduction in the number of parameters and a 75\% reduction in the number of operations was achieved.

\subsection{Quantization}
Quantization aims to reduce the number of bits needed to store the model parameters, thereby reducing memory footprint (see Table \ref{tab:modelSize}).
Additionally, it can speed up both data transfer and model inference for custom processor implementations.
Depending on the target device, quantization can be more or less fine-grained in terms of homogeneity, uniformity, scale factor, symmetry, and mixed-precision.

In this article, we have chosen a general and standardized heterogeneous quantization scheme, known as post-training integer quantization (PTQ), as implemented by TensorFlow Lite \cite{jacob2018quantization}.
This procedure converts 32-bit floating-point numbers (weights and activation outputs) to the nearest 8-bit fixed-point numbers, while biases, due to the greater sensitivity of the models to perturbation on these parameters, are converted to 32-bit fixed-point numbers.
This heterogeneous quantization scheme allows for model size reduction while preserving accuracy \cite{jacob2018quantization, vandersteegen2021integer}.
As shown in Table \ref{tab:metricsQuantizedUnprunedPrunedModel}, comparable segmentation metrics are obtained for the quantized model to those with the unquantized model, as quantization schemes applied to already trained models, are simple to apply and produce negligible accuracy degradation for most widely used ANN models \cite{shen2023efficient, wu2020integer}.

\begin{table}[h]
\caption{Segmentation metrics on the test set using the 8-bit integer unpruned and pruned models.}
\label{tab:metricsQuantizedUnprunedPrunedModel}
\centering
\resizebox{8cm}{!}{%
\begin{tabular}{c|c|c|c|c|c|c|}
\cline{2-7}
& \multicolumn{3}{c|}{\textbf{Unpruned}} & \multicolumn{3}{c|}{\textbf{Pruned}} \\ \hline
\multicolumn{1}{|c|}{\diagbox[]{\textbf{Class}}{\textbf{Metric}}} & \textbf{Rec.} & \textbf{Prec.} & \textbf{IoU} & \textbf{Rec.} & \textbf{Prec.} & \textbf{IoU} \\ \hline
\multicolumn{1}{|c|}{\textbf{Road}}       & 99.68 & 98.09 & 97.79 & 99.59 & 98.15 & 97.75 \\ \hline
\multicolumn{1}{|c|}{\textbf{Marks}}      & 91.67 & 95.79 & 88.12 & 92.51 & 95.46 & 88.61 \\ \hline
\multicolumn{1}{|c|}{\textbf{Vegetation}} & 97.63 & 95.76 & 93.59 & 97.38 & 95.88 & 93.47 \\ \hline
\multicolumn{1}{|c|}{\textbf{Sky}}        & 93.60 & 98.29 & 92.10 & 94.02 & 98.45 & 92.65 \\ \hline
\multicolumn{1}{|c|}{\textbf{Others}}     & 82.49 & 92.56 & 77.37 & 82.81 & 91.54 & 76.92 \\ \hline
\multicolumn{1}{|c|}{\textbf{Global}}     & 97.10 & 97.04 & 94.44 & 97.07 & 97.00 & 94.39 \\ \hline
\multicolumn{1}{|c|}{\textbf{Weighted}}   & 91.44 & 96.06 & 88.24 & 91.99 & 95.81 & 88.55 \\ \hline
\end{tabular}}
\end{table}

\subsection{Quantization of Pruning}
To fully compress the model, both techniques can be applied consecutively: first pruning and then quantization.
As shown in Table \ref{tab:metricsQuantizedUnprunedPrunedModel}, after the quantization of the pruned model segmentation accuracy on the test set remains mainly unaltered.
Table \ref{tab:modelSize} sums up model complexity figures for each version of the reference model: original (00), pruned (10), 8-bit quantized (01), and pruned and quantized (11). As can be seen, the original memory footprint is reduced from 118.77 MB to just 317.44 KB.

\begin{figure*}[b]
\centering
\includegraphics[width=16cm]{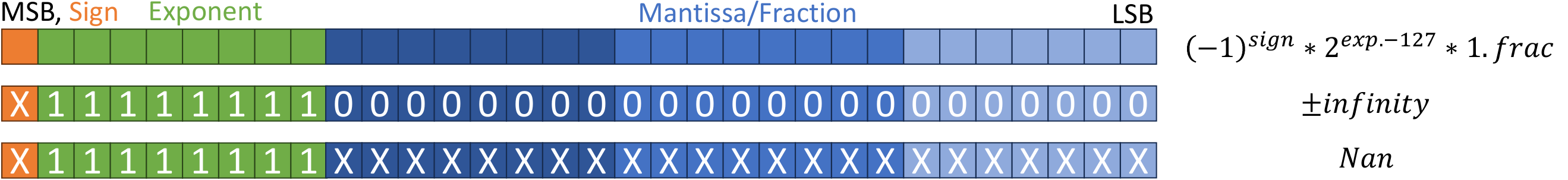}
\caption{Single-precision floating-point format (IEEE 754 format).}
\label{fig:fp32representation}
\end{figure*}

\section{Assessment of the fault injection campaign}\label{sec:fiCampaign}
To test the models' robustness against SBUs, an extensive fault injection campaign was conducted on the aforementioned models using a modified version of the TensorFI2 framework \cite{tensorfi2}.
This modification enables a more memory-efficient implementation by directly accessing the parameters of the model without creating additional copies or intermediate tensors, ensuring that FI on complex, large models does not result in an excessive memory overhead.
The original code has also been extended to support quantized TensorFlow Lite models and is available at \cite{myTensorFI2fork}.
To evaluate the perturbation produced by injected faults, the corrupted FCNs have been assessed using ten test images that represent the diversity of driving conditions in the dataset described in Section \ref{sec:datasetmodel}.

Performed FI campaign involved injecting single bit-flips into the parameters of the FCN model under analysis.
Parameter sets include the weights/kernels of the 2D convolution layers ($conv2D_k$), bias of the 2D convolution layers ($conv2D_b$), weights/kernels of the 2D transposed convolution layers ($conv2Dtr_k$), bias of the 2D transposed convolution layers ($conv2Dtr_b$), gamma of the batch normalization layers ($BN_\gamma$), and beta of the batch normalization layers ($BN_\beta$).
Thus, in what follows, $p1$ set of the network refers to the weights of the first $conv2D$ layer, while $p2$ set represents the biases of that layer.
In like manner, when it's mentioned that an error has been injected into parameter $x$, it means it has been injected into one of the $m$ elements that comprise parameter $x$ set.

As a general rule, for a given parameter, the higher the bit position in which the fault is injected, the greater its impact on the output, disregarding the sign bit.
However, not every layer and every parameter contributes equally to the output.
Previous studies \cite{bosio2019reliability, ruospo2020evaluating, fernandez2022methodology} focused on tasks such as image classification and object detection, have shown that faults injected in the exponent bits (interval $[23-30]$) are more likely to alter the output.
To verify whether this is also generally the case for this encoder-decoder model aimed to image segmentation and in order to set reasonable bounds for the range of bits to be modified, a first round of 150 faults per-layer spanning the entire range $[0-31]$ were injected.
We could verify that in some cases, bit-flips in the range $[20-23]$ and in the sign bit ($31$) also modified significantly the inference result.
Based on these preliminary results, a comprehensive fault injection campaign was conducted on bit positions $[20-31]$.

The quantity of fault injections was set to $1550$ single bit-flip faults per layer, totaling $155000$ injections to assure statistically representative experiments according to Equation \ref{equ:statisticallySignificant} as proposed in \cite{statisticalFaultInjection}.

\begin{equation}
    n = \frac{N}{1 + e^2 * \frac{N-1}{t^2 * p * (1-p)}}
    \label{equ:statisticallySignificant}
\end{equation}

where $N$ represents the number of possible faults per-layer (number of parameters * parameter bit-width), $e$ denotes the error margin, which was set to $0.025$ ($2.5\%$), $t$ is the cut-off point corresponding with the confidence level, set to $1.96$ ($95\%$), and $p$ is the estimated probability of faults resulting in a failure, set to $0.5$ since, as it is a priori unknown, a conservative approach is to use the value that maximizes the sample size \cite{statisticalFaultInjection}.
This is also done in other studies such as \cite{bosio2019reliability, ruospo2020evaluating}.
Finally, $n$ represents the minimum number of injections for the study to be statistically significant, capped at $1550$ for the chosen parameters.

To assess the impact of the fault injection campaign, it is necessary to first define what constitutes an error.
Contrary to the definition given in classification tasks, where the output is a single class and a prediction is deemed erroneous if the top-ranked class predicted by the original model changes, defining errors in segmentation tasks is more challenging.
Changes in the predicted class of some isolated pixels may not significantly alter the overall interpretation of the image, so in order to have an unequivocally defined metric, in this analysis the error rates are defined related to changes in the predicted classes at any pixels in the test images.

\subsection{Fault injections in unquantized models}
As a consequence of the 32-bit floating-point data representation, the effect of a bit-flip varies with its position (Figure \ref{fig:fp32representation}).
To maintain clarity in terminology, the most significant bit (MSB), the sign bit, is assigned position 31, while the least significant bit (LSB) is assigned position 0.

\begin{figure}[h!]
\centering
\includegraphics[width=7.75cm]{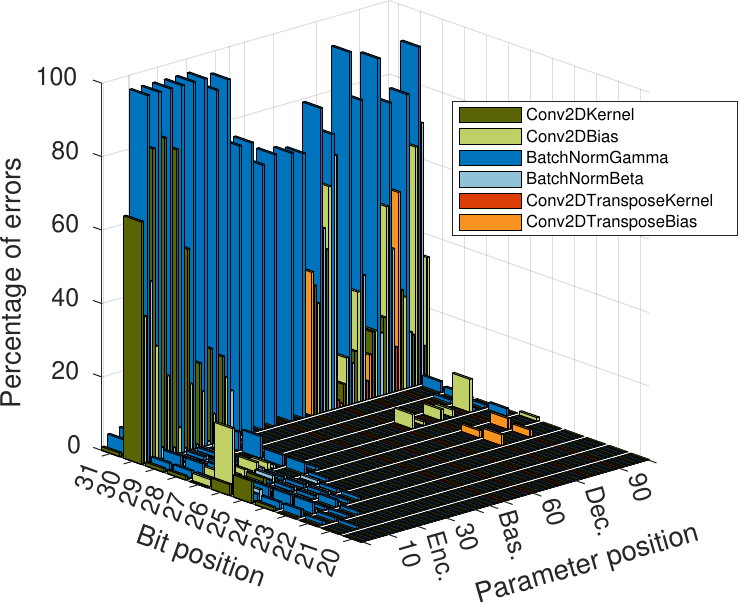}
\caption{Bit-flip error rate on the original unpruned model.}
\label{fig:bitFlipErrorNotPruned}
\end{figure}

\begin{figure}[h!]
\centering
\includegraphics[width=7.75cm]{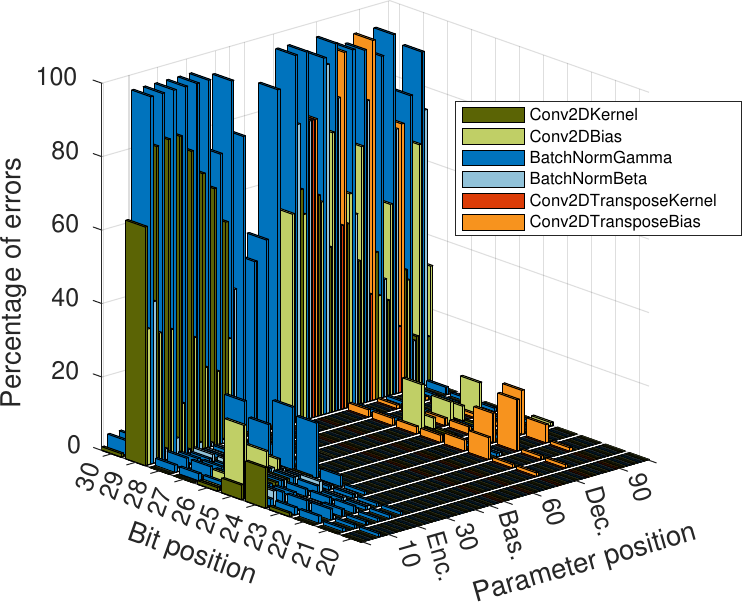}
\caption{Bit-flip error rate on the original pruned model.}
\label{fig:bitFlipErrorPruned}
\end{figure}

Figures \ref{fig:bitFlipErrorNotPruned} and \ref{fig:bitFlipErrorPruned} depict the bit-flip error rate according to the bit position and the parameter position for both the unpruned and pruned models.
As expected, the unpruned model exhibits greater robustness and can better withstand the injected faults.
Once the general statistical results have been observed, let us now analyse in detail the insights of the sensitivity of the model performance to perturbations in the parameters.

\subsubsection{Analysis of robustness of the unpruned model}\label{sec:lastConvolutionNotPrunedNotQuantized}

\paragraph{Output $conv2D$ layer}
$conv2D\_22$ ($p99$-$p100$ sets in Figure \ref{fig:unetModelNotPruned}) consists of $n_{class}$ filters of $1 * 1 * C$ dimensions and $n_{class}$ biases, where $n_{class}$ is the number of classes to be predicted ($6$ in this model) and C is the number of output channels from the previous layer, $conv2D\_21$ ($32$ in this model).

\begin{figure}[h!]
\centering
\includegraphics[width=7.75cm]{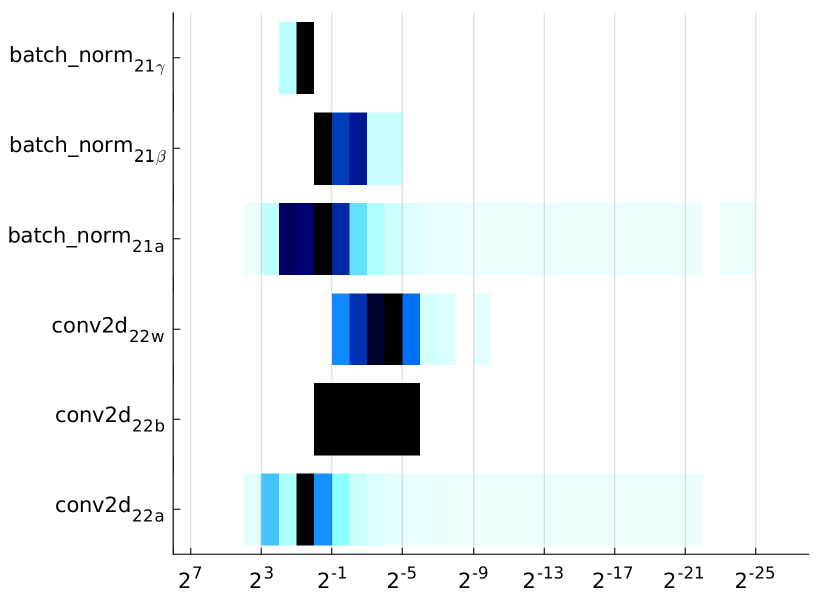}
\caption{Graphic representation of the unpruned model's calibration study (layers $conv2D\_21$ to $conv2D\_22$).}
\label{fig:calibrationUnpruned_21_22}
\end{figure}

A signal calibration analysis (see Figure \ref{fig:calibrationUnpruned_21_22}) reveals that the activations of the last convolutional layer lie within the range $[-6.7656$, $11.7859]$, while the absolute values of the bias parameters turn out to be smaller than $1$ ($-0.8494$, $0.3171$, $-0.0275$, $0.0394$, $-0.1706$, $0.1090$).
Consequently, a change in the sign (bit 31) of one of the biases is unlikely to alter the winning class.
However, a bit-flip in bit $30$ will significantly increase the bias' magnitude, because in 32-bit floating-point representation all six biases contain a '0' in their 30th bit.
The impact of a bit-flip on bit 30 depends on two factors: the sign of the bias and the probability of the model predicting each class.
In the scenario where the bit-flip occurs in a negative bias (biases 0, 2, and 4), the bias becomes so negative that the associated class will never be predicted by the model.
Consequently, if the original model predicted many pixels of that class in an image, produced error rate will be very high; conversely, if the presence of that class was low, produced error will be small.
If the bit-flip happens in a positive bias (biases 1, 3, and 5), the bias becomes large enough so that the associated class is always predicted by the model.
Again, depending on the frequency of that class in an image, the resulting error rate can be high or negligible.
Both the sign of the bias associated to each class and the probability of the model predicting each class can be known beforehand.
Therefore, the error rate can be predicted using Equation \ref{equ:ProbabilityBeforehand}.

\begin{equation}
  \% error_{30} = \sum_{j=0}^{5} P_{fi}^{j} * P_{m}^{j}
  \label{equ:ProbabilityBeforehand}
\end{equation}

where $j$ is the class index, $P_{fi}^j$ is the probability of a bit-flip occurring in class/bias $j$ (assumed to be $\frac{1}{6}$) and $P_{m}^{j}$ is the probability of the faultless model predicting $j$ as the output class.
Applying this equation yields:

\begin{equation*}
\begin{split}
    \% error_{30} & = \frac{1}{6} (0 + 55.09 + 4.41 + 73.05 + 7.47 + 83.73)
    \\ & = 37.29
\end{split}
\end{equation*}

The slight discrepancy between this value $(37.29\%)$ and the one obtained from the experimental data (34\% in Figure \ref{fig:bitFlipErrorNotPruned}) is due to the statistical error of the fault injection since it may not result in the exact same number of flips for each of the 6 classes (assuming a constant probability of $\frac{1}{6}$ is thus only an approximation).
Additionally, the highly unbalanced terms in the equation make the effect of the non-constant probability more noticeable.

Regarding the weights, $conv2D\_22_w$, they are multiplied by the output activation of layer $conv2D\_21$, which, due to the use of ReLU activation functions, is always $\geq 0$.
Hence, if the bit-flip occurs in a weight that is multiplied by a 0-valued activation from $conv2D\_21$, the model will be immune to it (unless the bit-flip produces a non-desirable special value such as $NaN$ in the weight itself).
The impact of a flip in the sign bit is considered negligible and the effect of a flip in bit 30, similar to biases, will only be relevant if the weight is negative and belongs to a filter of the class predicted by the model or if it is positive but does not belong to a filter of the class predicted by the model.

\paragraph{Gamma parameter in batch normalization layers}\label{sec:gammaUnpruned}
The gamma ($\gamma$) parameter of the BN layers is the most sensitive one in the model (located in the third, fourth, seventh and eighth positions in each of the 8-parameter blocks of Figure \ref{fig:unetModelNotPruned}).
This high sensitivity is due to the very nature of the BN operation, as specified in Equation \ref{equ:BNoperation} for a single pixel of a single channel of an activation map (the final result is a scalar value):

\begin{equation}
BN(Conv2D(\textbf{x})) = \gamma \frac{(\sum \textbf{w} \odot \textbf{x} + b) - \mu}{\sqrt{\sigma}} + \beta
\label{equ:BNoperation}
\end{equation}

where $\textbf{x}$ represents an input 3D array, $\textbf{w}$ is the weight 3D array of a filter of the $conv2D$ layer, $b$ is the bias of a filter of the $conv2D$ layer, $\gamma$ is the positive gamma parameter of the BN, $\beta$ is the bias parameter of the BN, and $\sigma$ and $\mu$ are the variance and the mean of the training data.

\begin{figure}[h!]
\centering
\includegraphics[width=7.75cm]{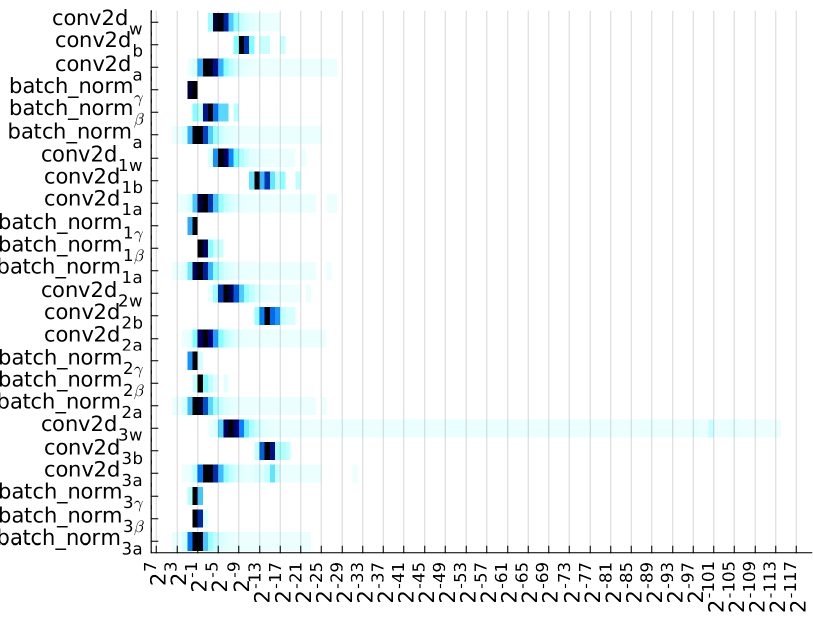}
\caption{Graphic representation of the unpruned model's calibration study (layers $conv2D$ to $conv2D\_3$).}
\label{fig:calibrationUnpruned_0_3}
\end{figure}

\begin{figure}[h!]
\centering
\includegraphics[width=7.75cm]{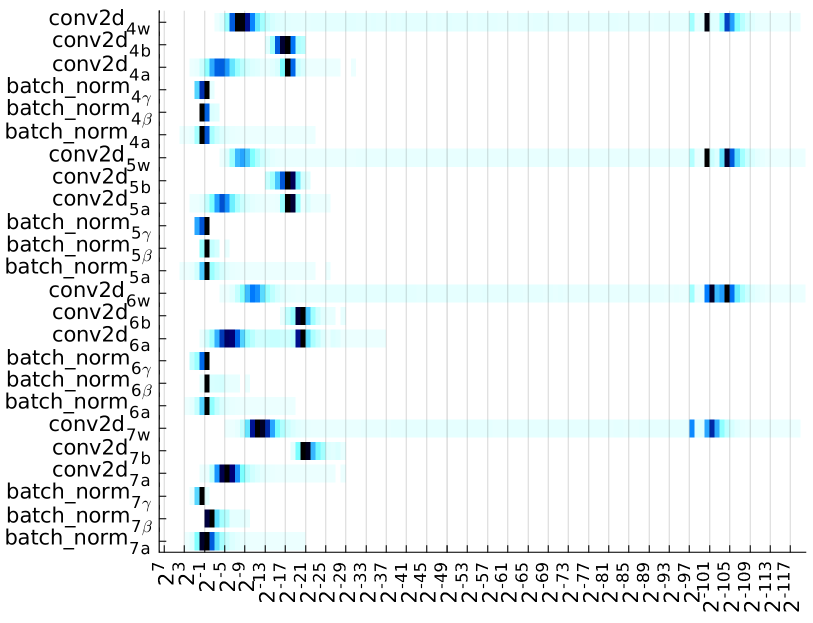}
\caption{Graphic representation of the unpruned model's calibration study (layers $conv2D\_4$ to $conv2D\_7$).}
\label{fig:calibrationUnpruned_4_7}
\end{figure}

It so happens that in this model, $\gamma$ values are positive, smaller than 2 but higher than 0.1.
Thus, in this case, three different scenarios can be considered: values higher than 1, equal to 1, and smaller than 1.
In the first case, except for bit 30, the remaining 7 bits of the exponent are '1's and the mantissa is non-zero, so a bit-flip in bit 30 converts the original value to a $NaN$ (Figure \ref{fig:fp32representation}).
The propagation of the $NaN$ through the network is unavoidable, resulting in a completely wrong output map.
In the second case, where the mantissa is zero, the same bit-flip converts the value 1 into $+\infty$.
This situation is akin to the previous one, as the network cannot digest this error except in situations where the value of the weight of the next $conv2D$ is negative so as the ReLU transforms the $-\infty$ into a 0.
In the last case, considering 0.1 as an example, the bit-flip in position 30 will increase the value to something around $e^{37}$, which is near the maximum representable value ($\approx 3.4028e^{38}$).
Moreover, as Equation \ref{equ:BNoperation} shows, that result is divided by the square root of the variance of the data, which is usually much smaller than 1.
Consequently, the resulting value would exceed the maximum value and would saturate to $\pm$ $\infty$.
In fact, as observed in the calibration analysis (e.g. Figures \ref{fig:calibrationUnpruned_0_3}, and \ref{fig:calibrationUnpruned_4_7}), BN layers always increase the range of the processed data.

\paragraph{Weight parameters in $conv2D$ layers}
As shown in Figure \ref{fig:bitFlipErrorNotPruned}, faults injected in weight parameters of $conv2D$ layers located in the encoder branch ($p$ sets below 40 in Figure \ref{fig:unetModelNotPruned}) have more impact on the output than when injected in the decoder branch ($p$ sets above 48).
This discrepancy is due to the presence of skip-connections between encoder and decoder branches since faults injected in the first layers will directly propagate to the outputs through skip connections.
Moreover, longer data paths will be also involved in the propagation of the errors through encoder/decoder branches, including BN layers which, as mentioned above, increase the range of the signals with the risk of overflows.

In the worst case, if a bit-flip occurs in bit 30th of a positive weight, when multiplied by a positive input, it produces a value close to the maximum representable number (otherwise it would be absorbed by the ReLU and 0 would be propagated, which is less problematic).
Thus, most likely, the value would saturate to infinity when passing through a BN layer.
This observation aligns with the fact that the weights of the first 5 convolutions (Figures \ref{fig:calibrationUnpruned_0_3} and \ref{fig:calibrationUnpruned_4_7}) are significantly larger than those of the rest of convolutions in the encoder.
Hence, it is more likely for a bit-flip in bit 30 to cause an output mismatch when it is produced in these initial layers, as can be verified in Figure \ref{fig:bitFlipErrorNotPruned}.

\begin{table}[h!]
\caption{Percentage of positive bias of the most sensitive $conv2D$ layers in the unpruned model.
Note: the tables are read in a U way (from top to bottom in the encoder layers and from bottom to top in the decoder layers).}
\label{tab:biasConv2D_unpruned}
\centering
\resizebox{8cm}{!}{
\begin{tabular}{|c|c|c|c|c|c|c|c|}
\hline
\textbf{Name} & \textbf{Pos.} & \textbf{Total} & \textbf{\%} & \textbf{Name} & \textbf{Pos.} & \textbf{Total} & \textbf{\%} \\ \hline
\textbf{conv2D}    & 15 & 32  & 46.88 & \textbf{conv2D\_21}  & 30  & 32  & 93.75 \\ \hline
\textbf{conv2D\_1} & 12 & 32  & 37.50 & \textbf{conv2D\_20}  & 11  & 32  & 34.38 \\ \hline
\textbf{conv2D\_2} & 19 & 64  & 29.69 & \textbf{conv2D\_19} & 31  & 64  & 48.44 \\ \hline
\textbf{conv2D\_3} & 6  & 64  & 9.38  & \textbf{conv2D\_18} & 10  & 64  & 15.63 \\ \hline
\textbf{conv2D\_4} & 18 & 128 & 14.06 & \textbf{conv2D\_17} & 34  & 128 & 26.56 \\ \hline
\textbf{conv2D\_5} & 16 & 128 & 12.50 & \textbf{conv2D\_16} & 20  & 128 & 15.63 \\ \hline
\textbf{conv2D\_6} & 23 & 256 & 8.98  & \textbf{conv2D\_15} & 174 & 256 & 67.97 \\ \hline
\textbf{conv2D\_7} & 55 & 256 & 21.48 & \textbf{conv2D\_14} & 109 & 256 & 42.58 \\ \hline
\end{tabular}}
\end{table}

\paragraph{Bias parameters in $conv2D$ layers}
According to obtained statistics (Figure \ref{fig:bitFlipErrorNotPruned}), biases of $conv2D$ layers in the decoder branch appear to be more sensitive than biases in the encoder branch.
The explanation of this is that, for a bit-flip in position 30 of a bias in a $conv2D$ layer to significantly alter the output, it must be a positive valued bias.
Errors in negative biases are absorbed by ReLU layers, since BN layers between $conv2D$ and the ReLU do not change the sign of that value, given that the multiplicative constants are always positive.
$conv2D$ bias values share the same characteristics, i.e. positive and smaller than unity, as case 3 of the $\gamma$ parameters described in Section \ref{sec:gammaUnpruned}.
Therefore, when passing through BN layers, its value will increase to infinity, causing errors in the outputs.

\begin{figure}[h!]
\centering
\includegraphics[width=7.75cm]{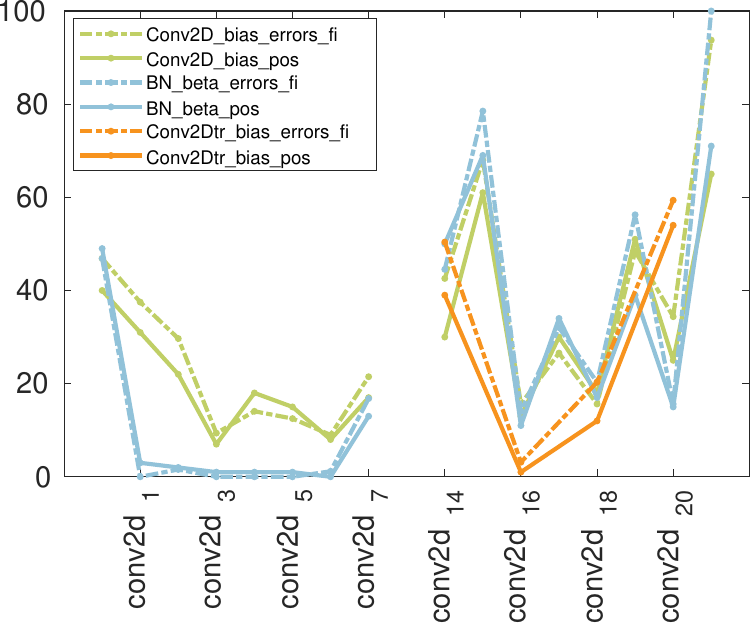}
\caption{Bit-flip error in the 30th bit (dashed) and ratio of positive parameters (solid) by parameter index: $conv2D_{b}$ (green), $BN_{beta}$ (blue) and $conv2Dtr_{b}$ (orange) in the unpruned model.}
\label{fig:interpretabilityPositiveBiasNotPruned}
\end{figure}

We can analyse the influence of bias parameters in more detail by inspecting the presence of positive biases in the primary $conv2D$ layers (deeper layers closer to the base of the model are analysed in a separate paragraph later on).
As shown in Table \ref{tab:biasConv2D_unpruned}, the percentage of positive biases in the decoder branch is much higher than in the encoder branch, which is consistent with our analysis and the obtained statistical observations.
In fact, if the correlation between the presence of positive bias and produces error rates is examined, parallel trends are observed (green lines in Figure \ref{fig:interpretabilityPositiveBiasNotPruned}).

\begin{table}[h!]
\caption{Percentage of positive betas of the most sensitive BN layers in the unpruned model.
Note: the tables are read in a U way (from top to bottom in the encoder layers and from bottom to top in the decoder layers).}
\label{tab:betaBN_unpruned}
\centering
\resizebox{7.5cm}{!}{
\begin{tabular}{|c|c|c|c|c|c|c|c|}
\hline
\textbf{Name} & \textbf{Pos.} & \textbf{Total} & \textbf{\%} & \textbf{Name} & \textbf{Pos.} & \textbf{Total} & \textbf{\%} \\ \hline
\textbf{bn}    & 15 & 32  & 46.88 & \textbf{bn\_21} & 32  & 32  & 100 \\ \hline
\textbf{bn\_1} & 0  & 32  & 0     & \textbf{bn\_20} & 5   & 32  & 15.63  \\ \hline
\textbf{bn\_2} & 1  & 64  & 1.56  & \textbf{bn\_19} & 36  & 64  & 56.25  \\ \hline
\textbf{bn\_3} & 0  & 64  & 0     & \textbf{bn\_18} & 13  & 64  & 20.31  \\ \hline
\textbf{bn\_4} & 0  & 128 & 0     & \textbf{bn\_17} & 41  & 128 & 32.03  \\ \hline
\textbf{bn\_5} & 0  & 128 & 0     & \textbf{bn\_16} & 18  & 128 & 14.06  \\ \hline
\textbf{bn\_6} & 3  & 256 & 1.17  & \textbf{bn\_15} & 201 & 256 & 78.52  \\ \hline
\textbf{bn\_7} & 43 & 256 & 16.80 & \textbf{bn\_14} & 114 & 256 & 44.53  \\ \hline
\end{tabular}}
\end{table}

\paragraph{Beta parameters in batch normalization layers}
Similarly to the analysis performed for biases in $conv2D$ layers, comparable error sensitivity for positive $\beta$ parameter values is observed.
As shown in Figure \ref{fig:interpretabilityPositiveBiasNotPruned} (blue lines), there is also a high degree of correlation between the error rate and the percentage of positive $\beta$ in each BN layer (Table \ref{tab:betaBN_unpruned}).

\paragraph{Bias parameters in $conv2Dtr$ layers}
Finally, in Table \ref{tab:biasConv2Dtr}, the percentage of biases that are positive for the primary $conv2D\_tr$ layers is grouped to see if the biases of the $conv2D\_tr$ also follow the above dynamics.
Plotted in orange in Figure \ref{fig:interpretabilityPositiveBiasNotPruned}, it is observed how, as in previous cases, both lines follow the same trend.

At this point, it is important to note that only bit-flips in the bit 30 of parameters have been considered so far.
Indeed, bit-flip on the rest of the bits produce many fewer errors, as seen in Figure \ref{fig:bitFlipErrorNotPruned}.
However, there are particular bits in specific parameters, such as bit 26 in the bias of the first $conv2D$ layer, associated with a significant error rate.
Explanation on this issue will be provided later on this paper.
Finally, it has to be remarked that the correlation between lines depicted in Figure \ref{fig:interpretabilityPositiveBiasNotPruned} is not perfect.
The reason is that faulty bits in negative biases can also result in a prediction error since they may turn positive activations into 0 values, and there are also corrupted positive biases that can be absorbed by a ReLU layer if they are multiplied by a null or negative weight (the error should be smaller then).

\paragraph{Deep layers: the base of the U-Net}\label{sec:baseUnpruned}
From the calibration and the FI analyses, it can be concluded that the deepest layers of the network ($p33$-$p58$ sets in Figure \ref{fig:unetModelNotPruned}) practically do not contribute to the inference process.
Nevertheless, training of shallower models (encoder depths 2, 3, and 4) with fewer convolution filters (8, 16, 32 in the first convolution) produced considerably worse results.
In fact, compressing trained large sparse models has been shown to be more effective than training smaller dense models \cite{zhu2017prune}.

\begin{figure}[h!]
\centering
\includegraphics[width=7.75cm]{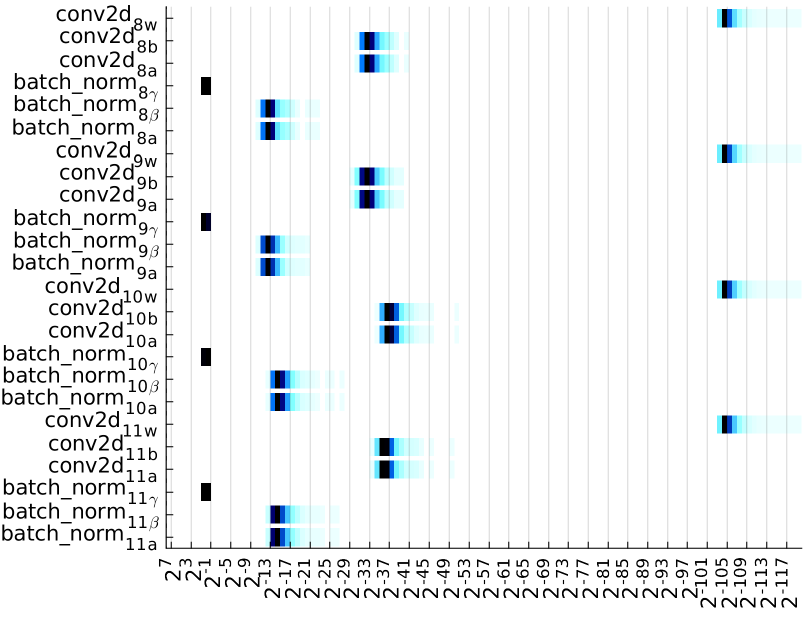}
\caption{Graphic representation of the unpruned model's calibration study (layers $conv2D\_8$ to $conv2D\_11$).}
\label{fig:calibrationUnpruned_8_11}
\end{figure}

\begin{figure}[h!]
\centering
\includegraphics[width=7.75cm]{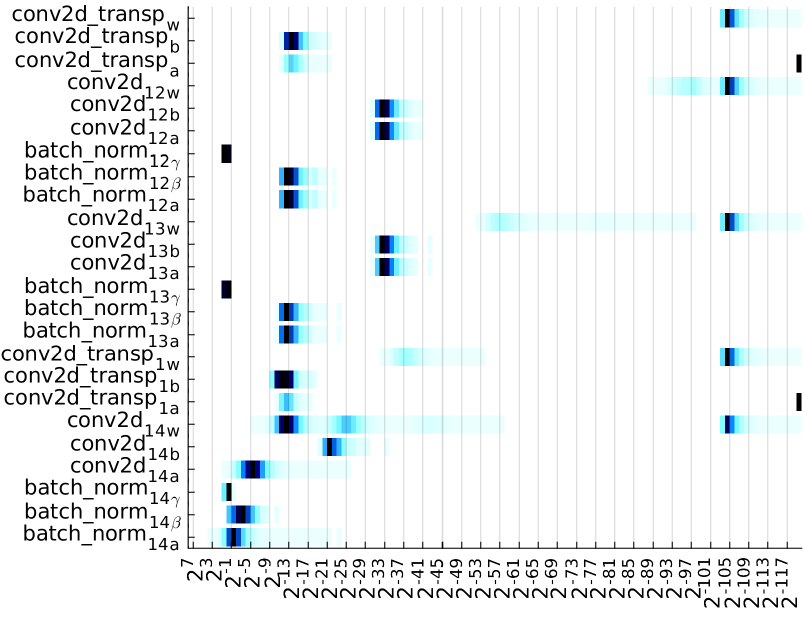}
\caption{Graphic representation of the unpruned model's calibration study (layers $conv2D\_12$ to $conv2D\_14$).}
\label{fig:calibrationUnpruned_12_14}
\end{figure}

Convolution weights in deep layers are so small that only the biases (Figures \ref{fig:calibrationUnpruned_8_11} and \ref{fig:calibrationUnpruned_12_14}) contribute to the activations.
Theoretically, even a bit-flip in the most significant bit of the biases is quickly absorbed by the network and should not influence the outputs.
However, it can be seen in Figure \ref{fig:bitFlipErrorNotPruned} that error statistics for FIs in deep layers ($p41$-$p50$ sets) are not negligible.
This is because the previous reasoning does not take into account events where a parameter value becomes $NaN$.
That is what precisely occurs when the bit 30 of one of the $\gamma$ parameters whose value is very close to 1 flips up.

\subsubsection{Analysis of robustness of the pruned model}
\begin{figure}[h!]
\centering
\includegraphics[width=7.75cm]{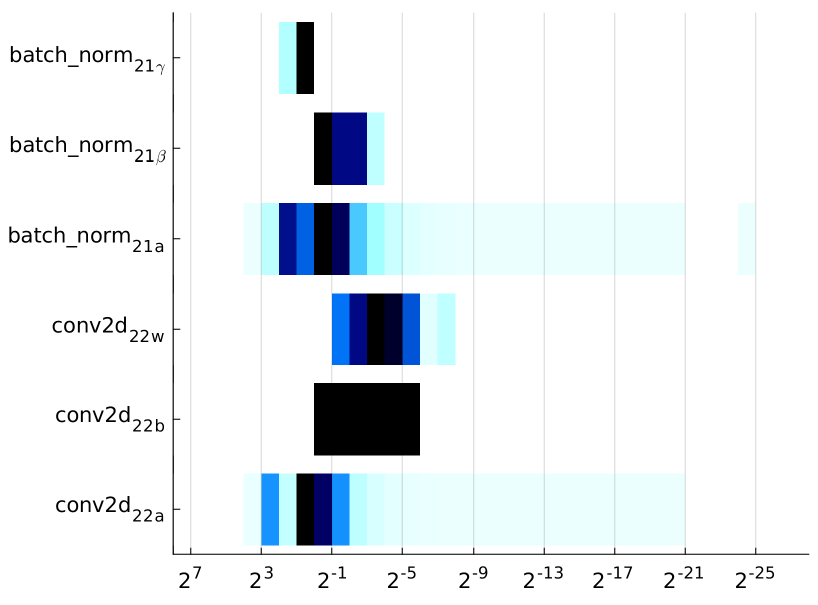}
\caption{Graphic representation of the pruned model's calibration studies (layers $conv2D\_21$ to $conv2D\_22$).}
\label{fig:calibrationPruned_21_22}
\end{figure}

\paragraph{Output $conv2D$ layer}\label{sec:conv22_pruned}
In the pruned model, the structure of $conv2D\_22$ is very similar to that of the unpruned model, as it also consists of 6 filters of size $1 * 1 * 32$ and 6 biases.
As shown in Figure \ref{fig:calibrationPruned_21_22}, there is a slight difference in the range of the activations, which is now $[-8.3144, 15.1673]$.
However, since the values of the bias parameters are essentially identical to those of the unpruned model ($-0.8495$, $0.3150$, $-0.02991$, $0.0373$, $-0.1736$, $0.1066$), the conclusions that can be drawn are quite the same.

\newpage

Applying Equation \ref{equ:ProbabilityBeforehand} to this case, the following value is obtained:

\begin{equation*}
\begin{split}
    \% error_{30} & = \frac{1}{6} (0 + 55.66 + 4.35 + 72.93 + 7.37 + 83.13)
    \\ & = 37.24
\end{split}
\end{equation*}

The theoretical value (37.24\%) is not far from that obtained experimentally and depicted in Figure \ref{fig:bitFlipErrorPruned} (32\%), being the difference a consequence of the statistical nature of the FI campaign.

\begin{figure}[h!]
\centering
\includegraphics[width=8.25cm]{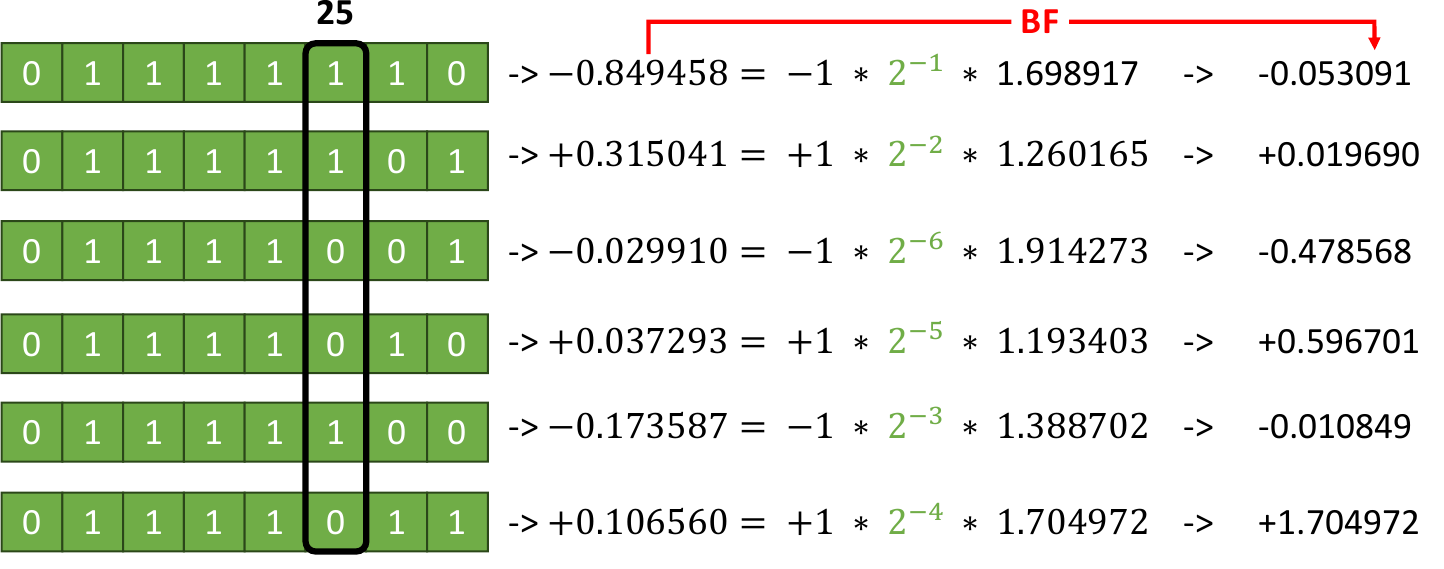}
\caption{Exponent part of the 32-bit floating-point representation of the biases of the last $conv2D$ of the pruned model.}
\label{fig:biasConv2D_22exponent_pruned}
\end{figure}

Let us now focus on lower position bits.
In Figure \ref{fig:bitFlipErrorPruned} it can also be seen that faults in bits 29, 28, 27, and 26 do not generate errors in the output, but changes in bit 25 do.
If we analyse the binary exponent of the 6 biases under consideration (see Figure \ref{fig:biasConv2D_22exponent_pruned}), it is observed that all of them contain a '1' in bits $[26-29]$.
A bit-flip in those positions will produce a reduction of the represented magnitude, and hence, the result of the bias addition will decrease too.
However, since the biases are small compared to the range of activations (see Figure \ref{fig:calibrationPruned_21_22}), it is reasonable to think that this reduction will not produce a meaningful change in the output map.
Same circumstance will happen if a bit-flip from 1 to 0 occurs in bit 25.
But, focusing on the last bias of Figure \ref{fig:biasConv2D_22exponent_pruned}, a bit-flip from 0 to 1 in bit 25 causes the partial exponent to fill with 1s.
Thus, the value increases from +0.1 to +1.7, which is a mayor perturbation that will probably produce a prediction error.

\paragraph{Gamma parameter in batch normalization layers}\label{sec:gammaPruned}
In the pruned model, it is also found that the $\gamma$ of the BN layers is the most sensitive parameter, and the same explanation holds.
However, in the seventh BN layer ($p27$ set in Figure \ref{fig:unetModelNotPruned}), there is a clear decrease in the number of errors associated with bit-flips in bit 30 and an increase in the number of errors associated with bit-flips in bits $[26-29]$.
The reason is that some of the $\gamma$ values in this layer, unlike in the rest of the BN layers, are greater than 2 but smaller than 3.
This means that the exponents in the binary representations change drastically from 01111111 to 1000000.
Then, a change in bit 30 of those $\gamma$s will no longer cause a value to be a $NaN$; instead, it will be a very small number that is less likely to produce a critical error.
In turn, a bit-flip in bits $[26-29]$ will now result in a considerable increase in magnitude, so the error rates for those bit positions grow compared to the layers with $\gamma$ values smaller than 2.

\begin{figure}[h!]
\centering
\includegraphics[width=7.75cm]{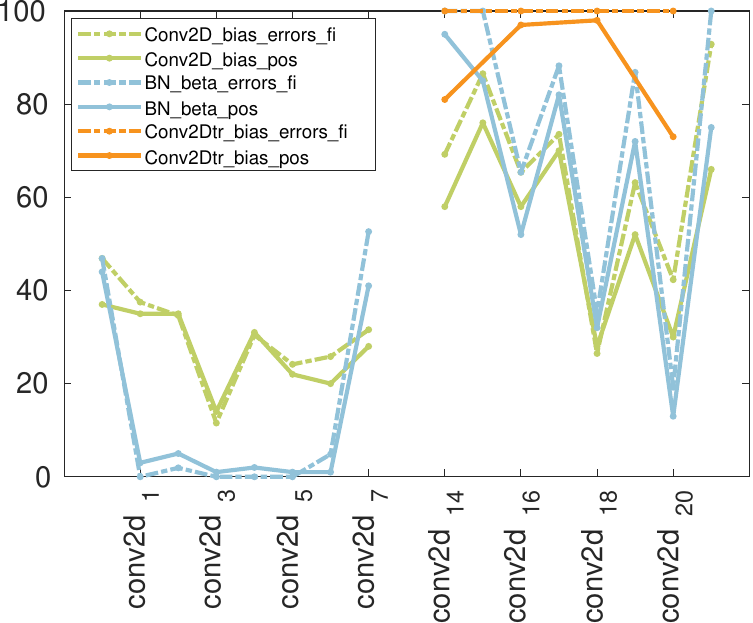}
\caption{Bit-flip error in the 30th bit (dashed) and ratio of positive parameters (solid) by parameter index: $conv2D_{b}$ (green), $BN_{beta}$ (blue) and $conv2Dtr_{b}$ (orange) in the pruned model.}
\label{fig:interpretabilityPositiveBiasPruned}
\end{figure}

\paragraph{Bias parameters in $conv2D$ layers}
As observed in the unpruned model, flips in bit 30 of the biases of $conv2D$ layers located in the encoder branch do not produce as many errors as in the decoder part.
Table \ref{tab:biasConv2D_pruned} contains the total number of positive biases compared to the total number of biases in those layers.
Again, as shown in Figure \ref{fig:interpretabilityPositiveBiasPruned} (green lines), there is a high correlation between the sign of the biases and the error rate.

\begin{figure}[h!]
\centering
\includegraphics[width=7.75cm]{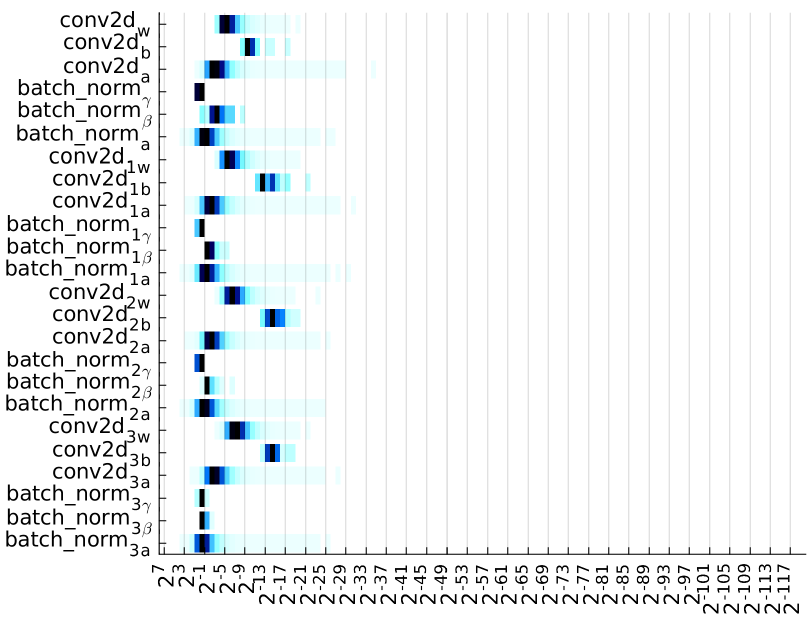}
\caption{Graphic representation of the pruned model's calibration study (layers $conv2D$ to $conv2D\_3$).}
\label{fig:calibrationPruned_0_3}
\end{figure}

As previously mentioned, Figure \ref{fig:bitFlipErrorPruned} exhibits a peak for the bias parameter of the first $conv2D$ layer in bit 26.
Furthering the discussion from the paragraph on the last $conv2D$, here 31 out of the 32 biases have an absolute value greater than $2^{-15}$ (all 0s in the exponent except bits 29, 28 and 27).
Hence, altering bit 26 produces the most significant increase in absolute value.
While this increment may seem insignificant, considering the first layer's activations (refer to Figure \ref{fig:calibrationPruned_0_3}) which are the lowest across the network ($[-2.6148, 2.1645]$), even small perturbations can induce notable changes.

\begin{table}[h!]
\caption{Percentage of positive bias of the most sensitive $conv2D$ layers in the pruned model.
Note: the tables are read in a U way (from top to bottom in the encoder layers and from bottom to top in the decoder layers).}
\label{tab:biasConv2D_pruned}
\centering
\resizebox{8cm}{!}{
\begin{tabular}{|c|c|c|c|c|c|c|c|}
\hline
\textbf{Name} & \textbf{Pos.} & \textbf{Total} & \textbf{\%} & \textbf{Name} & \textbf{Pos.} & \textbf{Total} & \textbf{\%} \\ \hline
\textbf{conv2D}    & 15 & 32  & 46.88 & \textbf{conv2D\_21} & 26 & 28 & 92.86 \\ \hline
\textbf{conv2D\_1} & 12 & 32  & 37.50 & \textbf{conv2D\_20} & 11 & 26 & 42.31 \\ \hline
\textbf{conv2D\_2} & 18 & 52  & 34.62 & \textbf{conv2D\_19} & 24 & 38 & 63.16 \\ \hline
\textbf{conv2D\_3} &  6 & 52  & 11.54 & \textbf{conv2D\_18} &  9 & 34 & 26.47 \\ \hline
\textbf{conv2D\_4} & 14 & 46  & 30.43 & \textbf{conv2D\_17} & 25 & 34 & 73.53 \\ \hline
\textbf{conv2D\_5} & 14 & 58  & 24.14 & \textbf{conv2D\_16} & 17 & 26 & 65.38 \\ \hline
\textbf{conv2D\_6} & 16 & 62  & 25.81 & \textbf{conv2D\_15} & 45 & 52 & 86.54 \\ \hline
\textbf{conv2D\_7} & 24 & 76  & 31.58 & \textbf{conv2D\_14} & 36 & 52 & 69.23 \\ \hline
\end{tabular}}
\end{table}

\begin{figure}[h!]
\centering
\includegraphics[width=7.75cm]{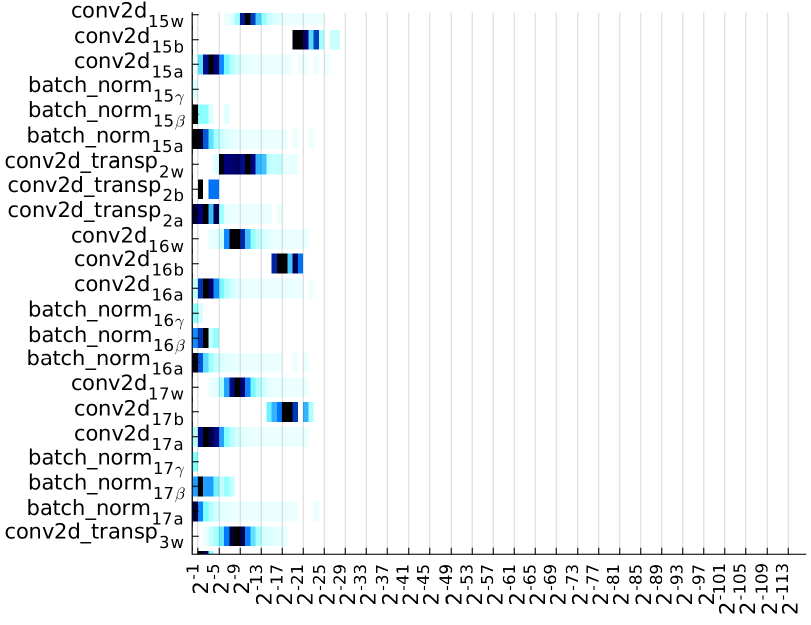}
\caption{Graphic representation of the pruned model's calibration study (layers $conv2D\_15$ to $conv2D\_17$).}
\label{fig:calibrationPruned_15_17}
\end{figure}

There is also a noticeable peak in the error rates of $conv2D\_16_b$ for changes in bit 27 ($p72$ set in Figure \ref{fig:unetModelNotPruned}).
Among the 26 biases, 4 of them have 6 of the 7 LSBs of the exponent set to 1 (all except bit 27), placing their absolute value in $[1.5259e^{-5}, 2*1.5259e^{-5}]$ range.
There are other 2 biases that are slightly outside this range.
Simultaneously, the activations of this layer are not very large (refer to Figure \ref{fig:calibrationPruned_15_17}).
In consequence, a bit-flip in bit 27 notably increases the value of those 4 biases to values larger than 1 but smaller than 2 (with the other 2 biases reaching a value close to 1).

According to this, we could expect small error rate peaks in certain bit positions for every layer in Figure \ref{fig:bitFlipErrorPruned}, given that all bias parameter sets contain some values with binary representations that are one bit away from completing the 7 least significant bits of the exponent (partial exponent).
However, this is not the case for every layer.
The reason is that layers with higher activation ranges are more robust against such errors.

An illustrative example is provided by the biases in $conv2D\_20$ ($p92$ set in Figure \ref{fig:unetModelNotPruned}), where 10 out of 26 biases show an absolute value in the range $[1.5259e^{-5}, 2*1.5259e^{-5})$.
There are other 2 biases that are close to this range.
In those cases, flipping bit 27 will cause a noticeable increase in the values, making them greater than one.
However, those bit-flips do not induce more errors than in $conv2D\_16$ because of the broader range of activations ($[-3.38, 4.93]$ for $conv2D\_16$ and $[-8.97, 6.94]$ for $conv2D\_20$).

\paragraph{Beta parameter in batch normalization layers}
The errors resulting from injecting a flipped bit 30 on the $\beta$ parameter in the batch normalization layers increase for the pruned model.
The percentage of $\beta$ parameters that are positive relative to the total is noted in Table \ref{tab:betaBN_pruned}.
As shown in Figure \ref{fig:interpretabilityPositiveBiasPruned} (blue line), the correlation between parameter soundness and the error rate is almost perfect.

\begin{table}[h!]
\caption{Percentage of positive betas of the most sensitive BN layers in the pruned model.
Note: the tables are read in a U way (from top to bottom in the encoder layers and from bottom to top in the decoder layers).}
\label{tab:betaBN_pruned}
\centering
\resizebox{7cm}{!}{
\begin{tabular}{|c|c|c|c|c|c|c|c|}
\hline
\textbf{Name} & \textbf{Pos.} & \textbf{Total} & \textbf{\%} & \textbf{Name} & \textbf{Pos.} & \textbf{Total} & \textbf{\%} \\ \hline
\textbf{bn}    & 15 & 32 & 46,88 & \textbf{bn\_21} & 28 & 28 & 100 \\ \hline
\textbf{bn\_1} & 0  & 32 & 0,00  & \textbf{bn\_20} &  5 & 26 & 19,23  \\ \hline
\textbf{bn\_2} & 1  & 52 & 1,92  & \textbf{bn\_19} & 33 & 38 & 86,84  \\ \hline
\textbf{bn\_3} & 0  & 52 & 0,00  & \textbf{bn\_18} & 12 & 34 & 35,29  \\ \hline
\textbf{bn\_4} & 0  & 46 & 0,00  & \textbf{bn\_17} & 30 & 34 & 88,24  \\ \hline
\textbf{bn\_5} & 0  & 58 & 0,00  & \textbf{bn\_16} & 17 & 26 & 65,38  \\ \hline
\textbf{bn\_6} & 3  & 62 & 4,84  & \textbf{bn\_15} & 52 & 52 & 100 \\ \hline
\textbf{bn\_7} & 40 & 76 & 52,63 & \textbf{bn\_14} & 52 & 52 & 100 \\ \hline
\end{tabular}}
\end{table}

\paragraph{Bias parameter in $conv2Dtr$ layers}
This parameter is one of the most pruned as Table \ref{tab:biasConv2Dtr} shows the difference in the number of parameters before and after applying pruning.
Visualizing the two lines plotted in orange in Figure \ref{fig:interpretabilityPositiveBiasPruned}, it is observed that this is the case for which the two curves are farther apart.
However, it should be noted that this is the oddest case, since all the biases are positive.
This would imply an error rate of 100\%, but we have previously discussed that very large positive biases can still be absorbed, reducing their adverse impact on the output.

\begin{table}[h!]
\caption{Percentage of positive bias of the most sensitive $conv2Dtr$ layers of the unpruned (left) and pruned (right) models.
Note: the table is read from bottom to top.}
\label{tab:biasConv2Dtr}
\parbox{.49\linewidth}{
\centering
\resizebox{4cm}{!}{
\begin{tabular}{|c|c|c|c|}
\hline
\textbf{Name} & \textbf{Pos.} & \textbf{Total} & \textbf{\%} \\ \hline
\textbf{conv2Dtr\_4} & 19  & 32  & 59.38 \\ \hline
\textbf{conv2Dtr\_3} & 13  & 64  & 20.31 \\ \hline
\textbf{conv2Dtr\_2} & 4   & 128 & 3.13  \\ \hline
\textbf{conv2Dtr\_1} & 129 & 256 & 50.39 \\ \hline
\end{tabular}}}
\parbox{.49\linewidth}{
\centering
\resizebox{3.85cm}{!}{
\begin{tabular}{|c|c|c|c|}
\hline
\textbf{Name} & \textbf{Pos.} & \textbf{Total} & \textbf{\%} \\ \hline
\textbf{conv2Dtr\_4} &  8  &  8  & 50  \\ \hline
\textbf{conv2Dtr\_3} & 10  & 10  & 100 \\ \hline
\textbf{conv2Dtr\_2} &  4  &  4  & 100 \\ \hline
\textbf{conv2Dtr\_1} &  2  &  2  & 100 \\ \hline
\end{tabular}}}
\end{table}

\begin{figure}[h!]
\centering
\includegraphics[width=8.35cm]{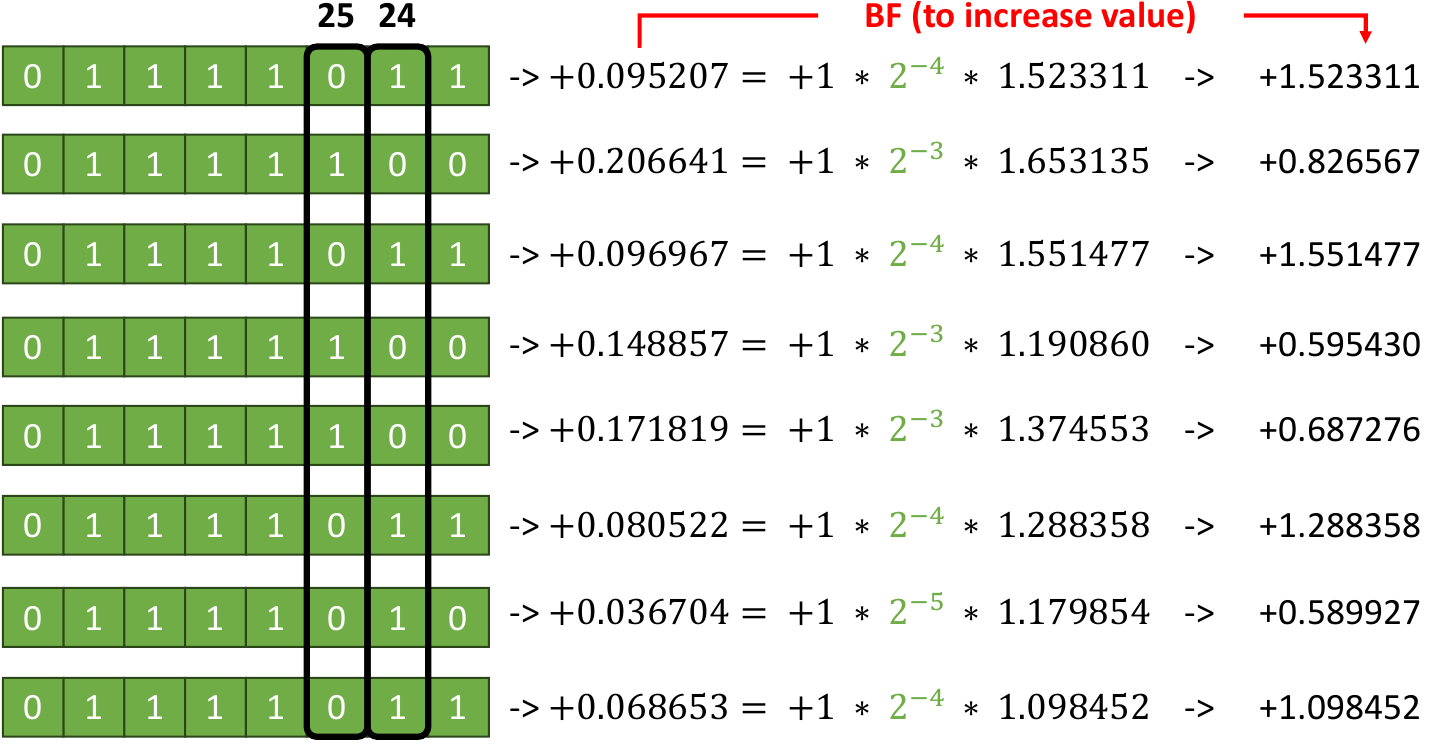}
\caption{Exponent part of the 32-bit floating-point representation of the $conv2Dtr\_4_b$, $p90$ set in Figure \ref{fig:unetModelNotPruned}.}
\label{fig:biasConv2Dtr_4exponent_pruned}
\end{figure}

Finally, $conv2Dtr\_4$ ($p90$ set in Figure \ref{fig:unetModelNotPruned}) is analysed in detail as it has a peak in bits 24 and 25 (Figure \ref{fig:bitFlipErrorPruned}).
Based on Figure \ref{fig:biasConv2Dtr_4exponent_pruned}, problems arise when the bit-flip increases the magnitude of a value.
Since all the biases of the parameter have bits in the range $[26-29]$ set to '1', a bit-flip in those bits will not increase their value, so an error is unlikely to happen.
However, in many cases, a bit-flip in bit 24, and especially in bit 25, causes bias values to increase to even above unity (when the least significant 7 bits of the exponent are set to '1').

\begin{figure}[h!]
\centering
\includegraphics[width=7.75cm]{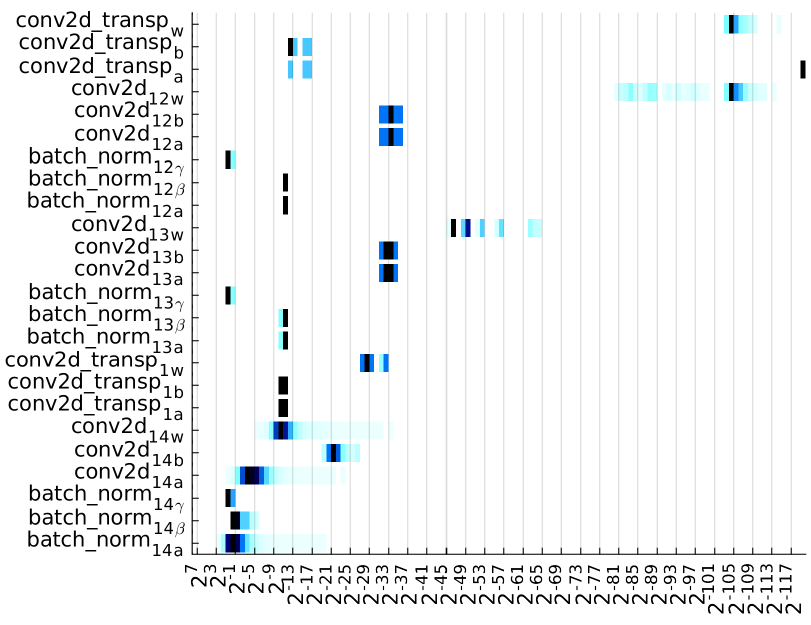}
\caption{Graphic representation of the pruned model's calibration study (layers $conv2D\_12$ to $conv2D\_14$).}
\label{fig:calibrationPruned_12_14}
\end{figure}

\paragraph{Deep layers: the base of the U-Net}
The base of the pruned U-Net is also quite insensitive to single bit-flips (Figure \ref{fig:bitFlipErrorPruned}) due to small parameter values.
However, it is observed that in the graph region located between the first two transposed $conv2D$ layers ($p51$-$p58$ sets in Figure \ref{fig:unetModelNotPruned}) there is an increase in the error rate.
This is not a surprise if we look at how the ranges of the weights in layer $conv2D\_13$ ($p55$ set in Figure \ref{fig:unetModelNotPruned}) have increased after the pruning and the subsequent fine-tuning process (there is a noticeable drift to the left when comparing Figure \ref{fig:calibrationUnpruned_12_14} and \ref{fig:calibrationPruned_12_14}).

\subsubsection{Sensitivity changes in pruned layers}
\begin{figure}[h!]
\centering
\includegraphics[width=7.75cm]{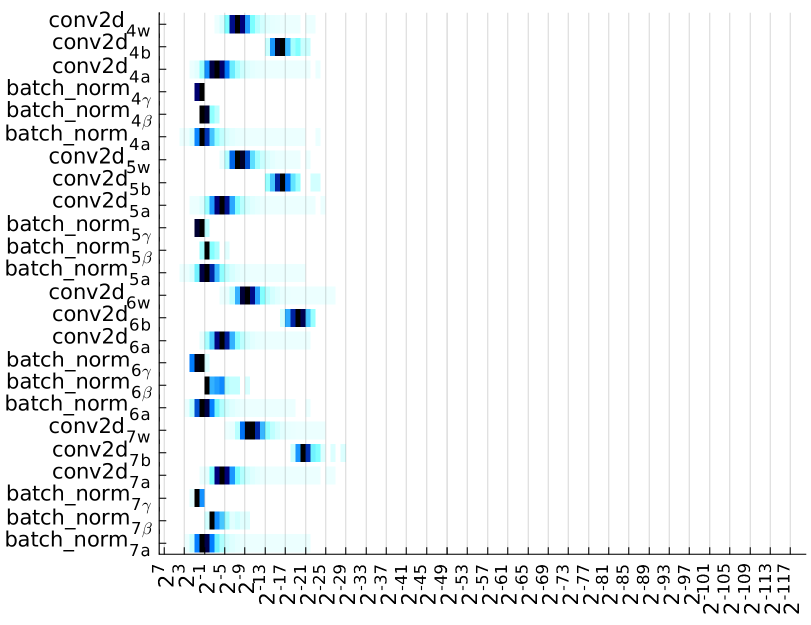}
\caption{Graphic representation of the pruned model's calibration study (layers $conv2D\_4$ to $conv2D\_7$).}
\label{fig:calibrationPruned_4_7}
\end{figure}

The most intensely pruned $conv2D$ layers of the encoder branch are $conv2D\_4$, $conv2D\_5$, $conv2D\_6$ and $conv2D\_7$ ($p17$-$p18$, $p21$-$p22$, $p25$-$p26$ and $p29$-$p30$ sets in Figures \ref{fig:unetModelNotPruned} respectively), which are precisely the layers with the largest increase in the measured error rate.
The pruning process did not result in a modification of the range of the weight values because, as expected, pruned filters contained very small weights.
As can be observed comparing parameter ranges depicted in Figures \ref{fig:calibrationUnpruned_4_7} and \ref{fig:calibrationPruned_4_7}, the original model contains many irrelevant weights that are not present in the pruned model.
In consequence, any SBU will be potentially more critical in the pruned model although, on the other hand, in most of implementations the probability of a SBU in smaller models is also lower.

\subsubsection{Batch normalization folding}
Since the $\gamma$ parameters of the BN layers have shown to be the most sensitive ones of the network upon SBUs, fusing the BN operation with the previous $conv2D$, technique known as BN folding, may produce an improvement in the robustness of the model.
Following the notation of Equation \ref{equ:BNoperation}, both operations can be fused into just one $conv2D$ operation as follows:

\begin{equation}
\begin{split}
BN(Conv2D(\textbf{x})) & = \gamma \frac{(\sum \textbf{w} \odot \textbf{x} + b) - \mu}{\sqrt{\sigma}} + \beta \\
                        & = \hat{Conv2D}(\textbf{x}) = \sum (\hat{\textbf{w}} \odot \textbf{x} + \hat{b})
\label{equ:BNoperationFolded}
\end{split}
\end{equation}

where $\hat{\textbf{w}} = \frac{\gamma}{\sqrt{\sigma}} \textbf{w}$ and $\hat{b} = \gamma \frac{b - \mu}{\sqrt{\sigma}} + \beta$.

\begin{figure}[h!]
\centering
\includegraphics[width=7.75cm]{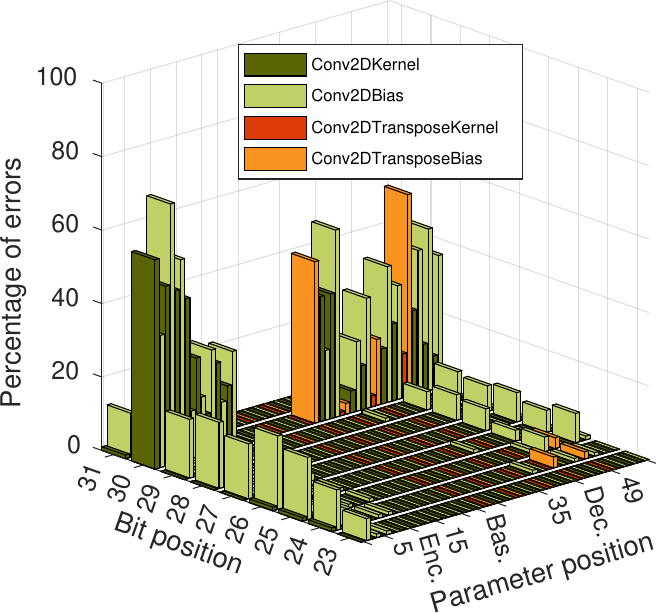}
\caption{Bit-flip error rate on the unpruned folded model.}
\label{fig:bitFlipErrorNotPrunedFolded}
\end{figure}

Although the original aim of this technique was to reduce the number of parameters to be stored and improve training performance, here we analyse if, indeed, it can help reduce prediction error rates produced by bit-flips.
Figure \ref{fig:bitFlipErrorNotPrunedFolded} shows the statistics of errors for the unpruned model after having folded all BN layers.
As can be seen, the layers associated with the highest error rates have been eliminated and, as a consequence of that, the base of the U-Net now does not produce any errors (there are neither $NaN$s nor $\pm$ $\infty$).
The error rates associated with the kernel of the $conv2D$ layers have decreased while the error rates in the kernel and bias parameters of the $conv2D\_tr$ layers remains unaltered.
However, the error rates to injected faults in the biases of $conv2D$ layers have increased.
To explain this increment, attention must be paid to the changes produced in the ranges of these values.

Table \ref{tab:valuesPerSegmentUnpruned} collects some bias range statistics for $conv2D$ layers, both for the unfolded and the folded models.
As expected, the folding operation has significantly increased the absolute value of the bias parameters.
Biases with absolute values between 1 and 2 in the folded model are a bit-flip away in bit 30 from producing a $NaN$ value.
Similarly, bit-flips in any of the remaining bits of the exponent of biases in the range above 2 will increase their value, augmenting the probability of producing output errors.
This analysis aligns with Figure \ref{fig:bitFlipErrorNotPrunedFolded}, showing that layers with the most values in the 'risky' range also have the highest error rates, and vice versa.

\begin{table}[h!]
\caption{Bias values per range before and after BN folding in some sensitive $conv2D$ layers of the unpruned model.}
\label{tab:valuesPerSegmentUnpruned}
\centering
\resizebox{8cm}{!}{
\begin{tabular}{c|ccccc|ccccc|}
\cline{2-11}
  & \multicolumn{5}{c|}{\textbf{Not Folded}} & \multicolumn{5}{c|}{\textbf{Folded}} \\ \hline
\multicolumn{1}{|c|}{\textbf{Range/conv2D}} & \multicolumn{1}{c|}{\textbf{}} & \multicolumn{1}{c|}{\textbf{\_3}} & \multicolumn{1}{c|}{\textbf{\_6}} & \multicolumn{1}{c|}{\textbf{\_19}} & \textbf{\_21} & \multicolumn{1}{c|}{\textbf{}} & \multicolumn{1}{c|}{\textbf{\_3}} & \multicolumn{1}{c|}{\textbf{\_6}} & \multicolumn{1}{c|}{\textbf{\_19}} & \textbf{\_21} \\ \hline
\multicolumn{1}{|c|}{\textbf{-$\infty$ < x $\leq$ -2}} & \multicolumn{1}{c|}{0} & \multicolumn{1}{c|}{0} & \multicolumn{1}{c|}{0} & \multicolumn{1}{c|}{0} & 0 & \multicolumn{1}{c|}{3} & \multicolumn{1}{c|}{0} & \multicolumn{1}{c|}{1} & \multicolumn{1}{c|}{0} & 0 \\ \hline
\multicolumn{1}{|c|}{\textbf{-2 < x $\leq$ -1}} & \multicolumn{1}{c|}{0} & \multicolumn{1}{c|}{0} & \multicolumn{1}{c|}{0} & \multicolumn{1}{c|}{0} & 0 & \multicolumn{1}{c|}{3} & \multicolumn{1}{c|}{1} & \multicolumn{1}{c|}{12} & \multicolumn{1}{c|}{6} & 0 \\ \hline
\multicolumn{1}{|c|}{\textbf{-1 < x $\leq$ 1}} & \multicolumn{1}{c|}{32} & \multicolumn{1}{c|}{64} & \multicolumn{1}{c|}{256} & \multicolumn{1}{c|}{64} & 32 & \multicolumn{1}{c|}{15} & \multicolumn{1}{c|}{63} & \multicolumn{1}{c|}{230} & \multicolumn{1}{c|}{28} & 26 \\ \hline
\multicolumn{1}{|c|}{\textbf{1 < x $\leq$ 2}} & \multicolumn{1}{c|}{0} & \multicolumn{1}{c|}{0} & \multicolumn{1}{c|}{0} & \multicolumn{1}{c|}{0} & 0 & \multicolumn{1}{c|}{5} & \multicolumn{1}{c|}{0} & \multicolumn{1}{c|}{7} & \multicolumn{1}{c|}{27} & 2 \\ \hline
\multicolumn{1}{|c|}{\textbf{2 < x $\leq$ +$\infty$}} & \multicolumn{1}{c|}{0} & \multicolumn{1}{c|}{0} & \multicolumn{1}{c|}{0} & \multicolumn{1}{c|}{0} & 0 & \multicolumn{1}{c|}{6} & \multicolumn{1}{c|}{0} & \multicolumn{1}{c|}{6} & \multicolumn{1}{c|}{3} & 4 \\ \hline
\end{tabular}}
\end{table}

\begin{figure}[h!]
\centering
\includegraphics[width=7.75cm]{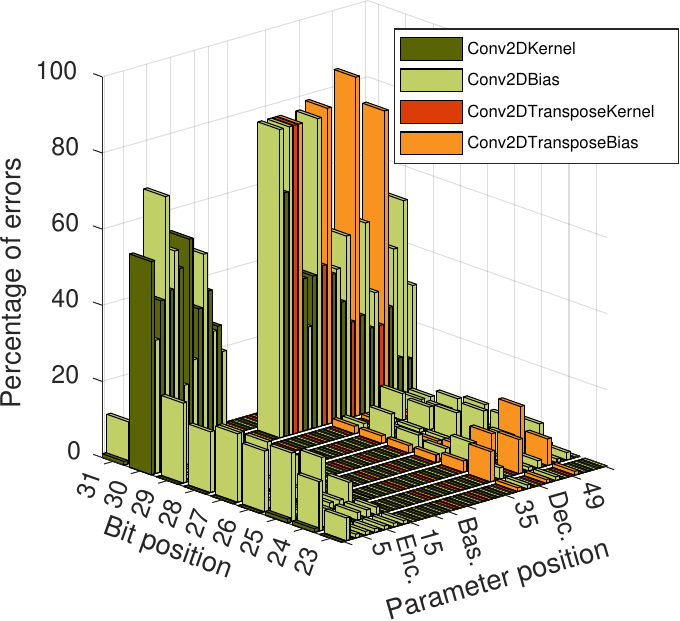}
\caption{Bit-flip error rate on the pruned folded model.}
\label{fig:bitFlipErrorPrunedFolded}
\end{figure}

The same significant differences observed in the unpruned folded model are also evident in the pruned folded model (compare statistics in Figure \ref{fig:bitFlipErrorPruned} and Figure \ref{fig:bitFlipErrorPrunedFolded}).
Firstly, the layers associated with the highest error rates have been removed.
Secondly, the base of the U-Net does not produce any errors either.
Thirdly, the error rates associated with the $conv2D$ kernel decreases and the error rates in $conv2Dtr$ kernel and bias parameters remains stable.
Lastly, the error rates associated with the bias of $conv2D$ layers have increased.
Once again, some bit positions $[24-29]$ which previously caused few or no errors, are now generating a significant number of them.
Looking to Table \ref{tab:valuesPerSegmentPruned}, we see the same range increase of bias values as in the unpruned model after BN folding, which leads to higher error rates in the corresponding layers.

In conclusion, although BN folding has positive effects on the general robustness of the model by eliminating layers with the highest error rates and the most dangerous situations from areas with low information, and reducing the number of parameters to protect, hence decreasing vulnerability, it also increases the probability of errors in some layers due to the increment of the absolute value of the $conv2D$ biases.

\begin{table}[h!]
\caption{Bias values per range before and after BN folding in some sensitive $conv2D$ layers of the pruned model.}
\label{tab:valuesPerSegmentPruned}
\centering
\resizebox{8cm}{!}{
\begin{tabular}{c|ccccc|ccccc|}
\cline{2-11}
  & \multicolumn{5}{c|}{\textbf{Not Folded}} & \multicolumn{5}{c|}{\textbf{Folded}} \\ \hline
\multicolumn{1}{|c|}{\textbf{Range/conv2D}} & \multicolumn{1}{c|}{\textbf{}} & \multicolumn{1}{c|}{\textbf{\_3}} & \multicolumn{1}{c|}{\textbf{\_6}} & \multicolumn{1}{c|}{\textbf{\_19}} & \textbf{\_21} & \multicolumn{1}{c|}{\textbf{}} & \multicolumn{1}{c|}{\textbf{\_3}} & \multicolumn{1}{c|}{\textbf{\_6}} & \multicolumn{1}{c|}{\textbf{\_19}} & \textbf{\_21} \\ \hline
\multicolumn{1}{|c|}{\textbf{-$\infty$ < x $\leq$ -2}} & \multicolumn{1}{c|}{0} & \multicolumn{1}{c|}{0} & \multicolumn{1}{c|}{0} & \multicolumn{1}{c|}{0} & 0 & \multicolumn{1}{c|}{3} & \multicolumn{1}{c|}{0} & \multicolumn{1}{c|}{0} & \multicolumn{1}{c|}{0} & 0 \\ \hline
\multicolumn{1}{|c|}{\textbf{-2 < x $\leq$ -1}} & \multicolumn{1}{c|}{0} & \multicolumn{1}{c|}{0} & \multicolumn{1}{c|}{0} & \multicolumn{1}{c|}{0} & 0 & \multicolumn{1}{c|}{3} & \multicolumn{1}{c|}{1} & \multicolumn{1}{c|}{5} & \multicolumn{1}{c|}{0} & 0 \\ \hline
\multicolumn{1}{|c|}{\textbf{-1 < x $\leq$ 1}} & \multicolumn{1}{c|}{32} & \multicolumn{1}{c|}{52} & \multicolumn{1}{c|}{62} & \multicolumn{1}{c|}{38} & 28 & \multicolumn{1}{c|}{12} & \multicolumn{1}{c|}{51} & \multicolumn{1}{c|}{34} & \multicolumn{1}{c|}{16} & 22 \\ \hline
\multicolumn{1}{|c|}{\textbf{1 < x $\leq$ 2}} & \multicolumn{1}{c|}{0} & \multicolumn{1}{c|}{0} & \multicolumn{1}{c|}{0} & \multicolumn{1}{c|}{0} & 0 & \multicolumn{1}{c|}{5} & \multicolumn{1}{c|}{0} & \multicolumn{1}{c|}{7} & \multicolumn{1}{c|}{19} & 3 \\ \hline
\multicolumn{1}{|c|}{\textbf{2 < x $\leq$ +$\infty$}} & \multicolumn{1}{c|}{0} & \multicolumn{1}{c|}{0} & \multicolumn{1}{c|}{0} & \multicolumn{1}{c|}{0} & 0 & \multicolumn{1}{c|}{6} & \multicolumn{1}{c|}{0} & \multicolumn{1}{c|}{6} & \multicolumn{1}{c|}{3} & 3 \\ \hline
\end{tabular}}
\end{table}

\subsection{8-bit integer quantized models (QNNs)}
The applied quantization scheme (Section \ref{sec:compressionTechniques}) involves converting weights to 8-bit integers and biases to 32-bit integers.
Additionally, BN layers have been folded into previous convolution layers to reduce the number of parameters and speed up computations.
According to \cite{jacob2018quantization}, the quantized version $\hat{r}$ of a real number $r$ is approximated by Equation \ref{eq:tfLiteQuantization}, where $S$ is a real positive scale factor, $q$ is an 8/32-bit integer value, and $Z$ is the zero-point integer value (0 for symmetric quantization).

\begin{equation}
    \centering
    \hat{r} \approx r = S (q - Z)
    \label{eq:tfLiteQuantization}
\end{equation}

Thus, the value $q_y$ resulting from $Y = Wx + b$ is:

\begin{equation}
    \centering
    q_y = \frac{S_w S_x}{S_y} (q_w q_x - Z_x q_w + q_b)
    \label{eq:tfLiteMAC}
\end{equation}

\begin{figure*}[t]
\centering
\includegraphics[width=11cm]{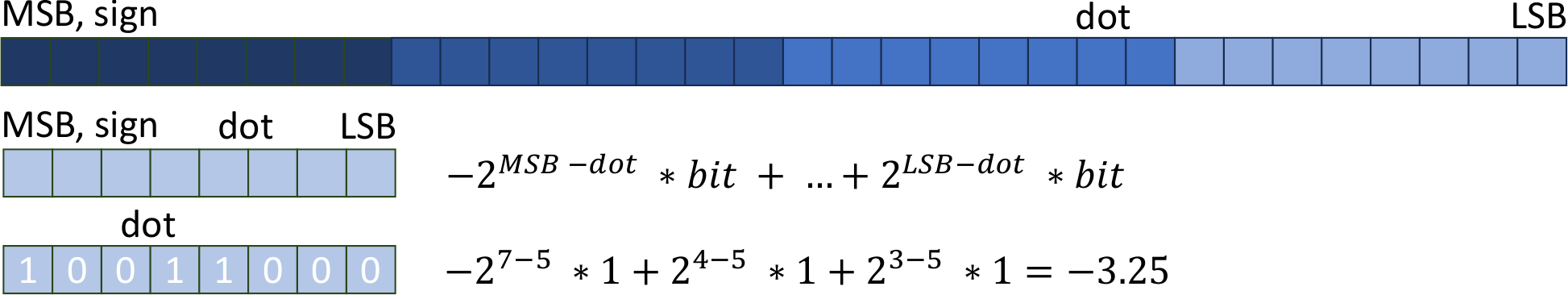}
\caption{Two's complement arithmetic for 8-bit and 32-bit integer/fixed-point representation.}
\label{fig:twoComplementArithmetic}
\end{figure*}

Due to the per-tensor and mainly symmetric quantization, the number of $q_i$ values that need to be stored is considerably greater than that of $Z_i$ and $S_i$ values.
Additionally, due to the quantization process itself, the S values have to meet certain restrictions \cite{jacob2018quantization} and are not of interest for the comparison being made.
Consequently, bit-flips were only injected on $q_w$ and $q_b$, which are represented in two's complement arithmetic, the typical computer representation for fixed-point (signed integers are just one example) binary values (Figure \ref{fig:twoComplementArithmetic}).

In this representation, a notable drawback arises from the double functionality of the sign bit: a bit-flip in the sign bit converts a very small negative number into a very large positive number (and vice versa) and a very small positive number into a very large negative number (and vice versa).
By contrast, two's complement does not define a specific representation for $NaN$ or infinity, seemingly alleviating the issues associated with these values.

\begin{figure}[h!]
\centering
\includegraphics[width=7.5cm]{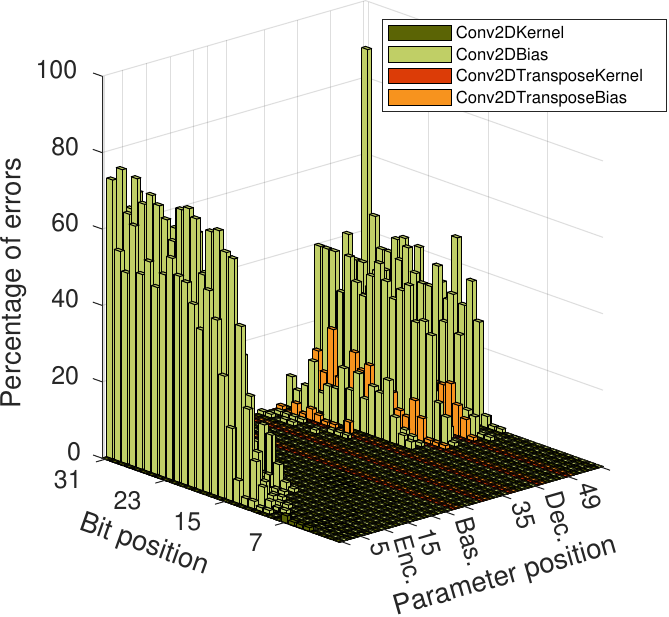}
\caption{Bit-flip error rate on the unpruned quantized model. Weights are quantized to 8 bits (bit 7 is the MSB). Biases are quantized to 32 bits and faults are injected across all bit positions due to disparities in their ranges.}
\label{fig:bitFlipErrorNotPrunedQuantized}
\end{figure}

\begin{figure}[h!]
\centering
\includegraphics[width=7.5cm]{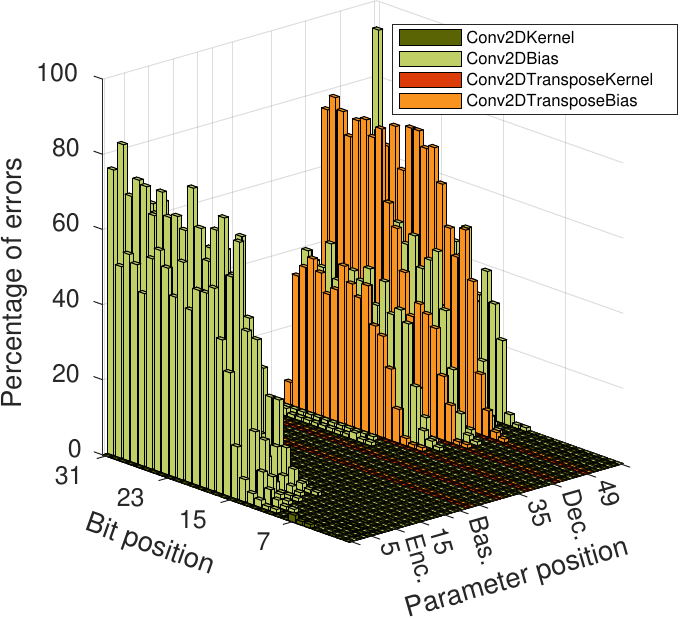}
\caption{Bit-flip error rate on the pruned quantized model. Weights are quantized to 8 bits (bit 7 is the MSB). Biases are quantized to 32 bits and faults are injected across all bit positions due to disparities in their ranges.}
\label{fig:bitFlipErrorPrunedQuantized}
\end{figure}

We repeated the statistical fault injection campaign on the QNN and obtained results are graphed in Figures \ref{fig:bitFlipErrorNotPrunedQuantized} and \ref{fig:bitFlipErrorPrunedQuantized} respectively.
Based on these results, the unpruned model shows, in principle, a superior robustness.

\subsubsection{Analysis of robustness of the unpruned model}

\paragraph{Relevance of weights compared to biases}
The injection of a fault in the convolution weights shows minimal impact on the output (see Figure \ref{fig:bitFlipErrorNotPrunedQuantized}).
Due to skip-connections in the model, the only critical scenario is the appearance of bit-flips in the first layer.
The significance of the perturbations in weights is negligible compared to the biases, which is a noteworthy observation.
This means that for an error protection technique based on redundancy, it could practically be limited to storing and checking the biases.

\begin{figure}[h!]
\centering
\includegraphics[width=7cm]{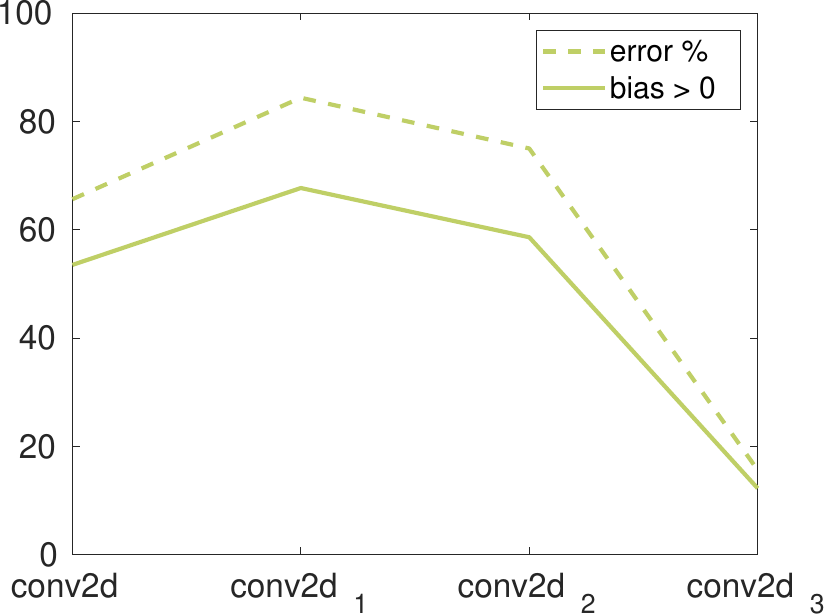}
\caption{Bit-flip error in $[17 - 30]$ bits (dashed) and ratio of positive bias (solid) in some $conv2D_{b}$ encoder layers of the quantized unpruned model.}
\label{fig:interpretabilityPositiveBiasNotPrunedQuantized}
\end{figure}

\paragraph{The biases} \label{sec:biasAnalysisUnpruned}
Contrary to what was observed for the unquantized models, encoder biases induce more errors than decoder biases.
Since bit-flips cannot result in $NaN$s or $\pm \infty$ in two's complement representation, the analysis differs.
For QNNs the layer position is more determinant, mainly due to the presence of skip connections and longer error propagation paths.
Indeed, the initial layers of the model are clearly the most sensitive ones.

\begin{figure}[h!]
\centering
\includegraphics[width=7.75cm]{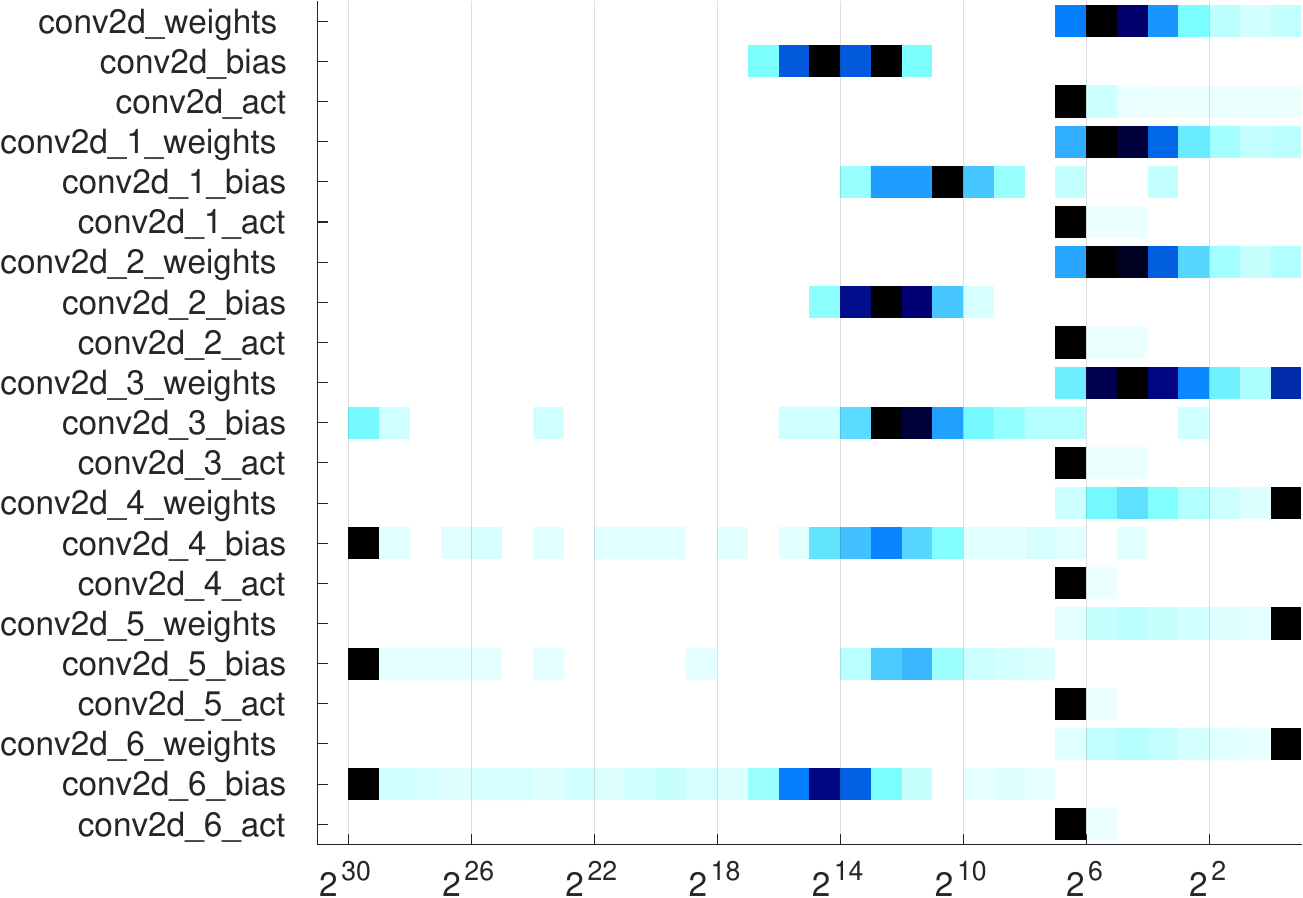}
\caption{Calibration study of the unpruned quantized model ($q_w$, $q_b$ and $q_a$) from layers $conv2D$ to $conv2D\_6$.}
\label{fig:calibrationUnprunedQuantized_0_6}
\end{figure}

However, analyzing the sign of the biases remains relevant to explain the bit-flip errors in the initial layers of the encoder.
For int-8 representation, perturbed parameters produce output errors that depend on the position of the flipped bits, being positions $[17 - 30]$ the most sensitive.
Once a bit-flip in one position produces an increase in the bias value high enough to produce an error at the output, changes in higher positions become inconsequential.
In order to get a general estimation of a single statistical error rate value generated by upset events in any of the considered bit positions, a linear weighting approximation based on bit significance has been applied.
Figure \ref{fig:interpretabilityPositiveBiasNotPrunedQuantized} reveals a practically identical trend between the calculated bit-flip error and the ratio of positive biases.

\begin{figure}[h!]
\centering
\includegraphics[width=7.75cm]{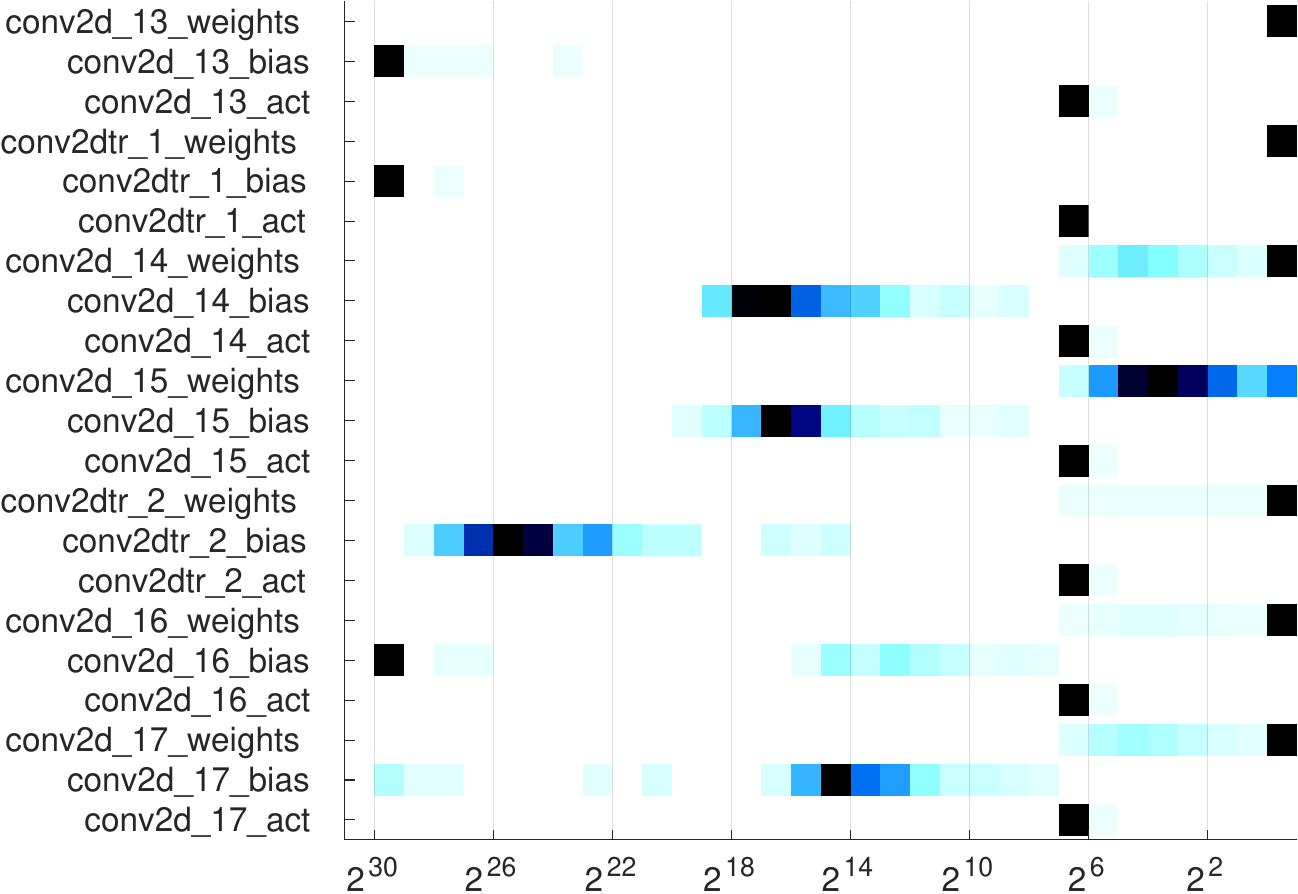}
\caption{Calibration study of the unpruned quantized model ($q_w$, $q_b$ and $q_a$) from layers $conv2D\_13$ to $conv2D\_17$.}
\label{fig:calibrationUnprunedQuantized_13_17}
\end{figure}

\begin{figure}[h!]
\centering
\includegraphics[width=7.75cm]{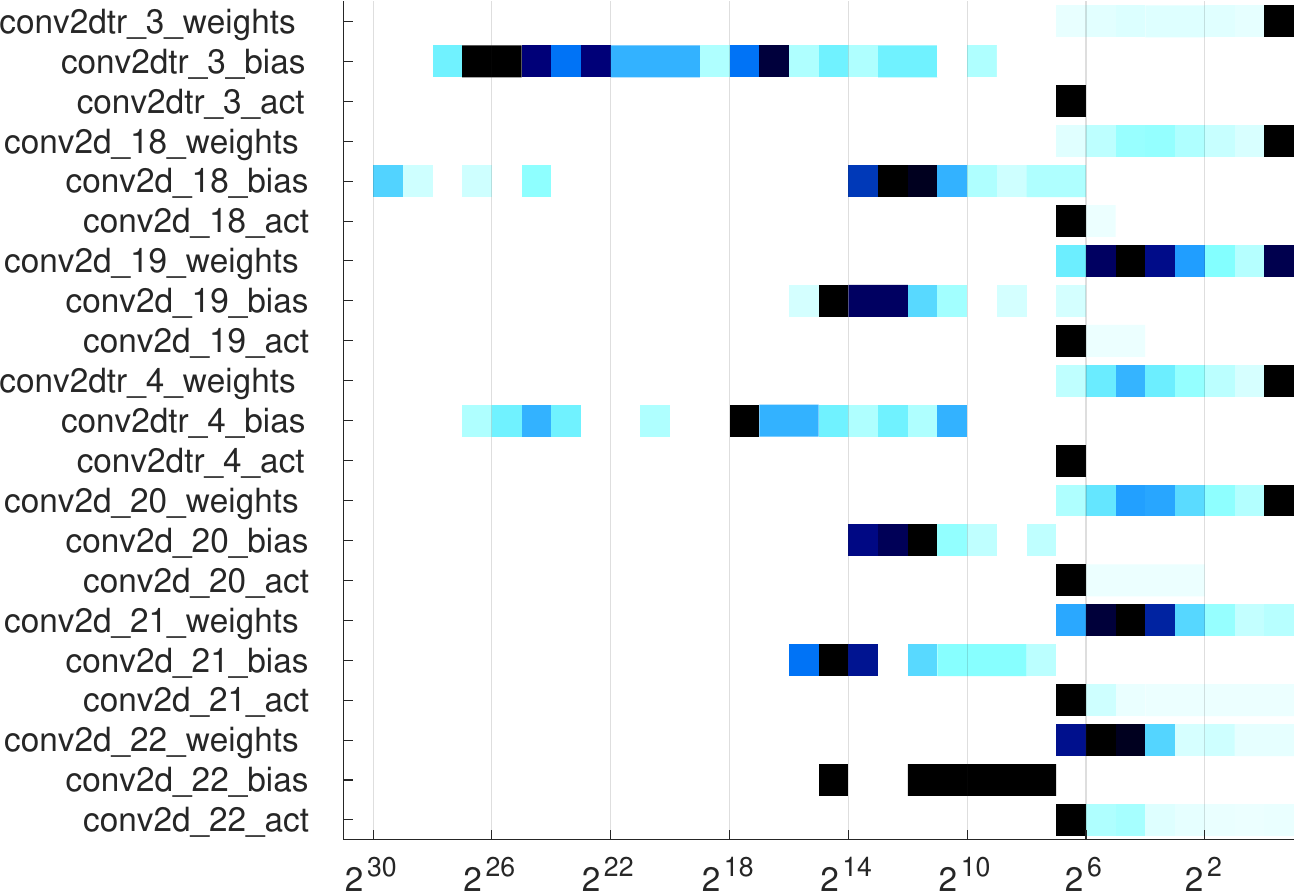}
\caption{Calibration study of the unpruned quantized model ($q_w$, $q_b$ and $q_a$) from layers $conv2\_18D$ to $conv2D\_22$.}
\label{fig:calibrationUnprunedQuantized_18_22}
\end{figure}

\begin{figure}[h!]
\centering
\includegraphics[width=7.75cm]{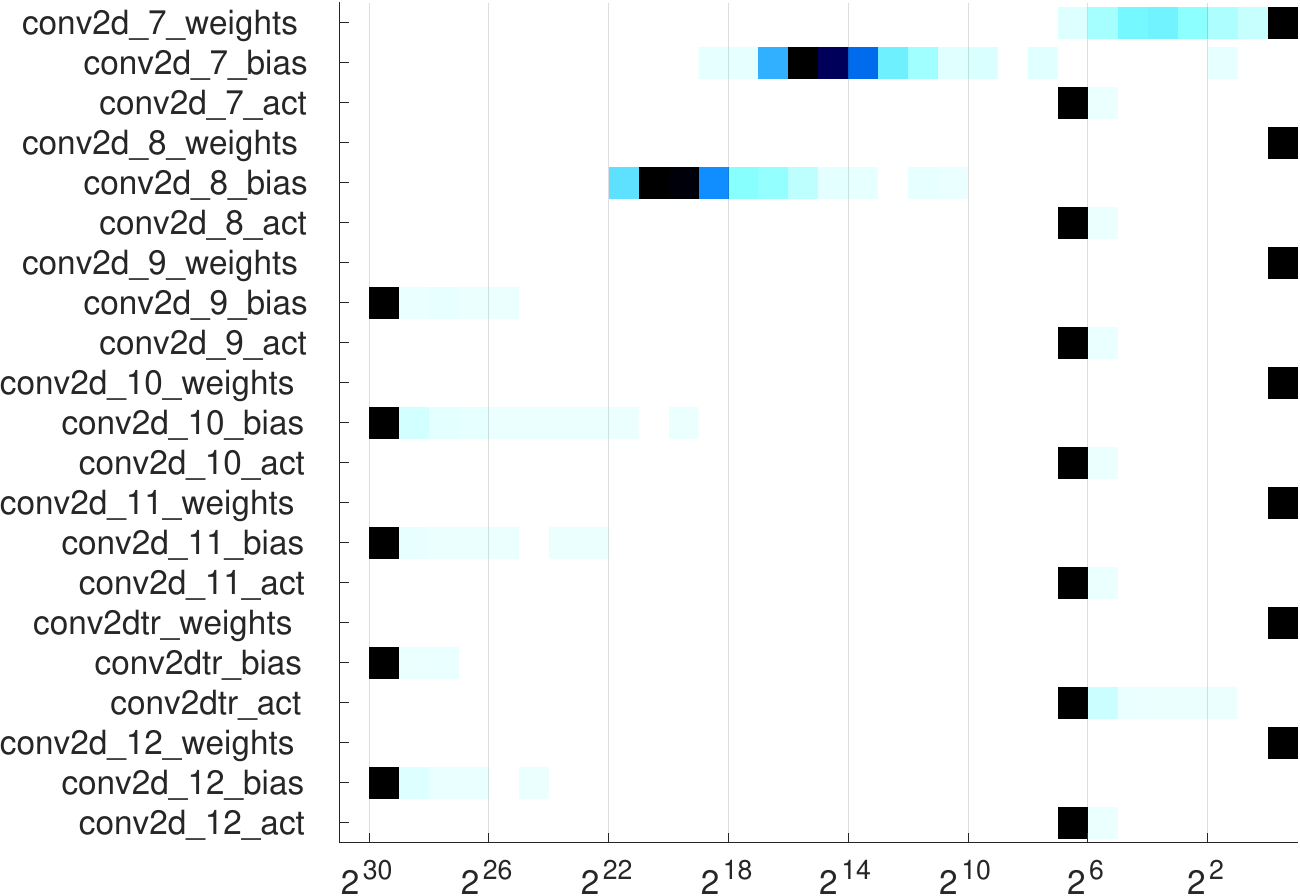}
\caption{Calibration study of the unpruned quantized model ($q_w$, $q_b$ and $q_a$) from layers $conv2D\_7$ to $conv2D\_12$.}
\label{fig:calibrationUnprunedQuantized_7_12}
\end{figure}

Finally, we analyse why bit-flips above bit 16, on average, result in significant errors, those below bit 13 cause very few errors, and below bit 9 induce none.
By referencing Table \ref{tab:biasParametersNotPrunedQuantized} and Figures \ref{fig:calibrationUnprunedQuantized_0_6}, \ref{fig:calibrationUnprunedQuantized_13_17}, \ref{fig:calibrationUnprunedQuantized_18_22} and \ref{fig:calibrationUnprunedQuantized_7_12}, it can be seen that the maximum bias values hovers around $2^{17}$.
Thus, bit-flips in higher positions lead to a magnified bias, resulting in a significant increase in the error rates.
Moreover, certain layers exhibit a requirement for 30/31 bits to represent all values.
To expound on this observation, we must differentiate between highly significant parameters (those in the first layers of the encoder and the last layers of the decoder) and those of low significance (found in the base zone).

If a parameter is placed in a highly relevant area of the model (e.g., $conv2D\_3$, $conv2D\_4$, $conv2D\_5$, $conv2D\_6$, $conv2D\_17$, $conv2Dtr\_3$, $conv2D\_18$, $conv2Dtr\_4$), the necessity for a high number of bits primarily stems from representing negative values (Table \ref{tab:biasParametersNotPrunedQuantized}).
However, there are relatively few instances of very negative values in these areas (as observed in Figures \ref{fig:calibrationUnprunedQuantized_0_6}, \ref{fig:calibrationUnprunedQuantized_13_17} and \ref{fig:calibrationUnprunedQuantized_18_22}).
Notably, channels with very negative biases typically have negligible relevance to the output, as the ReLU activation function effectively nullifies their contributions.

Conversely, in comparatively irrelevant areas of the model (e.g., $conv2D\_9$, $conv2D\_10$, $conv2D\_11$, $conv2Dtr$, $conv2D\_12$, $conv2D\_13$, $conv2Dtr\_1$), the necessity for a high number of bits arises from the need to represent both very positive and very negative values (Table \ref{tab:biasParametersNotPrunedQuantized}).
In these regions, there are numerous instances of very positive or very negative values (as observed in Figures \ref{fig:calibrationUnprunedQuantized_13_17} and \ref{fig:calibrationUnprunedQuantized_7_12}).
However, many of the channels either contain very negative biases (which are absorbed by ReLU activation) or very positive values that are subsequently multiplied by nearly zero weights (thus, contributing insignificantly).

\begin{table}[h!]
\caption{Necessary bits to represent all pos./neg. values of the $conv2D_b$ (green) and $conv2Dtr_b$ (orange) parameters in the quantized unpruned model.}
\label{tab:biasParametersNotPrunedQuantized}
\centering
\resizebox{7cm}{!}{
\begin{tabular}{|c|c|c|c|c|c|}
\hline
\textbf{Layer} & \textbf{Positive} & \textbf{Negative} & \textbf{Layer} & \textbf{Positive} & \textbf{Negative} \\ \hline
{\color[HTML]{009901} \textbf{0}} & 18 & 17 & {\color[HTML]{009901} \textbf{22}} & 13 & 16 \\ \hline
{\color[HTML]{009901} \textbf{1}} & 15 & 15 & {\color[HTML]{009901} \textbf{21}} & 17 & 13 \\ \hline
{\color[HTML]{009901} \textbf{2}} & 16 & 16 & {\color[HTML]{009901} \textbf{20}} & 15 & 15 \\ \hline
{\color[HTML]{009901} \textbf{3}} & 15 & 31 & {\color[HTML]{F8A102} \textbf{4}} & 19 & 28 \\ \hline
{\color[HTML]{009901} \textbf{4}} & 16 & 31 & {\color[HTML]{009901} \textbf{19}} & 16 & 17 \\ \hline
{\color[HTML]{009901} \textbf{5}} & 15 & 31 & {\color[HTML]{009901} \textbf{18}} & 15 & 31 \\ \hline
{\color[HTML]{009901} \textbf{6}} & 18 & 31 & {\color[HTML]{F8A102} \textbf{3}} & 20 & 29 \\ \hline
{\color[HTML]{009901} \textbf{7}} & 18 & 20 & {\color[HTML]{009901} \textbf{17}} & 17 & 31 \\ \hline
{\color[HTML]{009901} \textbf{8}} & 23 & 23 & {\color[HTML]{009901} \textbf{16}} & 16 & 31 \\ \hline
{\color[HTML]{009901} \textbf{9}} & 32 & 31 & {\color[HTML]{F8A102} \textbf{2}} & 18 & 30 \\ \hline
{\color[HTML]{009901} \textbf{10}} & 32 & 31 & {\color[HTML]{009901} \textbf{15}} & 19 & 21 \\ \hline
{\color[HTML]{009901} \textbf{11}} & 32 & 31 & {\color[HTML]{009901} \textbf{14}} & 19 & 20 \\ \hline
{\color[HTML]{F8A102} \textbf{0}} & 32 & 31 & {\color[HTML]{F8A102} \textbf{1}} & 32 & 31 \\ \hline
{\color[HTML]{009901} \textbf{12}} & 32 & 31 & {\color[HTML]{009901} \textbf{13}} & 32 & 31 \\ \hline
    \end{tabular}
  }
\end{table}%

In summary, although certain layers exhibit biases quantized to 30/31 bits, the high-order bits (typically 15/16 bits) are essentially redundant as they solely extend the sign.
These bits could potentially serve as a means of detecting and/or correcting sensitive bit positions.

\paragraph{Deep layers: the base of the U-Net}\label{sec:baseUNetNotPrunedQuantized}
Quantizing the model does not alter the significance of the base of the U-Net.
Even when a bit-flip occurs at the most sensitive bit (bit 31), the deepest layers ($p16$-$p32$ sets in Figure \ref{fig:bitFlipErrorNotPrunedQuantized}) remain unchanged.
This is because the convolution weights are very small, and only biases (as depicted in Figures \ref{fig:calibrationUnprunedQuantized_13_17} and \ref{fig:calibrationUnprunedQuantized_7_12}) contribute to the activations.

\paragraph{Output $conv2D$ layer} \label{sec:lastConvolutionNotPrunedQuantized}
The biases of the last convolution layer are $[-25983$, $2355$, $-187$, $300$, $-1494$, $923]$ (as illustrated in Figure \ref{fig:calibrationUnprunedQuantized_18_22}), consisting of 3 positive and 3 negative biases.
In the event of a bit-flip occurring in any of the bits within the $[17 - 30]$ range, the positive biases become more positive, while the negative biases become more negative.
Applying equation from Section \ref{sec:lastConvolutionNotPrunedNotQuantized}, the expected error rate for a bit-flip in these positions is:

\begin{equation*}
\begin{split}
    \% error_{17-30} & = \frac{1}{6} (0 + 54.91 + 4.28 + 72.42 + 6.91 + 83.86)
    \\ & = 37.06
\end{split}
\end{equation*}

This value is similar to that obtained experimentally by linearly weighting the error rates in the $[17-30]$ range (37.64\% in Figure \ref{fig:bitFlipErrorNotPrunedQuantized}).

Considering that a bit-flip in the sign bit significantly modifies the magnitude, interpreting the results requires an inverse explanation: if the bit-flip occurs in bias 0, the value becomes very positive and will invariably be the winning class, even though it should never win (error of 100\%); if it occurs in bias 1, the bias becomes very negative, and class 1 will never win (error of 45.09\%).
The same reasoning applies to the remaining biases.
Hence, the error rate associated with bit-flips in bit 31 should be approximately the complement of the error associated with bit-flips in bit $[17-30]$, i.e., 62.94\%.
This value is notably smaller than the one obtained experimentally, which is 88\%.
The discrepancy can be attributed to the uneven distribution of errors injected into each of the six biases in the statistical analysis for bit-flip 31 together with the presence of highly unbalanced terms in the error rate equation for bit-flip 31 (complement values are 100, 45.09, 95.72, 27.58, 93.09 and 16.14).

\begin{figure}[h!]
\centering
\includegraphics[width=7.5cm]{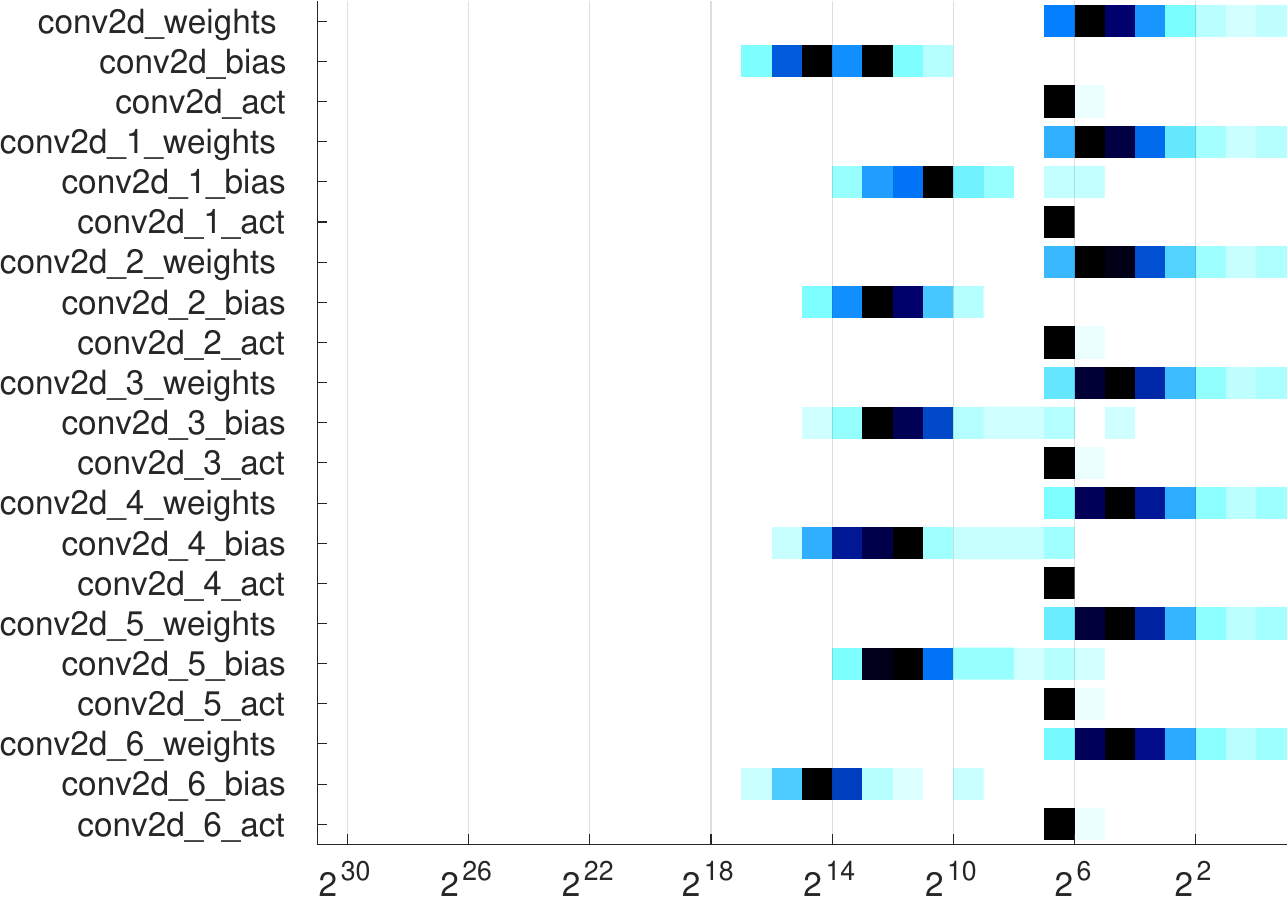}
\caption{Calibration study of the pruned quantized model ($q_w$, $q_b$ and $q_a$) from layers $conv2D$ to $conv2D\_6$.}
\label{fig:calibrationPrunedQuantized_0_6}
\end{figure}

\subsubsection{Analysis of robustness of the pruned model}
\paragraph{Relevance of weights compared to biases}
Despite the decrease in robustness of the pruned model (as illustrated in Figure \ref{fig:bitFlipErrorPrunedQuantized}), the relative negligible importance of weights compared to biases persists.

\begin{figure}[h!]
\centering
\includegraphics[width=6.75cm]{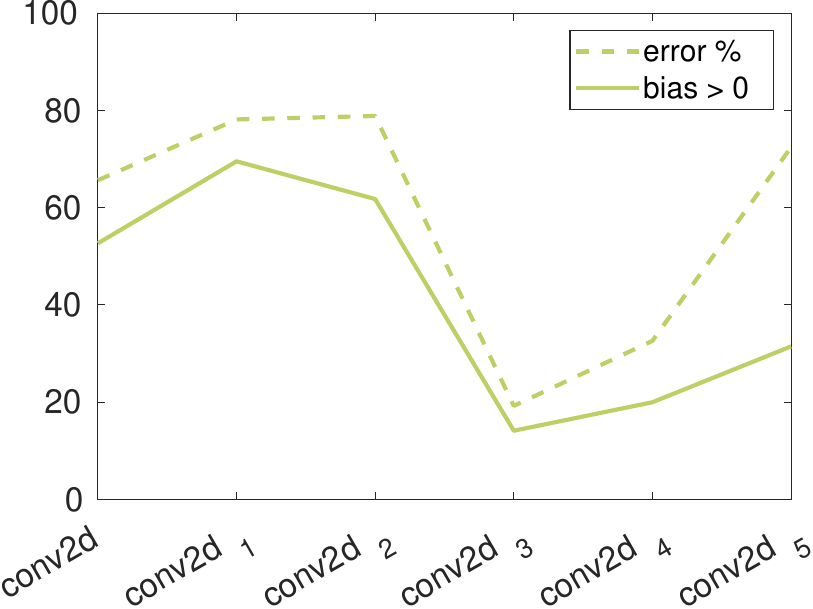}
\caption{Bit-flip error in $[17 - 30]$ bits (dashed) and percentage of positive bias (solid) in $conv2D_{b}$ layers of the pruned quantized model.}
\label{fig:interpretabilityPositiveBiasPrunedQuantized}
\end{figure}

\begin{figure}[h!]
\centering
\includegraphics[width=7.75cm]{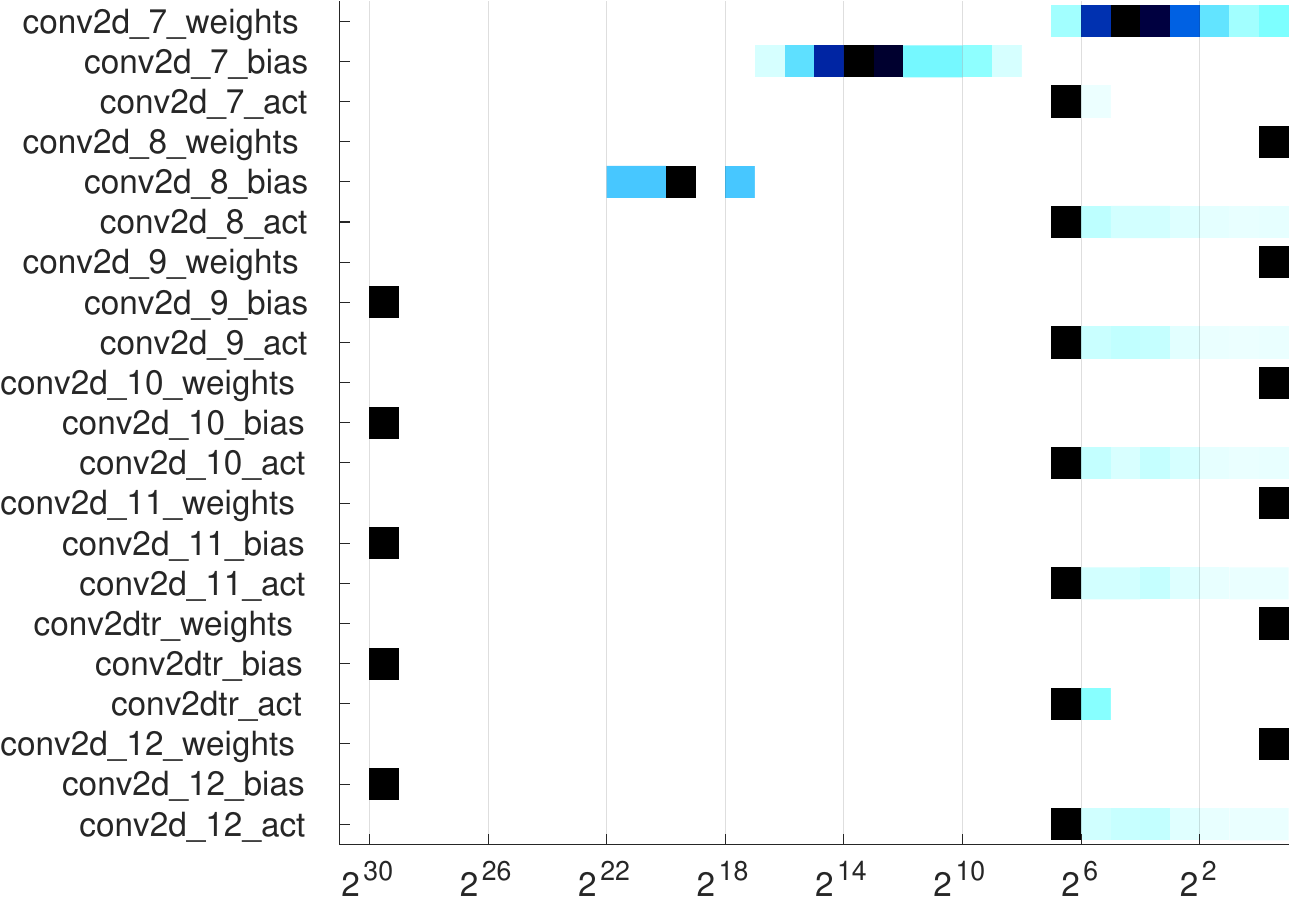}
\caption{Calibration study of the pruned quantized model ($q_w$, $q_b$ and $q_a$) from layers $conv2D\_7$ to $conv2D\_12$.}
\label{fig:calibrationPrunedQuantized_7_12}
\end{figure}

\paragraph{The biases}
The pruning process removes numerous irrelevant channels containing high negative biases along the whole model (Table \ref{tab:biasParametersPrunedQuantized}).
This can be seen by comparing the same figures for the unpruned and pruned case (e.g. Figures \ref{fig:calibrationUnprunedQuantized_0_6} and \ref{fig:calibrationPrunedQuantized_0_6}).
The only region where this cleanup may not be too pronounced is the central area (as seen in the comparison of Figure \ref{fig:calibrationUnprunedQuantized_7_12} with Figure \ref{fig:calibrationPrunedQuantized_7_12}).
However, even in this region, the activations show a less homogeneous distribution, which indicates that associated weights are more meaningful.

If we compare again the ratio of positive biases of the $conv2D$ layers with bit-flip error rate in the $[17 - 30]$ range (Figure \ref{fig:interpretabilityPositiveBiasPrunedQuantized}) the existing correlation between those two quantities for the first six $conv2D$ layers can be observed.

\begin{table}[h!]
\caption{Necessary bits to represent all pos./neg. values of the $conv2D_b$ (green) and $conv2Dtr_b$ (orange) parameters in the quantized pruned model (\textit{*means that there are no values of that sign}).}
\label{tab:biasParametersPrunedQuantized}
\centering
\resizebox{7cm}{!}{
\begin{tabular}{|c|c|c|c|c|c|}
\hline
\textbf{Layer} & \textbf{Positive} & \textbf{Negative} & \textbf{Layer} & \textbf{Positive} & \textbf{Negative} \\ \hline
{\color[HTML]{009901} \textbf{0}} & 18 & 17 & {\color[HTML]{009901} \textbf{22}} & 12 & 16 \\ \hline
{\color[HTML]{009901} \textbf{1}} & 15 & 15 & {\color[HTML]{009901} \textbf{21}} & 17 & 14 \\ \hline
{\color[HTML]{009901} \textbf{2}} & 16 & 16 & {\color[HTML]{009901} \textbf{20}} & 16 & 15 \\ \hline
{\color[HTML]{009901} \textbf{3}} & 15 & 16 & {\color[HTML]{F8A102} \textbf{4}} & 17 & -* \\ \hline
{\color[HTML]{009901} \textbf{4}} & 16 & 17 & {\color[HTML]{009901} \textbf{19}} & 17 & 14 \\ \hline
{\color[HTML]{009901} \textbf{5}} & 15 & 15 & {\color[HTML]{009901} \textbf{18}} & 15 & 15 \\ \hline
{\color[HTML]{009901} \textbf{6}} & 18 & 18 & {\color[HTML]{F8A102} \textbf{3}} & 19 & -* \\ \hline
{\color[HTML]{009901} \textbf{7}} & 18 & 17 & {\color[HTML]{009901} \textbf{17}} & 17 & 16 \\ \hline
{\color[HTML]{009901} \textbf{8}} & 23 & 22 & {\color[HTML]{009901} \textbf{16}} & 16 & 15 \\ \hline
{\color[HTML]{009901} \textbf{9}} & 32 & 31 & {\color[HTML]{F8A102} \textbf{2}} & 18 & -* \\ \hline
{\color[HTML]{009901} \textbf{10}} & 32 & 31 & {\color[HTML]{009901} \textbf{15}} & 18 & 12 \\ \hline
{\color[HTML]{009901} \textbf{11}} & 32 & 31 & {\color[HTML]{009901} \textbf{14}} & 18 & 18 \\ \hline
{\color[HTML]{F8A102} \textbf{0}} & 32 & 31 & {\color[HTML]{F8A102} \textbf{1}} & 32 & -* \\ \hline
{\color[HTML]{009901} \textbf{12}} & 32 & -* & {\color[HTML]{009901} \textbf{13}} & 32 & -* \\ \hline
\end{tabular}}
\end{table}

Finally, it is worth noting that, similar to the unpruned case, bit-flips occurring below position 9 do not significantly affect the output.
Similarly, errors become increasingly detrimental to the model from position 16 onwards.

\begin{figure}[h!]
\centering
\includegraphics[width=7.75cm]{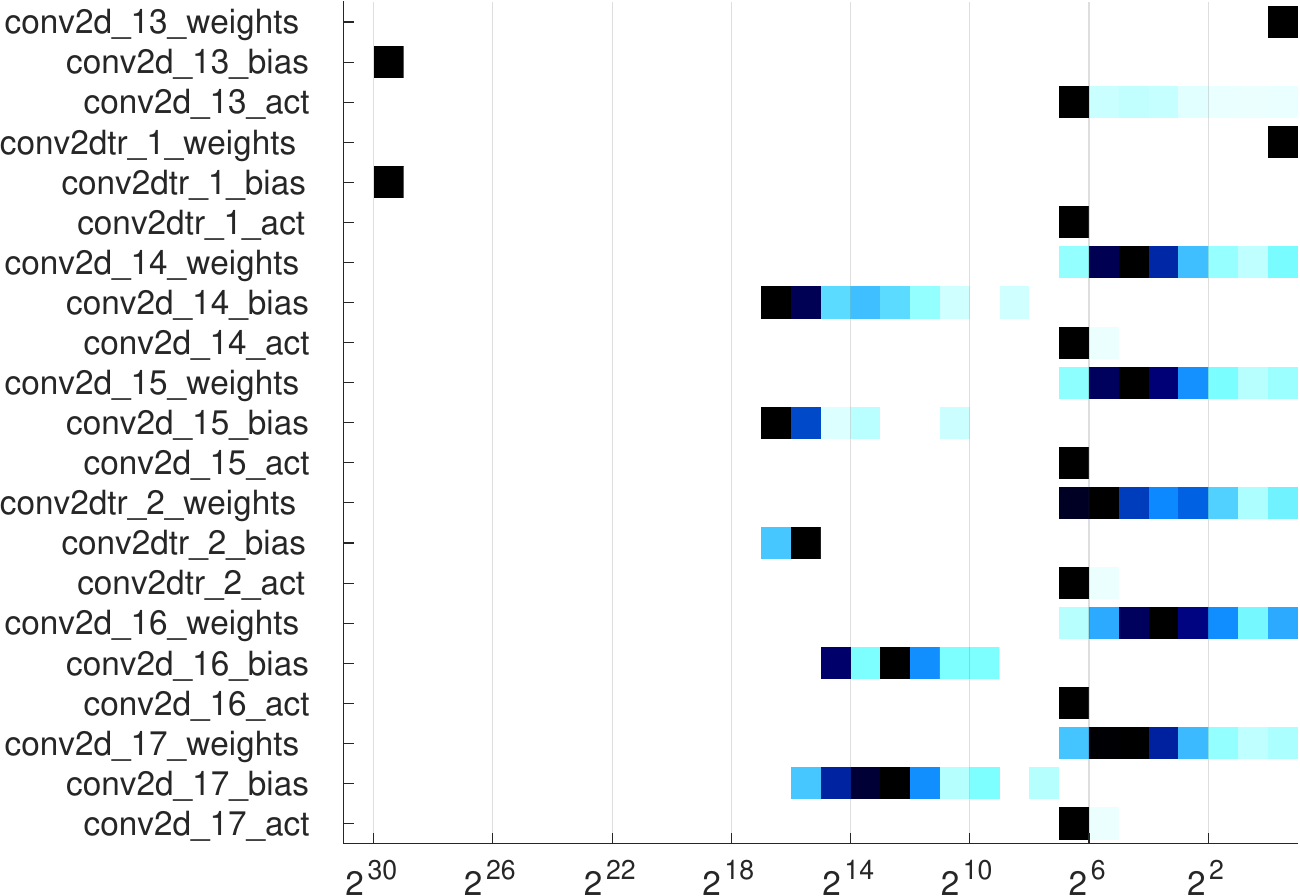}
\caption{Calibration study of the pruned quantized model ($q_w$, $q_b$ and $q_a$) from layers $conv2D\_13$ to $conv2D\_17$.}
\label{fig:calibrationPrunedQuantized_13_17}
\end{figure}

\paragraph{Deep layers: the base of the U-Net}
The QNN exhibits the same characteristics at the base of the U-Net as the quantized unpruned model.
Specifically, the values in $p16$-$p32$ sets do not generate any errors in the output, regardless of the position of bit-flips.
This is a consequence of the small values of convolution weights, and only biases (as depicted in Figures \ref{fig:calibrationPrunedQuantized_7_12} and \ref{fig:calibrationPrunedQuantized_13_17}) contribute significantly to the activations.

\begin{figure}[h!]
\centering
\includegraphics[width=7.5cm]{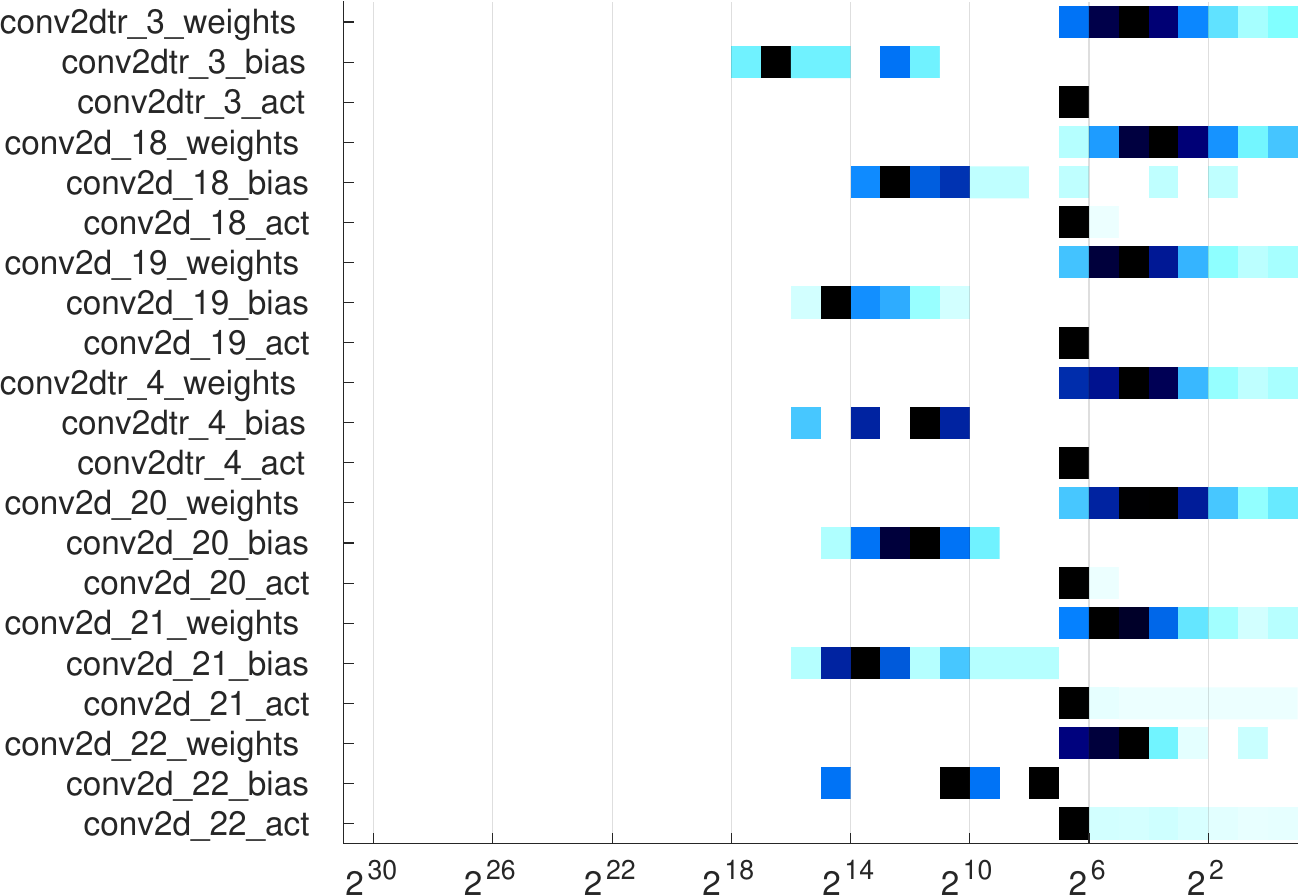}
\caption{Calibration study of the pruned quantized model ($q_w$, $q_b$ and $q_a$) from layers $conv2D\_18$ to $conv2D\_22$.}
\label{fig:calibrationPrunedQuantized_18_22}
\end{figure}

\paragraph{Output $conv2D$ layer}
The explanation given for the unpruned model remains applicable.
Despite having slightly smaller magnitude bias values ($[-19945$, $1730$, $-150$, $209$, $-1120$, $665]$), as depicted in Figure \ref{fig:calibrationPrunedQuantized_18_22}, these biases retain their sign bit.
The error rates experimentally obtained for bit-flips in positions $[17 - 30]$ extracted from the linear weighting of the data is $37.89\%$, consistent with the theoretical error value $37.06\%$.

\begin{equation*}
\begin{split}
    \% error_{17-30} & = \frac{1}{6} (0 + 55.68 + 4.24 + 73.03 + 6.96 + 82.48)
    \\ & = 37.06
\end{split}
\end{equation*}

The error associated with a bit-flip in bit 31 is 93\%, exceeding the expected value due to the uneven distribution of bit-flips among the biases.

\subsubsection{Analysis of multiple random bit-flips}
To estimate error rates under regular operating conditions, we have performed a statistical analysis of MBUs on both quantized unpruned and quantized pruned models.
Since FIT (Failures in Time) values reported by manufacturers such as AMD-Xilinx \cite{deviceReliabilityReport} for the most up-to-date FPGA technologies are in the range of tens of FIT/Mb, and given that targeted BFAs in embedded systems would be very improbable, this study has been performed based on a randomized fault injection campaign.

\begin{figure}[h!]
\centering
\includegraphics[width=7.75cm]{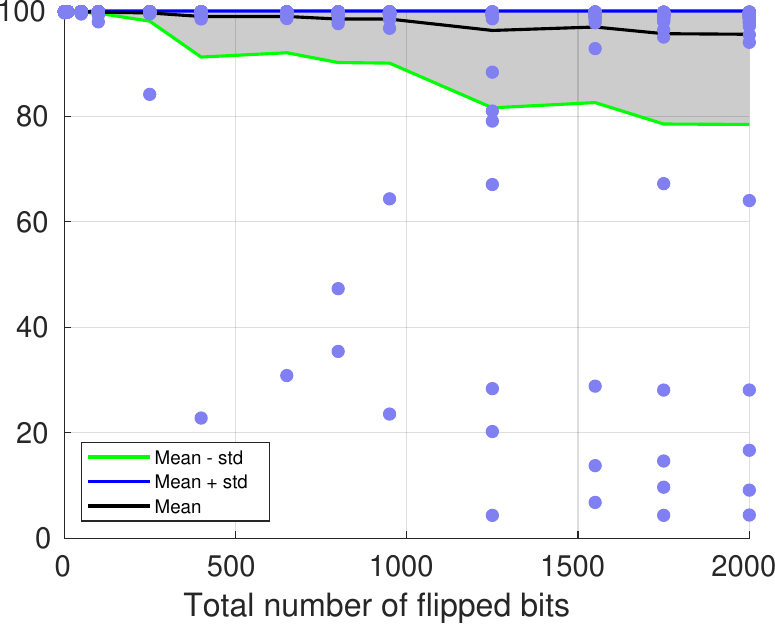}
\caption{Mean error rate due to multiple bit-flips for the quantized unpruned model. Each purple dot represents the error rate of individual repetitions of the injections.}
\label{fig:multipleBitFlipErrorNotPruned}
\end{figure}

\begin{figure}[h!]
\centering
\includegraphics[width=7.75cm]{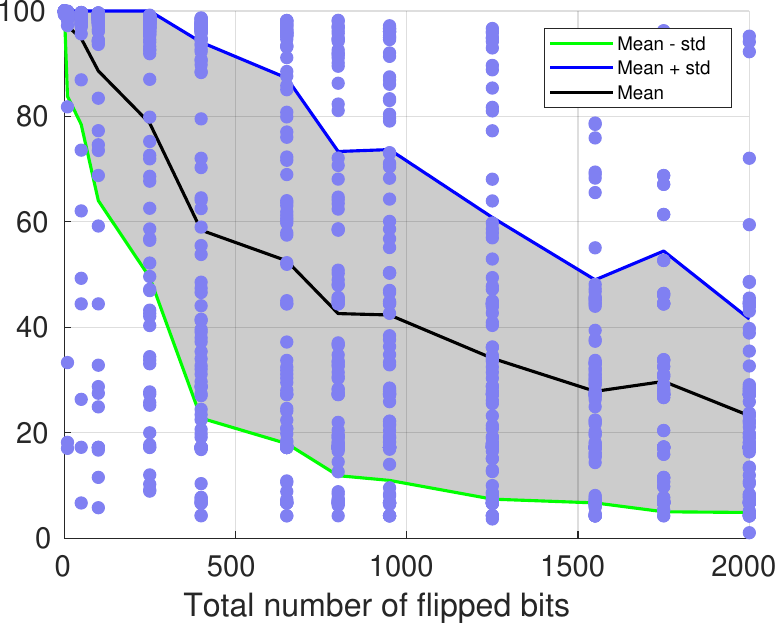}
\caption{Mean error rate due to multiple bit-flips for the quantized pruned model. Each purple dot represents the error rate of individual repetitions of the injections.}
\label{fig:multipleBitFlipErrorPruned}
\end{figure}

We randomly injected [1, 10, 50, 100, 250, 400, 650, 800, 950, 1250, 1550, 1750, 2000] bit-flips into the parameters of the entire model and repeated each injection 150 times to obtain average and standard deviation values.
Figures \ref{fig:multipleBitFlipErrorNotPruned} and \ref{fig:multipleBitFlipErrorPruned} show the mean multiple bit-flip error rate on the quantized unpruned and pruned DNNs, respectively.
As expected, the unpruned model is quite robust since the subset of highly sensitive parameters is small compared to the whole parameter set, so the probability of failure is low compared to that of the pruned model.
Nevertheless, it is worth noting that assuming a failure rate of 20 FiT/Mb (based on data from \cite{deviceReliabilityReport}), the unpruned model, with 100 times more parameters (see Table \ref{tab:modelSize}), experiences 5 SEUs/hour, whereas the pruned model experiences 0.05 SEUs/hour.
Specifically, for models quantized to INT8, the unpruned model would accumulate 1000 upsets in just 8.3 days of operation, whereas the pruned model would require approximately 2.3 years.

The results align with those reported by other authors in the context of image classification.
For instance, the accuracy loss in \cite{li2020defending} for the DNN when exposed to random BFAs is quite similar to that of the unpruned model (Figure \ref{fig:multipleBitFlipErrorNotPruned}).
Similarly, the accuracy loss when faults are injected into sensitive parameters is comparable to that of the pruned model (Figure \ref{fig:multipleBitFlipErrorPruned}).
In this regard, injecting random faults into an optimized and compressed model is analogous to injecting targeted faults into an uncompressed and not optimized model.
In \cite{rakin2019bit}, it is also noted that for any given network, a very high number of random faults must be injected to significantly degrade inference quality.
Finally, \cite{he2020defending} identifies several techniques as either beneficial or detrimental to network robustness, among which pruning is detrimental, whereas increasing network capacity is beneficial.

\section{Robustness enhancement techniques}\label{sec:protectionMethod}
As discussed in Subsection \ref{sec:hardeningTechniques}, neural network hardening methods against single bit-flips can be categorized into four groups: redundancy-based methods, activation function modifications, parameter modifications, and training/inference process modifications.
Key comparison features include recovered accuracy, storage overhead, time overhead, and retraining requirements.

Redundancy-based methods, such as triple-modular redundancy \cite{ruospo2022selective}, introduce significant storage and time overhead.
This is due to the need to compare the original model parameters against two additional copies.
However, these methods do not require retraining.

Methods that modify activation functions also introduce storage and time overhead, which can be either fixed \cite{hoang2020ft, chen2021low} or model-dependent and variable \cite{taheri2024exploration, zhan2021improving}.
Some methods, such as \cite{ghavami2022fitact}, require the modified activation function to be included during training, necessitating dataset availability and introducing a possible training overhead (the authors report around 6\% training overhead).

Methods based on parameter modifications generally do not introduce storage overhead.
For instance, \cite{schorn2019efficient} does not add storage or time overhead but requires retraining to create a more homogeneous architecture with fewer critical bottlenecks.
The MATE method proposed in \cite{jang2021mate} avoids retraining and memory overhead by replacing non-critical mantissa bits with error correction codes, although it does introduce a time overhead.
Similarly, \cite{burel2021zero} adjusts the parity of weights by flipping the least significant bits as needed, avoiding storage overhead but increasing inference time as faults are masked with zero values.

The requirements for methods that modify the training or inference process vary.
The method proposed in \cite{gambardella2022accelerated} relies on adversarial training, requiring dataset availability.
In contrast, the method in \cite{draghetti2019detecting} uses the inference results of previous frames, adding both storage and time overhead.
The method described in \cite{lee2022bipolar} has zero overhead, but is only applicable if the CNN ends in a fully connected layer.

Regarding methods that aim to mitigate or mask BFAs, the approach described in \cite{wang2023aegis} requires dataset availability because it involves training additional internal classifiers, which increases storage overhead.
However, due to the nature of this method, which promotes early exits, inference time is reported to be reduced to 46.1\% - 59.4\% of the original model's inference time.
In a different approach, the authors of \cite{9643556} extract a unique signature from the DNN offline and verify the integrity of the model on the fly.
This method introduces additional time and storage overheads.
While these approaches enhance resilience against faults, they often involve large memory and computation overheads, challenging their suitability for safety-critical DNN inference on edge devices.
Therefore, in this work, we aim to enhance robustness without increasing memory or computational costs.

\begin{figure*}[b]
\centering
\includegraphics[width=17cm]{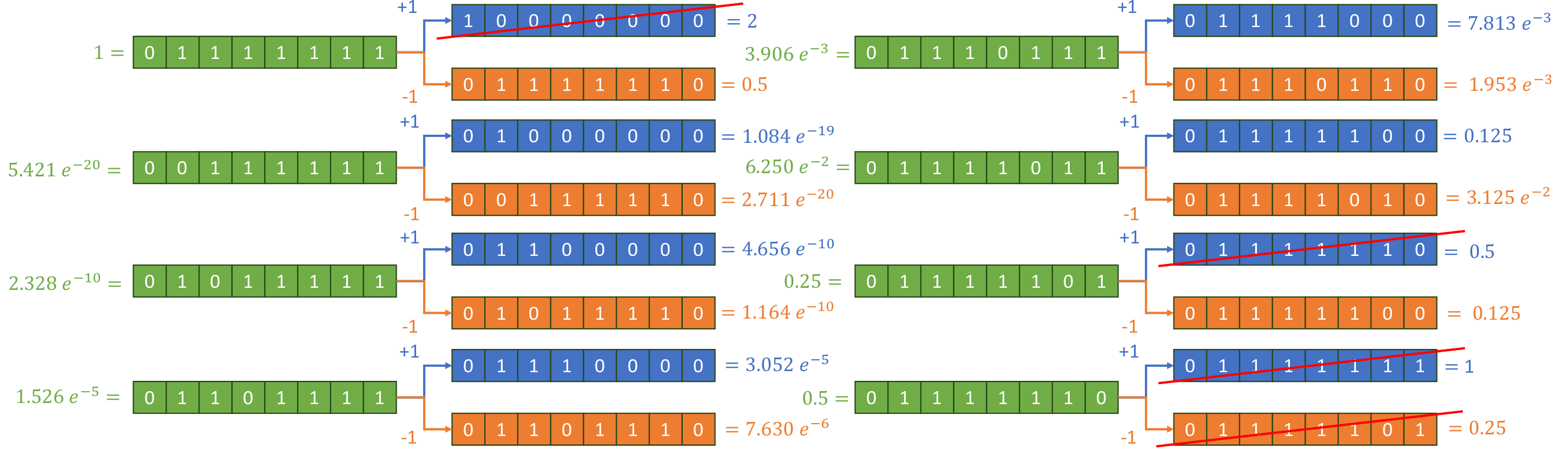}
\caption{Values of the partially filled exponents (in green) after increasing (in blue) or decreasing (in orange) its value by 1.
The three values which are crossed in red are not useful as either the number of '0'-valued positions does not increase or the value of the parameter increases above 2.}
\label{fig:exponentMantissaMethod}
\end{figure*}

The in detail analysis of error propagation described in the previous sections allows for identifying some key factors that shape the sensitivity of segmentation FCNs to SBUs.
As explained, this sensitivity depends mainly on the combination of two factors: the location of the affected parameters in the architecture of the models, and the parameter values (range, sign and their numeric representation).
Based on this knowledge, it is possible to establish some design rules to protect the model from being too sensitive to SBUs by avoiding "risky" states while preserving performance.
One of such states in single-precision floating-point (IEEE 754 format) is having too many parameters which are one bit-flip away from having the exponent filled, because this representation stands for two especially destructive values: $NaN$ and $\pm\infty$.
Consequently, when a SBU causes the exponent to be filled, undesirable results are produced during inference.
Similarly, partial exponent filling (leaving aside the MSB) as illustrated in Figures \ref{fig:biasConv2D_22exponent_pruned} and \ref{fig:biasConv2Dtr_4exponent_pruned}, also leads to a notable number of errors.
This occurs because a number with an absolute value smaller than 1 transforms into a number with an absolute value greater than 1 (but still smaller than 2).
Given that weight and bias values typically fall within the range of 0 to 1 (except for the $\gamma$ values of the BN layers), such situations are common and require attention.

\subsection{Modifying model's parameterization}\label{sec:ourMethod}
The proposed protection method involves identifying parameters with exponents containing either seven bits set to '1' or partial exponents with six bits set to '1'.
This is graphically illustrated on some example values in Figure \ref{fig:exponentMantissaMethod}, where coloured in green are the values with an exponent identified as "risky" by the tool.
Exponents containing seven '1's (with the MSB set to '1') have not been included, as parameters with such high values are not likely in neural networks.

The proposed method involves incrementing or decrementing the exponent by one, with the aim of making the number of '0's to be greater than one.
This prevents the partial exponent from being filled if a bit-flip event occurs.
However, not all the exponents with one '0' are candidates for this method.
For example, modifying exponent $01111110$ by 1 would not be useful (Figure \ref{fig:exponentMantissaMethod}), as it would either decrement the number of '0's or leave it unchanged.
Similarly, incrementing $01111101$ would not be useful either, as the number of '0's would remain unchanged.
Furthermore, $10000000$ has been omitted because bit-flips in the partial exponent would significantly increase the value of the parameter above 2.

After performing the increment/decrement of the exponent, the resultant value of the parameter is doubled/halved.
Thus, it has to be compensated by modifying the mantissa.
If the exponent is increased, the mantissa value needs to be reduced to the minimum value (1.0), while if the exponent is decreased, the mantissa value needs to be incremented to the maximum value ($\approx$ 1.999).
Depending on the original value of the mantissa, the protected value will differ more or less from the original.
This modification of the parameter values would produce a perturbation in the model's performance, thus the tool allows for setting upper and lower thresholds for the mantissas to be identified as "full" or "empty" and, thus, candidates for protection.
These thresholds define the "protection target" (PT) parameter of the tool.
According to this setting, the tool can first evaluate in what extent the performance of the model is modified.

\begin{table}[h!]
\centering
\caption{Number of candidate parameters to protect for the pruned and unpruned models according to the PT setting.}
\label{tab:numParametersProtectionTarget}
\resizebox{7.5cm}{!}{
\begin{tabular}{c|cccc|}
\cline{2-5}
  & \multicolumn{4}{c|}{\textbf{Protected parameters}} \\ \cline{2-5}
  & \multicolumn{2}{c|}{\textbf{Not Folded}} & \multicolumn{2}{c|}{\textbf{Folded}} \\ \hline
\multicolumn{1}{|c|}{\textbf{Protection}} & \multicolumn{1}{c|}{\textbf{NotPruned}} & \multicolumn{1}{c|}{\textbf{Pruned}} & \multicolumn{1}{c|}{\textbf{NotPruned}} & \textbf{Pruned} \\ \hline
\multicolumn{1}{|c|}{\textbf{PT1}} & \multicolumn{1}{c|}{2433}  & \multicolumn{1}{c|}{126}   & \multicolumn{1}{c|}{721}   & 160\\ \hline
\multicolumn{1}{|c|}{\textbf{PT2}} & \multicolumn{1}{c|}{5993}  & \multicolumn{1}{c|}{1095}  & \multicolumn{1}{c|}{7497}  & 1551 \\ \hline
\multicolumn{1}{|c|}{\textbf{PT3}} & \multicolumn{1}{c|}{21544} & \multicolumn{1}{c|}{5316}  & \multicolumn{1}{c|}{36534} & 7809 \\ \hline
\multicolumn{1}{|c|}{\textbf{PT4}} & \multicolumn{1}{c|}{40754} & \multicolumn{1}{c|}{10498} & \multicolumn{1}{c|}{71141} & 15368 \\ \hline
\end{tabular}}
\end{table}

To illustrate how this technique performs on the models under study, four different PTs have been explored: PT1 (1.999, 1.001), PT2 (1.99, 1.01), PT3 (1.95, 1.05), and PT4 (1.9, 1.1).
PT0 stands for unprotected models.
Table \ref{tab:numParametersProtectionTarget} shows the number of candidate parameters selected for protection by the tool according to the protection target.

\begin{table}[h!]
\caption{Comparison of IoU values of the original model (PT0) and the protected models (PT1-PT4). BN unpruned unfolded.}
\label{tab:notPrunednotFoldedProtectedEvaluation}
\centering
\resizebox{7cm}{!}{
\begin{tabular}{c|c|cccc|}
\hline
\multicolumn{1}{|c|}{\textbf{Class/Target}} & \textbf{PT0}        & \multicolumn{1}{c|}{\textbf{PT1}} & \multicolumn{1}{c|}{\textbf{PT2}} & \multicolumn{1}{c|}{\textbf{PT3}} & \textbf{PT4} \\ \hline
\multicolumn{1}{|c|}{\textbf{Road}}                   & 97.84             & \multicolumn{1}{c|}{97.84}      & \multicolumn{1}{c|}{97.85}      & \multicolumn{1}{c|}{9.31}       & 16.36      \\ \hline
\multicolumn{1}{|c|}{\textbf{Road Marks}}             & 87.99             & \multicolumn{1}{c|}{87.99}      & \multicolumn{1}{c|}{88.00}      & \multicolumn{1}{c|}{45.77}      & 5.50       \\ \hline
\multicolumn{1}{|c|}{\textbf{Vegetation}}             & 94.23             & \multicolumn{1}{c|}{94.23}      & \multicolumn{1}{c|}{94.24}      & \multicolumn{1}{c|}{1.93}       & 12.62      \\ \hline
\multicolumn{1}{|c|}{\textbf{Sky}}                    & 92.83             & \multicolumn{1}{c|}{92.83}      & \multicolumn{1}{c|}{92.85}      & \multicolumn{1}{c|}{5.05}       & 2.42       \\ \hline
\multicolumn{1}{|c|}{\textbf{Others}}                 & 78.12             & \multicolumn{1}{c|}{78.12}      & \multicolumn{1}{c|}{78.17}      & \multicolumn{1}{c|}{10.35}      & 12.26      \\ \hline
\multicolumn{1}{|c|}{\textbf{Global}}                 & 94.71             & \multicolumn{1}{c|}{94.71}      & \multicolumn{1}{c|}{94.72}      & \multicolumn{1}{c|}{8.76}       & 14.20      \\ \hline
\multicolumn{1}{|c|}{\textbf{Weighted}}               & 88.54             & \multicolumn{1}{c|}{88.54}      & \multicolumn{1}{c|}{88.56}      & \multicolumn{1}{c|}{25.23}      & 6.38       \\ \hline
\end{tabular}}
\end{table}

Tables \ref{tab:notPrunednotFoldedProtectedEvaluation} to \ref{tab:prunedFoldedProtectedEvaluation} compare the IoU metric on the test set between the original unprotected model and the modified protected ones with different PT settings.
Depending on the analysed model, the degradation of the IoU metric shows different sensitivity to the PT value although, as expected, the higher the PT, the higher the model perturbation.

\begin{table}[h!]
\caption{Comparison of IoU values of the original model (PT0) and the protected models (PT1-PT4). BN pruned unfolded.}
\label{tab:prunedNotFoldedProtectedEvaluation}
\centering
\resizebox{7cm}{!}{
\begin{tabular}{c|c|cccc|}
\hline
\multicolumn{1}{|c|}{\textbf{Class/Target}} & \textbf{PT0}        & \multicolumn{1}{c|}{\textbf{PT1}} & \multicolumn{1}{c|}{\textbf{PT2}} & \multicolumn{1}{c|}{\textbf{PT3}} & \textbf{PT4} \\ \hline
\multicolumn{1}{|c|}{\textbf{Road}}                   &  97.73            & \multicolumn{1}{c|}{97.73}      & \multicolumn{1}{c|}{97.73}      & \multicolumn{1}{c|}{97.73}      & 97.48      \\ \hline
\multicolumn{1}{|c|}{\textbf{Road Marks}}             &  88.17            & \multicolumn{1}{c|}{88.17}      & \multicolumn{1}{c|}{88.16}      & \multicolumn{1}{c|}{88.23}      & 88.13      \\ \hline
\multicolumn{1}{|c|}{\textbf{Vegetation}}             &  93.66            & \multicolumn{1}{c|}{93.66}      & \multicolumn{1}{c|}{93.67}      & \multicolumn{1}{c|}{93.76}      & 93.61      \\ \hline
\multicolumn{1}{|c|}{\textbf{Sky}}                    &  92.89            & \multicolumn{1}{c|}{92.89}      & \multicolumn{1}{c|}{92.88}      & \multicolumn{1}{c|}{92.77}      & 91.63      \\ \hline
\multicolumn{1}{|c|}{\textbf{Others}}                 &  77.34            & \multicolumn{1}{c|}{77.34}      & \multicolumn{1}{c|}{77.36}      & \multicolumn{1}{c|}{77.24}      & 76.46      \\ \hline
\multicolumn{1}{|c|}{\textbf{Global}}                 &  94.46            & \multicolumn{1}{c|}{94.46}      & \multicolumn{1}{c|}{94.46}      & \multicolumn{1}{c|}{94.46}      & 94.14      \\ \hline
\multicolumn{1}{|c|}{\textbf{Weighted}}               &  88.48            & \multicolumn{1}{c|}{88.48}      & \multicolumn{1}{c|}{88.48}      & \multicolumn{1}{c|}{88.47}      & 87.97      \\ \hline
\end{tabular}}
\end{table}

For the unpruned unfolded model, PT2 is the highest feasible PT (Table \ref{tab:notPrunednotFoldedProtectedEvaluation}).
For the unpruned folded model, it has to be assessed whether PT3 is too aggressive or not (Table \ref{tab:notPrunedFoldedProtectedEvaluation}).
Regarding pruned models, PT3 is adequate so it remains to be determined whether PT4 is too demanding or not (Tables \ref{tab:prunedNotFoldedProtectedEvaluation} and \ref{tab:prunedFoldedProtectedEvaluation}).

\begin{table}[h!]
\caption{Comparison of IoU values of the original model (PT0) and the protected models (PT1-PT4). BN unpruned folded.}
\label{tab:notPrunedFoldedProtectedEvaluation}
\centering
\resizebox{7cm}{!}{
\begin{tabular}{c|c|cccc|}
\hline
\multicolumn{1}{|c|}{\textbf{Class/Target}} & \textbf{PT0}        & \multicolumn{1}{c|}{\textbf{PT1}} & \multicolumn{1}{c|}{\textbf{PT2}} & \multicolumn{1}{c|}{\textbf{PT3}} & \textbf{PT4} \\ \hline
\multicolumn{1}{|c|}{\textbf{Road}}                   &  97.84            & \multicolumn{1}{c|}{97.84}      & \multicolumn{1}{c|}{97.84}      & \multicolumn{1}{c|}{97.75}      & 97.57      \\ \hline
\multicolumn{1}{|c|}{\textbf{Road Marks}}             &  87.99            & \multicolumn{1}{c|}{88.00}      & \multicolumn{1}{c|}{88.00}      & \multicolumn{1}{c|}{86.80}      & 85.96      \\ \hline
\multicolumn{1}{|c|}{\textbf{Vegetation}}             &  94.23            & \multicolumn{1}{c|}{94.23}      & \multicolumn{1}{c|}{94.23}      & \multicolumn{1}{c|}{94.15}      & 93.99      \\ \hline
\multicolumn{1}{|c|}{\textbf{Sky}}                    &  92.83            & \multicolumn{1}{c|}{92.83}      & \multicolumn{1}{c|}{92.83}      & \multicolumn{1}{c|}{92.13}      & 90.79      \\ \hline
\multicolumn{1}{|c|}{\textbf{Others}}                 &  78.12            & \multicolumn{1}{c|}{78.12}      & \multicolumn{1}{c|}{78.12}      & \multicolumn{1}{c|}{77.12}      & 75.18      \\ \hline
\multicolumn{1}{|c|}{\textbf{Global}}                 &  94.71            & \multicolumn{1}{c|}{94.71}      & \multicolumn{1}{c|}{94.71}      & \multicolumn{1}{c|}{94.47}      & 94.05      \\ \hline
\multicolumn{1}{|c|}{\textbf{Weighted}}               &  88.54            & \multicolumn{1}{c|}{88.54}      & \multicolumn{1}{c|}{88.54}      & \multicolumn{1}{c|}{87.61}      & 86.53      \\ \hline
\end{tabular}}
\end{table}

To evaluate the suitability of applying a certain PT setting, the following points have been taken into account.
As applying this methodology does not either create new potentially dangerous situations or solve the unprotected dangerous ones, PTs are going to be evaluated by injecting a single bit-flip in the "risky" position (so that exponent or partial exponent would be filled) in the parameters selected with each PT.
The assessment is based on the following three metrics: the global IoU (GIoU), the weighted IoU (WIoU) and the error rate.
IoU-based metrics evaluate the model just in the labelled pixels while error rate is a comparison against the prediction of the faultless model so, even in absence of bit-flips, protected models could have a certain error rate.
In this section, analysis will be focused only on the results with the maximum achievable PT values.

\begin{table}[h!]
\caption{Comparison of IoU values of the original model (PT0) and the protected models (PT1-PT4). BN pruned folded.}
\label{tab:prunedFoldedProtectedEvaluation}
\centering
\resizebox{7cm}{!}{
\begin{tabular}{c|c|cccc|}
\hline
\multicolumn{1}{|c|}{\textbf{Class/Target}} & \textbf{PT0}        & \multicolumn{1}{c|}{\textbf{PT1}} & \multicolumn{1}{c|}{\textbf{PT2}} & \multicolumn{1}{c|}{\textbf{PT3}} & \textbf{PT4} \\ \hline
\multicolumn{1}{|c|}{\textbf{Road}}                   & 97.73             & \multicolumn{1}{c|}{97.73}      & \multicolumn{1}{c|}{97.72}      & \multicolumn{1}{c|}{97.73}      &  97.78     \\ \hline
\multicolumn{1}{|c|}{\textbf{Road Marks}}             & 88.17             & \multicolumn{1}{c|}{88.17}      & \multicolumn{1}{c|}{88.16}      & \multicolumn{1}{c|}{88.08}      &  87.89     \\ \hline
\multicolumn{1}{|c|}{\textbf{Vegetation}}             & 93.66             & \multicolumn{1}{c|}{93.66}      & \multicolumn{1}{c|}{93.66}      & \multicolumn{1}{c|}{93.61}      &  93.56     \\ \hline
\multicolumn{1}{|c|}{\textbf{Sky}}                    & 92.89             & \multicolumn{1}{c|}{92.89}      & \multicolumn{1}{c|}{92.90}      & \multicolumn{1}{c|}{92.65}      &  92.03     \\ \hline
\multicolumn{1}{|c|}{\textbf{Others}}                 & 77.34             & \multicolumn{1}{c|}{77.33}      & \multicolumn{1}{c|}{77.31}      & \multicolumn{1}{c|}{77.02}      &  76.98     \\ \hline
\multicolumn{1}{|c|}{\textbf{Global}}                 & 94.46             & \multicolumn{1}{c|}{94.46}      & \multicolumn{1}{c|}{94.45}      & \multicolumn{1}{c|}{94.40}      &  94.38     \\ \hline
\multicolumn{1}{|c|}{\textbf{Weighted}}               & 88.48             & \multicolumn{1}{c|}{88.48}      & \multicolumn{1}{c|}{88.48}      & \multicolumn{1}{c|}{88.32}      &  88.05     \\ \hline
\end{tabular}}
\end{table}

\subsubsection{Protection of the unpruned unfolded model}
Applying PT2 yields notable benefits (see Table \ref{tab:NotPrunedNotFoldedOriginalVs2}), as the error rate is significantly reduced while the IoU metrics, especially for bit 26, show improvement.
There is a scenario where the three indicators do not align (bit 25): despite minor improvements in the IoU metrics, the error rate considerably increases for the protected model.
This could be attributed to the fact that, unintentionally, in the absence of faults, the protected model outperforms the non-protected model (see Table \ref{tab:notPrunednotFoldedProtectedEvaluation}), indicating a considerable number of pixels that change their value even before bit-flip injection.

\begin{table}[h!]
\caption{Comparison of the non-protected and PT2 protected unpruned unfolded models on potentially dangerous bit positions.}
\label{tab:NotPrunedNotFoldedOriginalVs2}
\centering
\resizebox{8cm}{!}{
\begin{tabular}{|c|c|c|c|c|c|c|c|}
\hline
\multicolumn{4}{|c|}{\textbf{Non-protected}} & \multicolumn{4}{c|}{\textbf{Protected}} \\ \hline
\textbf{Bit} & \textbf{GIoU} & \textbf{WIoU} & \textbf{\% error} & \textbf{Bit} & \textbf{GIoU} & \textbf{WIoU} & \textbf{\% error} \\ \hline
\color[HTML]{0066B3} \textbf{30} & 89.07 & 80.06 & 7.76 & \color[HTML]{0066B3} \textbf{30} & 90.63 & 82.69 &  0.72 \\ \hline
\color[HTML]{0066B3} \textbf{29} & 92.30 & 85.23 & 0.32 & \color[HTML]{0066B3} \textbf{29} & 92.31 & 85.28 &  0.04 \\ \hline
\color[HTML]{0066B3} \textbf{28} & 92.30 & 85.23 & 1.02 & \color[HTML]{0066B3} \textbf{28} & 92.31 & 85.28 &  0.04 \\ \hline
\color[HTML]{0066B3} \textbf{27} & 92.23 & 85.14 & 1.55 & \color[HTML]{0066B3} \textbf{27} & 92.29 & 85.25 &  0.08 \\ \hline
\color[HTML]{0066B3} \textbf{26} & 90.12 & 81.81 & 4.43 & \color[HTML]{0066B3} \textbf{26} & 91.65 & 83.90 &  1.36 \\ \hline
\color[HTML]{0066B3} \textbf{25} & 83.04 & 73.08 & 5.54 & \color[HTML]{0066B3} \textbf{25} & 83.11 & 73.17 & 10.01 \\ \hline
\color[HTML]{0066B3} \textbf{24} & 92.22 & 85.09 & 1.86 & \color[HTML]{0066B3} \textbf{24} & 92.23 & 85.13 &  0.54 \\ \hline
\end{tabular}}
\end{table}

\subsubsection{Protection of the pruned unfolded model}
Analyzing PT4 for the pruned unfolded model (Table \ref{tab:prunedNotFoldedOriginalVs4}), we can see that a straightforward conclusion cannot be extracted as the benefits of the protection are not shared among all bits.
As a consequence, it is necessary to assess how the protected model behaves beyond the labeled pixels, since erroneous pixels may be situated in regions that are not critical.
Otherwise, protection target should be lowered down to 3, as it has been verified that the protected model outperforms the non-protected one.

\begin{table}[h!]
\caption{Comparison of the non-protected and PT4 protected pruned unfolded models on potentially dangerous bit positions.}
\label{tab:prunedNotFoldedOriginalVs4}
\centering
\resizebox{8cm}{!}{
\begin{tabular}{|c|c|c|c|c|c|c|c|}
\hline
\multicolumn{4}{|c|}{\textbf{Non-protected}} & \multicolumn{4}{c|}{\textbf{Protected}} \\ \hline
\textbf{Bit} & \textbf{GIoU} & \textbf{WIoU} & \textbf{\% error} & \textbf{Bit} & \textbf{GIoU} & \textbf{WIoU} & \textbf{\% error} \\ \hline
\color[HTML]{0066B3} \textbf{30} &   0.00 &  0.00 &   100 & \color[HTML]{0066B3} \textbf{30}  & 26.03 & 24.52 & 79.50 \\ \hline
\color[HTML]{0066B3} \textbf{29}  & 92.48 & 84.50 &  0.00 & \color[HTML]{0066B3} \textbf{29}  & 91.93 & 84.11 &  1.61 \\ \hline
\color[HTML]{0066B3} \textbf{28}  & 92.48 & 84.50 &  0.00 & \color[HTML]{0066B3} \textbf{28}  & 91.93 & 84.11 &  1.61 \\ \hline
\color[HTML]{0066B3} \textbf{27}  & 90.83 & 82.33 &  2.35 & \color[HTML]{0066B3} \textbf{27}  & 91.55 & 83.71 &  2.14 \\ \hline
\color[HTML]{0066B3} \textbf{26}  & 89.82 & 80.76 &  3.51 & \color[HTML]{0066B3} \textbf{26}  & 90.92 & 82.51 &  2.93 \\ \hline
\color[HTML]{0066B3} \textbf{25}  & 78.25 & 67.49 & 14.58 & \color[HTML]{0066B3} \textbf{25}  & 79.24 & 68.26 & 13.98 \\ \hline
\color[HTML]{0066B3} \textbf{24}  & 92.32 & 84.35 &  1.03 & \color[HTML]{0066B3} \textbf{24}  & 91.83 & 83.89 &  2.12 \\ \hline
\end{tabular}}
\end{table}

\subsubsection{Protection of the unpruned folded model}
In the unpruned folded model, PT2 is the highest achievable one, albeit with unimpressive results, as there is neither loss nor significant improvement (see Table \ref{tab:NotPrunedFoldedOriginalVs2}).
Nevertheless, despite the lack of remarkable changes, this PT still yields benefits, as it increases the mean WIoU and reduces the mean error rates.
Thus, its application keeps being beneficial for the model.

\begin{table}[h!]
\caption{Comparison of the non-protected and PT2 protected unpruned folded models on potentially dangerous bit positions.}
\label{tab:NotPrunedFoldedOriginalVs2}
\centering
\resizebox{8cm}{!}{
\begin{tabular}{|c|c|c|c|c|c|c|c|}
\hline
\multicolumn{4}{|c|}{\textbf{Non-protected}} & \multicolumn{4}{c|}{\textbf{Protected}} \\ \hline
\textbf{Bit} & \textbf{GIoU} & \textbf{WIoU} & \textbf{\% error} & \textbf{Bit} & \textbf{GIoU} & \textbf{WIoU} & \textbf{\% error} \\ \hline
\color[HTML]{0066B3} \textbf{30}  & 10.25 &  9.48 & 88.91 & \color[HTML]{0066B3} \textbf{30} & 31.11 & 26.58 & 65.23 \\ \hline
\color[HTML]{0066B3} \textbf{29}  & 92.30 & 85.23 &  0.00 & \color[HTML]{0066B3} \textbf{29} & 92.29 & 85.23 &  0.01 \\ \hline
\color[HTML]{0066B3} \textbf{28}  & 92.30 & 85.23 &  0.00 & \color[HTML]{0066B3} \textbf{28} & 92.30 & 85.23 &  0.02 \\ \hline
\color[HTML]{0066B3} \textbf{27}  & 92.30 & 85.23 &  0.00 & \color[HTML]{0066B3} \textbf{27} & 92.29 & 85.23 &  0.02 \\ \hline
\color[HTML]{0066B3} \textbf{26}  & 92.30 & 85.23 &  0.05 & \color[HTML]{0066B3} \textbf{26} & 92.29 & 85.23 &  0.03 \\ \hline
\color[HTML]{0066B3} \textbf{25}  & 92.27 & 85.15 &  0.43 & \color[HTML]{0066B3} \textbf{25} & 92.28 & 85.20 &  0.31 \\ \hline
\color[HTML]{0066B3} \textbf{24}  & 92.29 & 85.15 &  0.69 & \color[HTML]{0066B3} \textbf{24} & 92.29 & 85.15 &  0.69 \\ \hline
\end{tabular}}
\end{table}

\subsubsection{Protection of the pruned folded model}
In the pruned folded model, the highest achievable PT is 3 (Table \ref{tab:prunedFoldedOriginalVs3}).
Even though the error rate slightly increases (but not more than 0.31\%), positive differences in terms of GIoU and WIoU are found for all bit positions.
After applying PT4, the improvements in WIoU disappear, while the error rate still increases.

\begin{table}[h!]
\centering
\caption{Comparison of the non-protected and PT3 protected pruned folded models on potentially dangerous bit positions.}
\label{tab:prunedFoldedOriginalVs3}
\centering
\resizebox{8cm}{!}{
\begin{tabular}{|c|c|c|c|c|c|c|c|}
\hline
\multicolumn{4}{|c|}{\textbf{Non-protected}} & \multicolumn{4}{c|}{\textbf{Protected}} \\ \hline
\textbf{Bit} & \textbf{GIoU} & \textbf{WIoU} & \textbf{\% error} & \textbf{Bit} & \textbf{GIoU} & \textbf{WIoU} & \textbf{\% error} \\ \hline
\color[HTML]{0066B3} \textbf{30} &  0.00 &  0.00 &  100 & \color[HTML]{0066B3} \textbf{30} & 23.84 & 18.86 & 72.09 \\ \hline
\color[HTML]{0066B3} \textbf{27} & 92.48 & 84.50 & 0.08 & \color[HTML]{0066B3} \textbf{27} & 92.54 & 84.54 &  0.39 \\ \hline
\color[HTML]{0066B3} \textbf{26} & 92.44 & 84.43 & 0.27 & \color[HTML]{0066B3} \textbf{26} & 92.53 & 84.52 &  0.43 \\ \hline
\color[HTML]{0066B3} \textbf{25} & 92.39 & 84.35 & 0.55 & \color[HTML]{0066B3} \textbf{25} & 92.52 & 84.49 &  0.61 \\ \hline
\color[HTML]{0066B3} \textbf{24} & 92.34 & 84.33 & 0.97 & \color[HTML]{0066B3} \textbf{24} & 92.40 & 84.34 &  1.07 \\ \hline
\end{tabular}}
\end{table}

After assessing all the models, it can be concluded that the target from which to start the protection can be deduced just by inspecting the results of Tables \ref{tab:notPrunednotFoldedProtectedEvaluation} to \ref{tab:prunedFoldedProtectedEvaluation}.
For all the models, this target has been at least 2.
Depending on the parameters of the model, a PT of 3, 4, or even higher may still be beneficial.
As this method is completely memory and computation-free, it can be combined with memory-hungry methods, such as modular redundancy, which would focus on protecting just the critical parameters that have not been protected with this method.

\subsection{Additional protection techniques: revisiting model's parameterization}
Additional protection techniques to that proposed in Section \ref{sec:ourMethod} can be applied to enhance model protection with no memory and computational overhead.
In contrast to the previous method, the following methods exclusively focus on the redefinition of the model parameters.
This adjustments ensures that, in the absence of bit-flips, the model retains its original behaviour while preventing potentially critical value changes due to bit-flip events.

\subsubsection{Applying sparse pruning}
Applying sparse pruning can be beneficial in terms of robustness regardless of whether the processor is designed to improve inference performance in the presence of sparse matrices.
In case of identifying that the model has parameters in the range $1 \leq |x| < 2$ that are not relevant for inference, applying sparse pruning (setting irrelevant parameters to 0) prevents the appearance of $NaN$ values in case of bit-flips.

\subsubsection{Redefining batch normalization parameters} \label{sec:memoryFreeBatchNorm}
If folding of batch normalization layers, which can be viewed as depthwise convolutions (see Equation \ref{equ:batchNormalizationAsConvolution}), is not applied, $\gamma$, $\sigma$, $\mu$ and $\beta$ parameters can be reconditioned as long as the weight ($W$) and bias ($b$) values do not change.
Parameter reconditioning can be applied with two aims: either directly reduce the amount of parameters at risk of having the exponent filled or just modify their mantissa values so that the protection method described in the previous section can be applied.

\begin{equation}
    BN(x) = \gamma \frac{(x - \mu)}{\sqrt{\sigma}} + \beta = \frac{\gamma}{\sqrt{\sigma}} x + (\beta - \gamma \frac{\mu}{\sqrt{\sigma}}) = Wx + b
    \label{equ:batchNormalizationAsConvolution}
\end{equation}

\subsubsection{Applying Cross Layer Equalization} \label{sec:memoryFreeCLE}
This method, which is similar to CLE \cite{nagel2019data} for quantization, allows for the protection of convolutional kernels, which contain most of the parameters in the model.
The method is based on the positive scaling equivariance property ($f(s\textbf{x}) = s f(\textbf{x})$) which is held for models containing Parametric Rectified Linear Unit (PReLU) functions.
The activation map $\textbf{y}$ of a certain layer $n$ can be calculated as $f_{n}(\textbf{W}_{n} \cdot \textbf{x} + \textbf{b}_{n})$ where $\textbf{W}_{n}$ and $\textbf{b}_{n}$ are the weights and biases of the convolutional kernel applied to layer $n$ and $\textbf{x}$ is the input cube.
If we calculate the activation map of the next layer $n+1$ in terms of the previous layer $n$, Equation \ref{equ:reparametrizationEquivariance} shows that model parameters can be reconditioned as long as the activation function of the $n$-th layer fulfills the equivariance property (e.g. ReLU and Identity).

\begin{equation}
    \begin{split}
    \textbf{z} & = f_{n+1}(\textbf{W}_{n+1} \cdot f_{n}(\textbf{W}_{n}  \cdot \textbf{x} + \textbf{b}_{n}) + \textbf{b}_{n+1}) \\
        & = f_{n+1}(\textbf{W}_{n+1}  \cdot \textbf{S} \cdot \hat{f}_{n}(\textbf{S}^{-1} \cdot \textbf{W}_{n} \cdot \textbf{x} + \textbf{S}^{-1} \cdot \textbf{b}_{n}) + \textbf{b}_{n+1}) \\
        & = f_{n+1}(\hat{\textbf{W}}_{n+1} \cdot f_{n}(\hat{\textbf{W}}_{n} \cdot \textbf{x} + \hat{\textbf{b}}_{n}) + \textbf{b}_{n+1})
    \end{split}
    \label{equ:reparametrizationEquivariance}
\end{equation}

where $\textbf{S}$ is a diagonal matrix containing the scale factor of the parameters for each of the filters of a convolutional layer.

\subsubsection{Applying bias absorption} \label{sec:memoryFreeBiasAbsorption}
Bias absorption \cite{nagel2019data}, which is applied during post training quantization procedures, can be implemented with a different purpose in this domain: reducing the number of critical parameters in the model.
The only requirement in order to apply this method is that $f(\textbf{W} \textbf{x} + \textbf{b} - \textbf{c}) = f(\textbf{W} \textbf{x} + \textbf{b}) - \textbf{c} $ must be fulfilled.
It is straightforward to see that this condition will not be met by any non-linear activation function, will always be met by identity and will be met by ReLU activations just in certain situations ($\textbf{c} = max(0, \textbf{W} \textbf{x} + \textbf{b}) \forall \textbf{x}$) that are closely related to $\textbf{x}$ data distribution and $\textbf{W}$ and $\textbf{b}$ parameters.
The method is described in Equation \ref{equ:biasAbsorption}, where $\hat{\textbf{b}}_{n} = \textbf{b}_{n} - \textbf{c}$ and $\hat{\textbf{b}}_{n + 1} = \textbf{W}_{n+1} * \textbf{c} + \textbf{b}_{n+1}$.

\begin{equation}
    \begin{split}
    \textbf{z} & = f_{n+1}(\textbf{W}_{n+1} \cdot f_{n}(\textbf{W}_{n}  \cdot \textbf{x} + \textbf{b}_{n}) + \textbf{b}_{n+1}) \\
        & = f_{n+1}(\textbf{W}_{n+1} \cdot (f_{n}(\textbf{W}_{n} \cdot \textbf{x} + \textbf{b}_{n}) + \textbf{c} - \textbf{c}) + \textbf{b}_{n+1}) \\
        & = f_{n+1}(\textbf{W}_{n+1} \cdot (f_{n}(\textbf{W}_{n} \cdot \textbf{x} + \hat{\textbf{b}}_{n}) + \textbf{c}) + \textbf{b}_{n+1}) \\
        & = f_{n+1}(\textbf{W}_{n+1} \cdot (f_{n}(\textbf{W}_{n} \cdot \textbf{x} + \hat{\textbf{b}}_{n}) + \hat{\textbf{b}}_{n+1}) \\
    \end{split}
    \label{equ:biasAbsorption}
\end{equation}

\newpage

\section{Conclusions}\label{sec:conclusiones}
This article analyses the robustness of encoder-decoder type FCNs against SBUs in semantic segmentation tasks and how compression techniques applied for deployments in embedded systems alter this robustness.
The analysis is based on a statistical campaign of software-injected SBUs in different network parameters to conduct a layer-by-layer and bit-by-bit in-depth study of the sources of critical errors.
The reference models used in this study were designed and optimized for hyperspectral image segmentation with application to autonomous driving systems, and were trained on the HSI-Drive v2.0 dataset.
We identified the potentially most problematic bit-flips according to the location of the parameters in the model architecture, their binary representation, range and sign, and relevance to the inference process.
From this analysis, a tool for estimating the model's robustness and a simple memory and computation-free error mitigation method is proposed.

This study reveals that in 32-bit floating-point FCN models with lower-bounded activation functions, such as the commonly used ReLU, critical errors are primarily associated with SBUs that result in an increment in the parameter values.
This is because, as noted by other authors, commonly used machine learning computations are often monotonic.
However, not all increments pose the same risk.
The most sensitive bit is the MSB of the exponent, as its bit-flip significantly increases the model parameters' values from below 2 to $\infty$ or $NaN$, in the worst-case scenario, propagating the error throughout the model.
Given the typical values of the parameters in BN layers, this layer is the most sensitive to such bit-flips.
Furthermore, we have found that this layer extends the range of output values, thus contributing negatively to the propagation of errors that may have occurred in previous layers.
In this regard, it is noted that layers with many positive biases are particularly vulnerable.
Another particularly "risky" situation arises from parameters with almost filled partial exponents.
When a partial exponent is filled by a bit-flip, the parameter undergoes a transition from a value significantly below unity to a new value above unity, leading to a noticeable amount of errors.
Due to the specific characteristics of encoder-decoder type FCNs with skip connections, SBUs in the first layers of the encoder branch and in the last layers of the decoder branch, which are closer to the model's output, produce comparatively higher error rates.
On the other hand, it is observed that perturbations in the parameters at the base of the model have little influence on the output.

Regarding compression techniques, pruned models show higher sensibility to SBUs than unpruned models.
This is attributed to the overparameterized and parallel nature of original unpruned DNN models.
However, the primary advantage of pruned models lies in their smaller size, as it is less likely for an SBU to occur in a model with fewer parameters.
Hence, to accurately assess the influence of pruning on the vulnerability to SBUs, model implementation details such as circuit design and selected target device should be considered.

With respect to data representation, integer-quantized models show higher robustness to bit-flips.
This is due to the particularities of integer binary representation, since there is no option to extreme values such as $NaN$s or $\infty$s to occur.
The employed quantization method, where weights are quantized to 8 bits while biases are quantized to 32 bits, makes the latter more sensitive to SBUs.
Furthermore, it is generally observed that the upper word of the bias representation simply serves to extend the sign, so it could be used to implement error detection and/or correction methods with no memory overhead.
Finally, since the model sensitivity is fundamentally linked to perturbations in the biases and these constitute a tiny subset of the whole parameter-set, memory-intensive error mitigation techniques such as duplication or triple modular redundancy applied only to biases would ensure high model protection with low memory burden.

When combining prunning and quantization it was observed that the fully compressed model is more robust than the pruned unquantized model but less robust than the unpruned quantized model.
It can be concluded that integer quantization always contributes to robustness enhancement, whereas the evaluation of the effects of pruning is tied to the applied compression degree and the final implementation of the model.

Concerning the robustness analysis of quantized models, if instead of a targeted fault injection technique random multiple fault injection is performed, it is observed that the unpruned model exhibits a considerably greater robustness.
This is because, in quantized models, the sensitive parameters are the biases, and not the weights, which constitute a small subset of the total parameters, reducing the likelihood of a random bit-flip occurring in them.
Additionally, due to the network architecture, most parameters are concentrated in the base of the U-Net, where the least critical layers are located.
As a result, after pruning, the robustness against bit-flips is significantly reduced since nearly any bit-flip can affect a sensitive parameter.
Nonetheless, assuming a failure rate of 20 FiT/Mb, it is worth noting that the unpruned model experiences 5 SEUs/hour whereas the pruned model experiences 0.05 SEUs/hour.
This implies that the unpruned quantized model would need just 8.3 days of operation to accumulate 1000 upsets, whereas the pruned model would require approximately 2.3 years.

Finally, based on the performed analysis, a set of complementary protection methods is proposed to mitigate some delicate situations like partial exponent completion.
In particular, we describe a new memory and computation overhead-free method for the protection of certain model parameters in floating-point representation. 
This technique enables the selection of various protection targets, where a higher target involves attempting to protect more parameters in the model.
However, setting too ambitious protection targets may result in an excessive modification of the original parameterization, thus a model performance analysis must be accomplished beforehand.

\bibliographystyle{unsrtnat}
\bibliography{biblio}

\begin{thebibliography}{75}
\providecommand{\natexlab}[1]{#1}
\providecommand{\url}[1]{\texttt{#1}}
\expandafter\ifx\csname urlstyle\endcsname\relax
  \providecommand{\doi}[1]{doi: #1}\else
  \providecommand{\doi}{doi: \begingroup \urlstyle{rm}\Url}\fi

\bibitem[Le~Clainche et~al.(2023)Le~Clainche, Ferrer, Gibson, Cross, Parente,
  and Vinuesa]{le2023improving}
Soledad Le~Clainche, Esteban Ferrer, Sam Gibson, Elisabeth Cross, Alessandro
  Parente, and Ricardo Vinuesa.
\newblock Improving aircraft performance using machine learning: A review.
\newblock \emph{Aerospace Science and Technology}, page 108354, 2023.

\bibitem[{SAE International Aerospace}(2010)]{ARP4754}
{SAE International Aerospace}.
\newblock {Guidelines for Development of Civil Aircraft and Systems}.
\newblock SAE ARP4754A, 2010.
\newblock URL \url{https://www.sae.org/standards/content/arp4754a/}.
\newblock Revision A.

\bibitem[{International Organization for Standardization}(2018)]{ISO26262}
{International Organization for Standardization}.
\newblock {Road vehicles -- Functional safety}.
\newblock ISO 26262, 2018.
\newblock URL \url{https://www.iso.org/standard/68383.html}.

\bibitem[Martinella et~al.(2021)Martinella, Al{\'\i}a, Stark, Coronetti,
  Cazzaniga, Kastriotou, Kadi, Gaillard, Grossner, and
  Javanainen]{martinella2021impact}
Corinna Martinella, Rub{\'e}n~G Al{\'\i}a, Roger Stark, Andrea Coronetti, Carlo
  Cazzaniga, Maria Kastriotou, Yacine Kadi, R{\'e}mi Gaillard, Ulrike Grossner,
  and Arto Javanainen.
\newblock Impact of terrestrial neutrons on the reliability of sic vd-mosfet
  technologies.
\newblock \emph{IEEE Transactions on Nuclear Science}, 68\penalty0
  (5):\penalty0 634--641, 2021.

\bibitem[Baumann(2005)]{baumann2005soft}
Robert Baumann.
\newblock Soft errors in advanced computer systems.
\newblock \emph{IEEE design \& test of computers}, 22\penalty0 (3):\penalty0
  258--266, 2005.

\bibitem[Hirokawa et~al.(2016)Hirokawa, Harada, Sakuta, Watanabe, and
  Hashimoto]{hirokawa2016multiple}
Soichi Hirokawa, Ryo Harada, Kenshiro Sakuta, Yukinobu Watanabe, and Masanori
  Hashimoto.
\newblock Multiple sensitive volume based soft error rate estimation with
  machine learning.
\newblock In \emph{2016 16th European Conference on Radiation and Its Effects
  on Components and Systems (RADECS)}, pages 1--4. IEEE, 2016.

\bibitem[Yan et~al.(2020)Yan, Shi, Liao, Hashimoto, Zhou, and
  Zhuo]{yan2020single}
Zheyu Yan, Yiyu Shi, Wang Liao, Masanori Hashimoto, Xichuan Zhou, and Cheng
  Zhuo.
\newblock When single event upset meets deep neural networks: Observations,
  explorations, and remedies.
\newblock In \emph{2020 25th Asia and South Pacific Design Automation
  Conference (ASP-DAC)}, pages 163--168. IEEE, 2020.

\bibitem[Wang et~al.(2022)Wang, Gao, Wang, Li, Yue, Niu, Yin, Guo, and
  Shen]{wang2022advances}
Wenxiao Wang, Song Gao, Yaqi Wang, Yang Li, Wenjing Yue, Hongsen Niu, Feifei
  Yin, Yunjian Guo, and Guozhen Shen.
\newblock Advances in emerging photonic memristive and memristive-like devices.
\newblock \emph{Advanced Science}, 9\penalty0 (28):\penalty0 2105577, 2022.

\bibitem[Zhang et~al.(2022{\natexlab{a}})Zhang, Wang, Cai, Zhu, Kline, Yang,
  and Wang]{zhang2022wesco}
Jiangwei Zhang, Chong Wang, Yi~Cai, Zhenhua Zhu, Donald Kline, Huazhong Yang,
  and Yu~Wang.
\newblock Wesco: Weight-encoded reliability and security co-design for
  in-memory computing systems.
\newblock In \emph{2022 IEEE Computer Society Annual Symposium on VLSI
  (ISVLSI)}, pages 296--301. IEEE, 2022{\natexlab{a}}.

\bibitem[Kim et~al.(2014)Kim, Daly, Kim, Fallin, Lee, Lee, Wilkerson, Lai, and
  Mutlu]{kim2014flipping}
Yoongu Kim, Ross Daly, Jeremie Kim, Chris Fallin, Ji~Hye Lee, Donghyuk Lee,
  Chris Wilkerson, Konrad Lai, and Onur Mutlu.
\newblock Flipping bits in memory without accessing them: An experimental study
  of dram disturbance errors.
\newblock \emph{ACM SIGARCH Computer Architecture News}, 42\penalty0
  (3):\penalty0 361--372, 2014.

\bibitem[Kim et~al.(2011)Kim, Chandra, Aitken, Blaauw, and
  Sylvester]{kim2011variation}
Daeyeon Kim, Vikas Chandra, Robert Aitken, David Blaauw, and Dennis Sylvester.
\newblock Variation-aware static and dynamic writability analysis for
  voltage-scaled bit-interleaved 8-t srams.
\newblock In \emph{IEEE/ACM International Symposium on Low Power Electronics
  and Design}, pages 145--150. IEEE, 2011.

\bibitem[Razavi et~al.(2016)Razavi, Gras, Bosman, Preneel, Giuffrida, and
  Bos]{razavi2016flip}
Kaveh Razavi, Ben Gras, Erik Bosman, Bart Preneel, Cristiano Giuffrida, and
  Herbert Bos.
\newblock Flip feng shui: Hammering a needle in the software stack.
\newblock In \emph{25th USENIX Security Symposium (USENIX Security 16)}, pages
  1--18, 2016.

\bibitem[Yao et~al.(2020)Yao, Rakin, and Fan]{yao2020deephammer}
Fan Yao, Adnan~Siraj Rakin, and Deliang Fan.
\newblock $\{$DeepHammer$\}$: Depleting the intelligence of deep neural
  networks through targeted chain of bit flips.
\newblock In \emph{29th USENIX Security Symposium (USENIX Security 20)}, pages
  1463--1480, 2020.

\bibitem[Wang et~al.(2023{\natexlab{a}})Wang, Zhang, Wang, Qiu, Zhang, Li, Li,
  Wei, and Zhang]{wang2023aegis}
Jialai Wang, Ziyuan Zhang, Meiqi Wang, Han Qiu, Tianwei Zhang, Qi~Li, Zongpeng
  Li, Tao Wei, and Chao Zhang.
\newblock Aegis: Mitigating targeted bit-flip attacks against deep neural
  networks.
\newblock In \emph{32nd USENIX Security Symposium (USENIX Security 23)}, pages
  2329--2346, 2023{\natexlab{a}}.

\bibitem[Wang et~al.(2023{\natexlab{b}})Wang, Dai, Chen, Huang, Li, Zhu, Hu,
  Lu, Lu, Li, Wang, and Qiao]{wang2023internimage}
Wenhai Wang, Jifeng Dai, Zhe Chen, Zhenhang Huang, Zhiqi Li, Xizhou Zhu,
  Xiaowei Hu, Tong Lu, Lewei Lu, Hongsheng Li, Xiaogang Wang, and Yu~Qiao.
\newblock Internimage: Exploring large-scale vision foundation models with
  deformable convolutions, 2023{\natexlab{b}}.

\bibitem[Violante et~al.(2004)Violante, Sterpone, Ceschia, Bortolato, Bernardi,
  Reorda, and Paccagnella]{1369494}
M.~Violante, L.~Sterpone, M.~Ceschia, D.~Bortolato, P.~Bernardi, M.S. Reorda,
  and A.~Paccagnella.
\newblock Simulation-based analysis of seu effects in sram-based fpgas.
\newblock \emph{IEEE Transactions on Nuclear Science}, 51\penalty0
  (6):\penalty0 3354--3359, 2004.
\newblock \doi{10.1109/TNS.2004.839516}.

\bibitem[Ruospo et~al.(2022)Ruospo, Gavarini, Bragaglia, Traiola, Bosio, and
  Sanchez]{ruospo2022selective}
Annachiara Ruospo, Gabriele Gavarini, Ilaria Bragaglia, Marcello Traiola,
  Alberto Bosio, and Ernesto Sanchez.
\newblock Selective hardening of critical neurons in deep neural networks.
\newblock In \emph{2022 25th International Symposium on Design and Diagnostics
  of Electronic Circuits and Systems (DDECS)}, pages 136--141. IEEE, 2022.

\bibitem[Taheri et~al.(2024)Taheri, Cherezova, Ansari, Jenihhin, Mahani,
  Daneshtalab, and Raik]{taheri2024exploration}
Mahdi Taheri, Natalia Cherezova, Mohammad~Saeed Ansari, Maksim Jenihhin, Ali
  Mahani, Masoud Daneshtalab, and Jaan Raik.
\newblock Exploration of activation fault reliability in quantized systolic
  array-based dnn accelerators, 2024.

\bibitem[Gholami et~al.(2022)Gholami, Kim, Dong, Yao, Mahoney, and
  Keutzer]{gholami2022survey}
Amir Gholami, Sehoon Kim, Zhen Dong, Zhewei Yao, Michael~W Mahoney, and Kurt
  Keutzer.
\newblock A survey of quantization methods for efficient neural network
  inference.
\newblock In \emph{Low-Power Computer Vision}, pages 291--326. Chapman and
  Hall/CRC, 2022.

\bibitem[Gutiérrez-Zaballa(2024{\natexlab{a}})]{myRepo}
Jon Gutiérrez-Zaballa.
\newblock parameterprotection, 2024{\natexlab{a}}.
\newblock URL \url{https://github.com/jonGuti13/parameterProtection}.

\bibitem[Gutiérrez-Zaballa(2024{\natexlab{b}})]{myTensorFI2fork}
Jon Gutiérrez-Zaballa.
\newblock Tensorfi2, 2024{\natexlab{b}}.
\newblock URL \url{https://github.com/jonGuti13/TensorFI2}.

\bibitem[Narayanan et~al.(2023)Narayanan, Chen, Fang, Li, Pattabiraman, and
  DeBardeleben]{tensorfi2}
Niranjhana Narayanan, Zitao Chen, Bo~Fang, Guanpeng Li, Karthik Pattabiraman,
  and Nathan DeBardeleben.
\newblock Fault injection for tensorflow applications.
\newblock \emph{IEEE Transactions on Dependable and Secure Computing},
  20\penalty0 (4):\penalty0 2677--2695, 2023.
\newblock \doi{10.1109/TDSC.2022.3175930}.

\bibitem[Ahmadilivani et~al.(2023)Ahmadilivani, Taheri, Raik, Daneshtalab, and
  Jenihhin]{systematicReview}
Mohammad~Hasan Ahmadilivani, Mahdi Taheri, Jaan Raik, Masoud Daneshtalab, and
  Maksim Jenihhin.
\newblock A systematic literature review on hardware reliability assessment
  methods for deep neural networks.
\newblock \emph{ACM Comput. Surv.}, dec 2023.
\newblock ISSN 0360-0300.
\newblock \doi{10.1145/3638242}.
\newblock URL \url{https://doi.org/10.1145/3638242}.
\newblock Just Accepted.

\bibitem[Arechiga and Michaels(2018{\natexlab{a}})]{arechiga2018robustness}
Austin~P Arechiga and Alan~J Michaels.
\newblock The robustness of modern deep learning architectures against single
  event upset errors.
\newblock In \emph{2018 IEEE High Performance extreme Computing Conference
  (HPEC)}, pages 1--6. IEEE, 2018{\natexlab{a}}.

\bibitem[Malekzadeh et~al.(2021)Malekzadeh, Rohbani, Lu, and
  Ebrahimi]{malekzadeh2021impact}
Elaheh Malekzadeh, Nezam Rohbani, Zhonghai Lu, and Masoumeh Ebrahimi.
\newblock The impact of faults on dnns: A case study.
\newblock In \emph{2021 IEEE International Symposium on Defect and Fault
  Tolerance in VLSI and Nanotechnology Systems (DFT)}, pages 1--6. IEEE, 2021.

\bibitem[Arechiga and Michaels(2018{\natexlab{b}})]{arechiga2018effect}
Austin~P Arechiga and Alan~J Michaels.
\newblock The effect of weight errors on neural networks.
\newblock In \emph{2018 IEEE 8th Annual Computing and Communication Workshop
  and Conference (CCWC)}, pages 190--196. IEEE, 2018{\natexlab{b}}.

\bibitem[Neggaz et~al.(2019)Neggaz, Alouani, Niar, and Kurdahi]{neggaz2019cnns}
Mohamed~A Neggaz, Ihsen Alouani, Smail Niar, and Fadi Kurdahi.
\newblock Are cnns reliable enough for critical applications? an exploratory
  study.
\newblock \emph{IEEE Design \& Test}, 37\penalty0 (2):\penalty0 76--83, 2019.

\bibitem[Sabbagh et~al.(2019)Sabbagh, Gongye, Fei, and
  Wang]{sabbagh2019evaluating}
Majid Sabbagh, Cheng Gongye, Yunsi Fei, and Yanzhi Wang.
\newblock Evaluating fault resiliency of compressed deep neural networks.
\newblock In \emph{2019 IEEE International Conference on Embedded Software and
  Systems (ICESS)}, pages 1--7. IEEE, 2019.

\bibitem[Goldstein et~al.(2020)Goldstein, Srinivasan, Das, Banerjee, Santiago,
  Ferreira, Nery, Kundu, and Fran{\c{c}}a]{goldstein2020reliability}
Brunno~F Goldstein, Sudarshan Srinivasan, Dipankar Das, Kunal Banerjee, Leandro
  Santiago, Victor~C Ferreira, Alexandre~S Nery, Sandip Kundu, and Felipe~MG
  Fran{\c{c}}a.
\newblock Reliability evaluation of compressed deep learning models.
\newblock In \emph{2020 IEEE 11th Latin American Symposium on Circuits \&
  Systems (LASCAS)}, pages 1--5. IEEE, 2020.

\bibitem[Bosio et~al.(2019)Bosio, Bernardi, Ruospo, and
  Sanchez]{bosio2019reliability}
Alberto Bosio, Paolo Bernardi, Annachiara Ruospo, and Ernesto Sanchez.
\newblock A reliability analysis of a deep neural network.
\newblock In \emph{2019 IEEE Latin American Test Symposium (LATS)}, pages 1--6.
  IEEE, 2019.

\bibitem[Ruospo et~al.(2023)Ruospo, Gavarini, De~Sio, Guerrero, Sterpone,
  Reorda, S{\'a}nchez, Mariani, Aribido, and Athavale]{ruospo2023assessing}
Annachiara Ruospo, Gabriele Gavarini, Corrado De~Sio, J~Guerrero, Luca
  Sterpone, M~Sonza Reorda, Ernesto S{\'a}nchez, Riccardo Mariani, J~Aribido,
  and Jyotika Athavale.
\newblock Assessing convolutional neural networks reliability through
  statistical fault injections.
\newblock In \emph{2023 Design, Automation \& Test in Europe Conference \&
  Exhibition (DATE)}, pages 1--6. IEEE, 2023.

\bibitem[Hong et~al.(2019)Hong, Frigo, Kaya, Giuffrida, and
  Dumitraș]{hong2019terminal}
Sanghyun Hong, Pietro Frigo, Yi{\u{g}}itcan Kaya, Cristiano Giuffrida, and
  Tudor Dumitraș.
\newblock Terminal brain damage: Exposing the graceless degradation in deep
  neural networks under hardware fault attacks.
\newblock In \emph{28th USENIX Security Symposium (USENIX Security 19)}, pages
  497--514, 2019.

\bibitem[Narayanan(2021)]{narayanan2021fault}
Niranjhana Narayanan.
\newblock \emph{Fault injection in Machine Learning applications}.
\newblock PhD thesis, University of British Columbia, 2021.

\bibitem[Ruospo et~al.(2020)Ruospo, Bosio, Ianne, and
  Sanchez]{ruospo2020evaluating}
Annachiara Ruospo, Alberto Bosio, Alessandro Ianne, and Ernesto Sanchez.
\newblock Evaluating convolutional neural networks reliability depending on
  their data representation.
\newblock In \emph{2020 23rd Euromicro Conference on Digital System Design
  (DSD)}, pages 672--679. IEEE, 2020.

\bibitem[Ruospo et~al.(2021)Ruospo, Sanchez, Traiola, O’connor, and
  Bosio]{ruospo2021investigating}
Annachiara Ruospo, Ernesto Sanchez, Marcello Traiola, Ian O’connor, and
  Alberto Bosio.
\newblock Investigating data representation for efficient and reliable
  convolutional neural networks.
\newblock \emph{Microprocessors and Microsystems}, 86:\penalty0 104318, 2021.

\bibitem[Syed et~al.(2021)Syed, Ulbricht, Piotrowski, and
  Krstic]{syed2021fault}
Rizwan~Tariq Syed, Markus Ulbricht, Krzysztof Piotrowski, and Milos Krstic.
\newblock Fault resilience analysis of quantized deep neural networks.
\newblock In \emph{2021 IEEE 32nd International Conference on Microelectronics
  (MIEL)}, pages 275--279. IEEE, 2021.

\bibitem[He et~al.(2020)He, Rakin, Li, Chakrabarti, and Fan]{he2020defending}
Zhezhi He, Adnan~Siraj Rakin, Jingtao Li, Chaitali Chakrabarti, and Deliang
  Fan.
\newblock Defending and harnessing the bit-flip based adversarial weight
  attack.
\newblock In \emph{Proceedings of the IEEE/CVF Conference on Computer Vision
  and Pattern Recognition}, pages 14095--14103, 2020.

\bibitem[Li et~al.(2020)Li, Rakin, Xiong, Chang, He, Fan, and
  Chakrabarti]{li2020defending}
Jingtao Li, Adnan~Siraj Rakin, Yan Xiong, Liangliang Chang, Zhezhi He, Deliang
  Fan, and Chaitali Chakrabarti.
\newblock Defending bit-flip attack through dnn weight reconstruction.
\newblock In \emph{2020 57th ACM/IEEE Design Automation Conference (DAC)},
  pages 1--6. IEEE, 2020.

\bibitem[Rakin et~al.(2019)Rakin, He, and Fan]{rakin2019bit}
Adnan~Siraj Rakin, Zhezhi He, and Deliang Fan.
\newblock Bit-flip attack: Crushing neural network with progressive bit search.
\newblock In \emph{Proceedings of the IEEE/CVF International Conference on
  Computer Vision}, pages 1211--1220, 2019.

\bibitem[Bach et~al.(2015)Bach, Binder, Montavon, Klauschen, M{\"u}ller, and
  Samek]{bach2015pixel}
Sebastian Bach, Alexander Binder, Gr{\'e}goire Montavon, Frederick Klauschen,
  Klaus-Robert M{\"u}ller, and Wojciech Samek.
\newblock On pixel-wise explanations for non-linear classifier decisions by
  layer-wise relevance propagation.
\newblock \emph{PloS one}, 10\penalty0 (7):\penalty0 e0130140, 2015.

\bibitem[Choi et~al.(2019)Choi, Shin, Park, and Ghosh]{choi2019sensitivity}
Wonseok Choi, Dongyeob Shin, Jongsun Park, and Swaroop Ghosh.
\newblock Sensitivity based error resilient techniques for energy efficient
  deep neural network accelerators.
\newblock In \emph{Proceedings of the 56th Annual Design Automation Conference
  2019}, pages 1--6, 2019.

\bibitem[Zhang et~al.(2022{\natexlab{b}})Zhang, Itsuji, Uezono, Toba, and
  Hashimoto]{zhang2022estimating}
Yangchao Zhang, Hiroaki Itsuji, Takumi Uezono, Tadanobu Toba, and Masanori
  Hashimoto.
\newblock Estimating vulnerability of all model parameters in dnn with a small
  number of fault injections.
\newblock In \emph{2022 Design, Automation \& Test in Europe Conference \&
  Exhibition (DATE)}, pages 60--63. IEEE, 2022{\natexlab{b}}.

\bibitem[Chen et~al.(2019)Chen, Li, Pattabiraman, and
  DeBardeleben]{chen2019binfi}
Zitao Chen, Guanpeng Li, Karthik Pattabiraman, and Nathan DeBardeleben.
\newblock Binfi: An efficient fault injector for safety-critical machine
  learning systems.
\newblock In \emph{Proceedings of the International Conference for High
  Performance Computing, Networking, Storage and Analysis}, pages 1--23, 2019.

\bibitem[Zhan et~al.(2021)Zhan, Sun, Jiang, Jiang, Yin, and
  Zhuo]{zhan2021improving}
Jinyu Zhan, Ruoxu Sun, Wei Jiang, Yucheng Jiang, Xunzhao Yin, and Cheng Zhuo.
\newblock Improving fault tolerance for reliable dnn using boundary-aware
  activation.
\newblock \emph{IEEE Transactions on Computer-Aided Design of Integrated
  Circuits and Systems}, 41\penalty0 (10):\penalty0 3414--3425, 2021.

\bibitem[BurelT et~al.(2022)BurelT, EvansT, and Anghel]{burelt2022improving}
St{\'e}phane BurelT, Adrian EvansT, and Lorena Anghel.
\newblock Improving dnn fault tolerance in semantic segmentation applications.
\newblock In \emph{2022 IEEE International Symposium on Defect and Fault
  Tolerance in VLSI and Nanotechnology Systems (DFT)}, pages 1--6. IEEE, 2022.

\bibitem[Esposito(2023)]{esposito2023reliability}
Giuseppe Esposito.
\newblock \emph{Reliability Evaluation of Split Computing Neural Networks}.
\newblock PhD thesis, Politecnico di Torino, 2023.

\bibitem[Govarini et~al.(2023)Govarini, Ruospo, and Sanchez]{govarini2023fast}
G~Govarini, Annachiara Ruospo, and E~Sanchez.
\newblock A fast reliability analysis of image segmentation neural networks
  exploiting statistical fault injections.
\newblock In \emph{2023 IEEE 24th Latin American Test Symposium (LATS)}, pages
  1--6. IEEE, 2023.

\bibitem[Sabena et~al.(2013)Sabena, Reorda, Sterpone, Rech, and
  Carro]{sabena2013evaluation}
Davide Sabena, M~Sonza Reorda, Luca Sterpone, Paolo Rech, and Luigi Carro.
\newblock On the evaluation of soft-errors detection techniques for gpgpus.
\newblock In \emph{2013 8th IEEE Design and Test Symposium}, pages 1--6. IEEE,
  2013.

\bibitem[Oliveira et~al.(2014)Oliveira, Rech, Quinn, Fairbanks, Monroe,
  Michalak, Anderson-Cook, Navaux, and Carro]{oliveira2014modern}
Daniel~AG Oliveira, Paolo Rech, Heather~M Quinn, Thomas~D Fairbanks, Laura
  Monroe, Sarah~E Michalak, Christine Anderson-Cook, Philippe~OA Navaux, and
  Luigi Carro.
\newblock Modern gpus radiation sensitivity evaluation and mitigation through
  duplication with comparison.
\newblock \emph{IEEE Transactions on Nuclear Science}, 61\penalty0
  (6):\penalty0 3115--3122, 2014.

\bibitem[Weigel et~al.(2017)Weigel, Fernandes, Navaux, and
  Rech]{weigel2017kernel}
Lucas Weigel, Fernando Fernandes, Philippe Navaux, and Paolo Rech.
\newblock Kernel vulnerability factor and efficient hardening for histogram of
  oriented gradients.
\newblock In \emph{2017 IEEE International Symposium on Defect and Fault
  Tolerance in VLSI and Nanotechnology Systems (DFT)}, pages 1--6. IEEE, 2017.

\bibitem[Libano et~al.(2018)Libano, Wilson, Anderson, Wirthlin, Cazzaniga,
  Frost, and Rech]{libano2018selective}
Fabiano Libano, Brittany Wilson, J~Anderson, Michael~J Wirthlin, Carlo
  Cazzaniga, Christopher Frost, and Paolo Rech.
\newblock Selective hardening for neural networks in fpgas.
\newblock \emph{IEEE Transactions on Nuclear Science}, 66\penalty0
  (1):\penalty0 216--222, 2018.

\bibitem[Bolchini et~al.(2022)Bolchini, Cassano, Miele, and
  Nazzari]{bolchini2022selective}
Cristiana Bolchini, Luca Cassano, Antonio Miele, and Alessandro Nazzari.
\newblock Selective hardening of cnns based on layer vulnerability estimation.
\newblock In \emph{2022 IEEE International Symposium on Defect and Fault
  Tolerance in VLSI and Nanotechnology Systems (DFT)}, pages 1--6. IEEE, 2022.

\bibitem[dos Santos et~al.(2018)dos Santos, Pimenta, Lunardi, Draghetti, Carro,
  Kaeli, and Rech]{dos2018analyzing}
Fernando~Fernandes dos Santos, Pedro~Foletto Pimenta, Caio Lunardi, Lucas
  Draghetti, Luigi Carro, David Kaeli, and Paolo Rech.
\newblock Analyzing and increasing the reliability of convolutional neural
  networks on gpus.
\newblock \emph{IEEE Transactions on Reliability}, 68\penalty0 (2):\penalty0
  663--677, 2018.

\bibitem[Fern{\'a}ndez et~al.(2022)Fern{\'a}ndez, Agirre, Perez-Cerrolaza,
  Abella, and Cazorla]{fernandez2022methodology}
Javier Fern{\'a}ndez, Irune Agirre, Jon Perez-Cerrolaza, Jaume Abella, and
  Francisco~J Cazorla.
\newblock A methodology for selective protection of matrix multiplications: A
  diagnostic coverage and performance trade-off for cnns executed on gpus.
\newblock In \emph{2022 6th International Conference on System Reliability and
  Safety (ICSRS)}, pages 9--18. IEEE, 2022.

\bibitem[dos Santos et~al.(2021)dos Santos, Brandalero, Sullivan, Basso,
  H{\"u}bner, Carro, and Rech]{dos2021reduced}
Fernando~F dos Santos, Marcelo Brandalero, Michael~B Sullivan, Pedro~M Basso,
  Michael H{\"u}bner, Luigi Carro, and Paolo Rech.
\newblock Reduced precision dwc: An efficient hardening strategy for
  mixed-precision architectures.
\newblock \emph{IEEE Transactions on Computers}, 71\penalty0 (3):\penalty0
  573--586, 2021.

\bibitem[Hoang et~al.(2020)Hoang, Hanif, and Shafique]{hoang2020ft}
Le-Ha Hoang, Muhammad~Abdullah Hanif, and Muhammad Shafique.
\newblock Ft-clipact: Resilience analysis of deep neural networks and improving
  their fault tolerance using clipped activation.
\newblock In \emph{2020 Design, Automation \& Test in Europe Conference \&
  Exhibition (DATE)}, pages 1241--1246. IEEE, 2020.

\bibitem[Chen et~al.(2021)Chen, Li, and Pattabiraman]{chen2021low}
Zitao Chen, Guanpeng Li, and Karthik Pattabiraman.
\newblock A low-cost fault corrector for deep neural networks through range
  restriction.
\newblock In \emph{2021 51st Annual IEEE/IFIP International Conference on
  Dependable Systems and Networks (DSN)}, pages 1--13. IEEE, 2021.

\bibitem[Ghavami et~al.(2022)Ghavami, Sadati, Fang, and
  Shannon]{ghavami2022fitact}
Behnam Ghavami, Mani Sadati, Zhenman Fang, and Lesley Shannon.
\newblock Fitact: Error resilient deep neural networks via fine-grained
  post-trainable activation functions.
\newblock In \emph{2022 Design, Automation \& Test in Europe Conference \&
  Exhibition (DATE)}, pages 1239--1244. IEEE, 2022.

\bibitem[Schorn et~al.(2019)Schorn, Guntoro, and Ascheid]{schorn2019efficient}
Christoph Schorn, Andre Guntoro, and Gerd Ascheid.
\newblock An efficient bit-flip resilience optimization method for deep neural
  networks.
\newblock In \emph{2019 Design, Automation \& Test in Europe Conference \&
  Exhibition (DATE)}, pages 1507--1512. IEEE, 2019.

\bibitem[Jang and Hong(2021)]{jang2021mate}
Myeungjae Jang and Jeongkyu Hong.
\newblock Mate: Memory-and retraining-free error correction for convolutional
  neural network weights.
\newblock \emph{Journal of Information \& Communication Convergence
  Engineering}, 19\penalty0 (1), 2021.

\bibitem[Burel et~al.(2021)Burel, Evans, and Anghel]{burel2021zero}
St{\'e}phane Burel, Adrian Evans, and Lorena Anghel.
\newblock Zero-overhead protection for cnn weights.
\newblock In \emph{2021 IEEE International Symposium on Defect and Fault
  Tolerance in VLSI and Nanotechnology Systems (DFT)}, pages 1--6. IEEE, 2021.

\bibitem[Lee et~al.(2022)Lee, Choi, and Yang]{lee2022bipolar}
Suyong Lee, Insu Choi, and Joon-Sung Yang.
\newblock Bipolar vector classifier for fault-tolerant deep neural networks.
\newblock In \emph{Proceedings of the 59th ACM/IEEE Design Automation
  Conference}, pages 673--678, 2022.

\bibitem[Gambardella et~al.(2022)Gambardella, Fraser, Zahid, Furano, and
  Blott]{gambardella2022accelerated}
Giulio Gambardella, Nicholas~J Fraser, Ussama Zahid, Gianluca Furano, and
  Michaela Blott.
\newblock Accelerated radiation test on quantized neural networks trained with
  fault aware training.
\newblock In \emph{2022 IEEE Aerospace Conference (AERO)}, pages 1--7. IEEE,
  2022.

\bibitem[Draghetti et~al.(2019)Draghetti, dos Santos, Carro, and
  Rech]{draghetti2019detecting}
Lucas~Klein Draghetti, Fernando~Fernandes dos Santos, Luigi Carro, and Paolo
  Rech.
\newblock Detecting errors in convolutional neural networks using inter frame
  spatio-temporal correlation.
\newblock In \emph{2019 IEEE 25th International Symposium on On-Line Testing
  and Robust System Design (IOLTS)}, pages 310--315. IEEE, 2019.

\bibitem[Javaheripi and Koushanfar(2021)]{9643556}
Mojan Javaheripi and Farinaz Koushanfar.
\newblock Hashtag: Hash signatures for online detection of fault-injection
  attacks on deep neural networks.
\newblock In \emph{2021 IEEE/ACM International Conference On Computer Aided
  Design (ICCAD)}, pages 1--9, 2021.
\newblock \doi{10.1109/ICCAD51958.2021.9643556}.

\bibitem[Guti{\'e}rrez-Zaballa et~al.(2023)Guti{\'e}rrez-Zaballa, Basterretxea,
  Echanobe, Mart{\'\i}nez, and Martinez-Corral]{gutierrez2023hsi}
Jon Guti{\'e}rrez-Zaballa, Koldo Basterretxea, Javier Echanobe, M~Victoria
  Mart{\'\i}nez, and Unai Martinez-Corral.
\newblock Hsi-drive v2. 0: More data for new challenges in scene understanding
  for autonomous driving.
\newblock In \emph{2023 IEEE Symposium Series on Computational Intelligence
  (SSCI)}, pages 207--214. IEEE, 2023.

\bibitem[AMD-Xilinx(2023)]{deviceReliabilityReport}
AMD-Xilinx.
\newblock Device reliability report. ug116 v10.17.
\newblock \url{https://docs.amd.com/r/en-US/ug116}, 2023.

\bibitem[Tian et~al.(2024)Tian, Ibrahim, Chen, Wang, Jin, Belev, and
  Chen]{TIAN2024115392}
Haonan Tian, Younis Ibrahim, Rui Chen, Yixiu Wang, Chen Jin, George Belev, and
  Li~Chen.
\newblock Comparative study: Autodpr-sem for enhancing cnn reliability in
  sram-based fpgas through autonomous reconfiguration.
\newblock \emph{Microelectronics Reliability}, 157:\penalty0 115392, 2024.
\newblock ISSN 0026-2714.
\newblock \doi{https://doi.org/10.1016/j.microrel.2024.115392}.
\newblock URL
  \url{https://www.sciencedirect.com/science/article/pii/S0026271424000726}.

\bibitem[Jacob et~al.(2018)Jacob, Kligys, Chen, Zhu, Tang, Howard, Adam, and
  Kalenichenko]{jacob2018quantization}
Benoit Jacob, Skirmantas Kligys, Bo~Chen, Menglong Zhu, Matthew Tang, Andrew
  Howard, Hartwig Adam, and Dmitry Kalenichenko.
\newblock Quantization and training of neural networks for efficient
  integer-arithmetic-only inference.
\newblock In \emph{Proceedings of the IEEE conference on computer vision and
  pattern recognition}, pages 2704--2713, 2018.

\bibitem[Vandersteegen et~al.(2021)Vandersteegen, Van~Beeck, and
  Goedem{\'e}]{vandersteegen2021integer}
Maarten Vandersteegen, Kristof Van~Beeck, and Toon Goedem{\'e}.
\newblock Integer-only cnns with 4 bit weights and bit-shift quantization
  scales at full-precision accuracy.
\newblock \emph{Electronics}, 10\penalty0 (22):\penalty0 2823, 2021.

\bibitem[Shen et~al.(2023)Shen, Mellempudi, He, Gao, Wang, and
  Wang]{shen2023efficient}
Haihao Shen, Naveen Mellempudi, Xin He, Qun Gao, Chang Wang, and Mengni Wang.
\newblock Efficient post-training quantization with fp8 formats.
\newblock \emph{arXiv preprint arXiv:2309.14592}, 2023.

\bibitem[Wu et~al.(2020)Wu, Judd, Zhang, Isaev, and
  Micikevicius]{wu2020integer}
Hao Wu, Patrick Judd, Xiaojie Zhang, Mikhail Isaev, and Paulius Micikevicius.
\newblock Integer quantization for deep learning inference: Principles and
  empirical evaluation.
\newblock \emph{arXiv preprint arXiv:2004.09602}, 2020.

\bibitem[Leveugle et~al.(2009)Leveugle, Calvez, Maistri, and
  Vanhauwaert]{statisticalFaultInjection}
R.~Leveugle, A.~Calvez, P.~Maistri, and P.~Vanhauwaert.
\newblock Statistical fault injection: Quantified error and confidence.
\newblock In \emph{2009 Design, Automation \& Test in Europe Conference \&
  Exhibition}, pages 502--506, 2009.
\newblock \doi{10.1109/DATE.2009.5090716}.

\bibitem[Zhu and Gupta(2017)]{zhu2017prune}
Michael Zhu and Suyog Gupta.
\newblock To prune, or not to prune: exploring the efficacy of pruning for
  model compression.
\newblock \emph{arXiv preprint arXiv:1710.01878}, 2017.

\bibitem[Nagel et~al.(2019)Nagel, Baalen, Blankevoort, and
  Welling]{nagel2019data}
Markus Nagel, Mart~van Baalen, Tijmen Blankevoort, and Max Welling.
\newblock Data-free quantization through weight equalization and bias
  correction.
\newblock In \emph{Proceedings of the IEEE/CVF International Conference on
  Computer Vision}, pages 1325--1334, 2019.

\end{thebibliography}

\end{document}